\documentclass[12pt]{article}
\usepackage{times}  
\usepackage{helvet} 
\usepackage{courier}  
\usepackage[hyphens]{url}  
\usepackage{graphicx} 
\usepackage{graphicx}  
\usepackage{amsmath}
\usepackage{booktabs}
\usepackage{algorithm}
\usepackage{amssymb}
\usepackage{subfig}
\usepackage{array}
\usepackage{multirow}
\usepackage{placeins}
\usepackage{natbib}
\usepackage{hyperref}
\usepackage{algorithmic}
\renewcommand{\cite}{\citep}

\usepackage[ruled,noresetcount,algo2e]{algorithm2e}
\newenvironment{sciabstract}{%
\begin{quote} \baselineskip14pt\small\hfil {\bf Abstract} \hfil\\[3pt]}
{\end{quote}\vspace{6pt}}

\newcounter{lastnote}

\title{NeoRL: A Near Real-World Benchmark for Offline Reinforcement Learning}

\author{
Rongjun Qin$^{1,2,\star}$, Songyi Gao$^{2,\star}$, Xingyuan Zhang$^{2,\star}$,
Zhen Xu$^2$,\\
Shengkai Huang$^2$,
Zewen Li$^{2}$,
Weinan Zhang$^{3}$,
Yang Yu$^{1,2,\diamond}$\\
\normalsize{
$^1$National Key Laboratory for Novel Software Technology, Nanjing University, China}\\
\normalsize{$^2$Polixir Technologies, China}\\
\normalsize{$^3$Shanghai Jiao Tong University, China}\\
\normalsize{
\{qinrj,xingyuan.zhang, songyi.gao, zhen.xu, shengkai.huang, zewen.li,yuy\}@polixir.ai,}\\
\normalsize{wnzhang@apex.sjtu.edu.cn}\\
\normalsize{$^\star$These authors contribute equally. $^\diamond$Corresponding author.}
}

\date{}

\begin{document}

\baselineskip16pt

\maketitle 

\begin{sciabstract}
	 Offline reinforcement learning (RL) aims at learning a good policy from a batch of collected data, without extra interactions with the environment during training. However, current offline RL benchmarks commonly have a large \emph{reality gap}, because they involve large datasets collected by highly exploratory policies, and the trained policy is directly evaluated in the environment. In real-world situations, running a highly exploratory policy is prohibited to ensure system safety, the data is commonly very limited, and a trained policy should be well validated before deployment.
	 In this paper, we present a \textbf{N}ear r\textbf{e}al-world \textbf{o}ffline \textbf{RL} benchmark, named NeoRL, which contains datasets from various domains with controlled sizes, and extra test datasets for policy validation. We evaluate existing offline RL algorithms on NeoRL and argue that the performance of a policy should also be compared with the deterministic version of the behavior policy, instead of the dataset reward. The empirical results demonstrate that the tested offline RL algorithms become less competitive to the deterministic policy on many datasets, and the offline policy evaluation hardly helps. The NeoRL suit can be found at \url{http://polixir.ai/research/neorl}.  We hope this work will shed some light on future research and draw more attention when deploying RL in real-world systems.
\end{sciabstract}

\section{Introduction}
Recent years have witnessed the great success of machine learning, especially deep learning systems, in computer vision, and natural language processing tasks. These tasks are usually based on a large dataset and can be divided into training and test phases. The deep learning algorithm updates its model, validates this model, and tunes its hyper-parameters on the training dataset. In general, the trained model will be evaluated on the \emph{unseen} test dataset before deployment. On the contrary, reinforcement learning (RL) agents interact with the environment, and collect trajectory data online maximize the expected return. Combined with deep learning, RL shows impressive ability in simulated environments without human knowledge \cite{DQN15,alphaGoZero}. However, beyond the scope of cheap simulated environments, current RL algorithms are hard to leverage in real-world applications, because the lack of a simulator makes it unrealistic to train an online RL agent in critical applications. Fortunately, the running systems will produce data, which come from expert demonstrations, human-designed rules, learned prediction models, and so on. The goal of offline RL (also batch RL) \cite{batchRL} is to learn an optimal policy from these static data, without extra online interactions. Thus it is a promising direction to scale RL to more real-world applications, such as industrial control, quantitative trading, and robotics, where online training agents may incur safety, cost, and ethical problems.

As deep learning benefits from a large dataset, current offline RL methods also assume a large batch of data at hand. The requirements of a large dataset limit the use of offline RL because collecting enough data will be both time-consuming and costly, especially in industrial tasks with huge policy spaces. Therefore, the out-of-data problem is more challenging in the low-data regime for offline RL, yet it is still crucial. The naive behavioral cloning (BC) approach can hardly outperform the behavior policy that produced the offline data, and the behavior policy is sub-optimal in general, so that BC is seldom applied in practice. Current offline RL methods are often pessimistic about the out-of-data distribution, and constrain the RL agent towards the behavior policy that produces the offline data, or use ensemble models to measure the uncertainty of out-of-data demonstrations \cite{BCQ,MOPO}. Although these offline RL approaches have achieved great success in previous benchmarks, we find them to be eclipsed by the simply implemented BC in near half of the tasks, when stepping further towards real-world settings.

\begin{figure*}[!h]
	\centering
	\includegraphics[width=0.8\textwidth,height=0.4\textwidth] {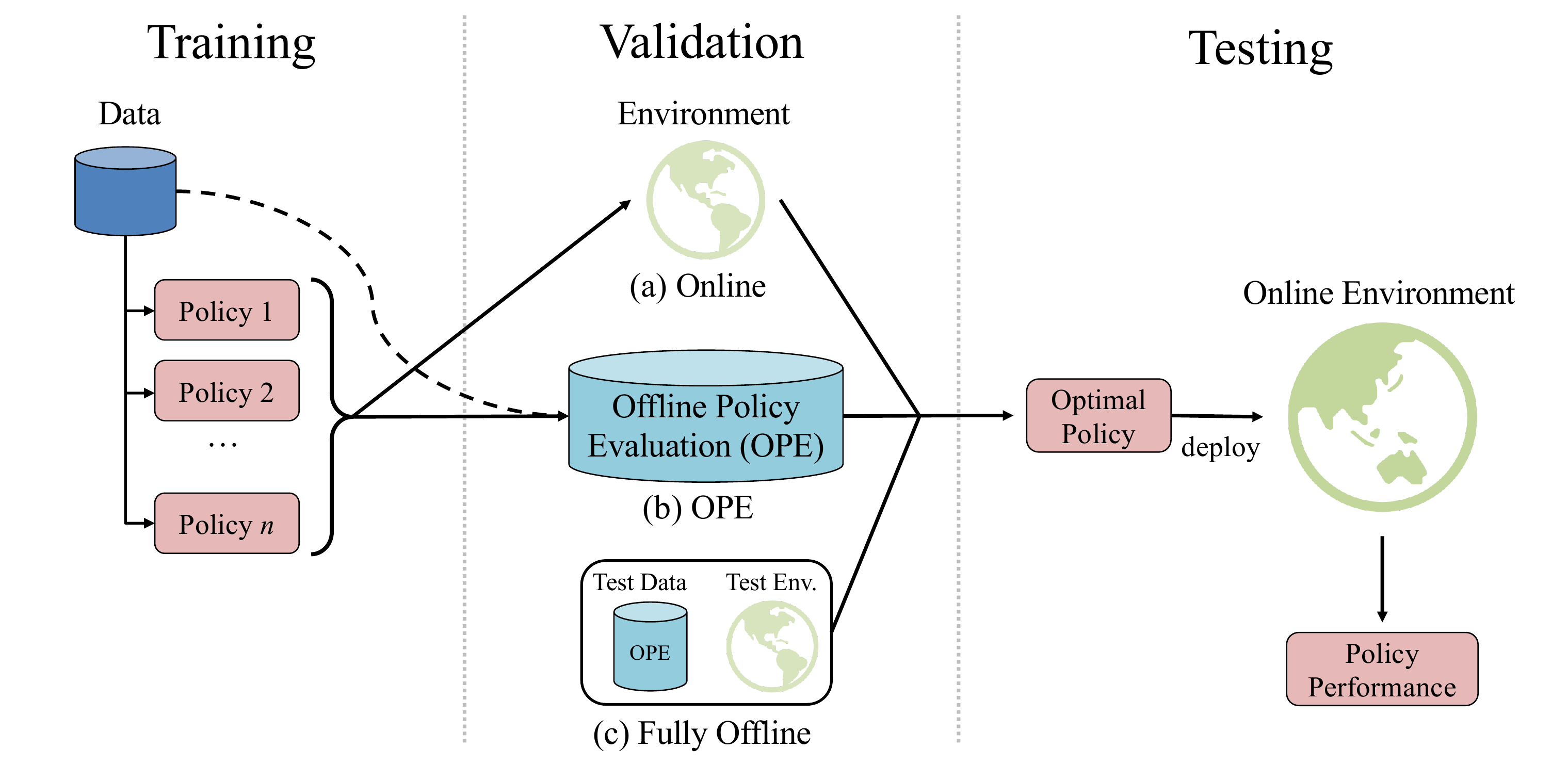}
	\caption{
		The pipeline of training and deploying offline RL, including training, validation (or offline test before deployment), and test (deploying) phases. In the validation phase, (a) uses the online environment to validate the trained policy. (b) uses offline policy evaluation models on training data. (c) uses offline policy evaluation method on an extra offline test data or uses a test environment, where the test environment can be learned from test data or uses other cheap simulators instead. After validating, an optimal policy is obtained and deployed in the online environment.
	}
	\label{diag}
\end{figure*}

Another critical issue is about evaluating and hyper-parameter selection for the trained policy before deploying. Online policy evaluation refers to directly run the trained policy in the original simulator or the production environment, while offline evaluation uses off-policy evaluation (OPE) methods \cite{studyOPE} or builds an evaluator through Bellman backup \cite{useFQE}. Figure \ref{diag} shows the pipeline of training and deploying offline RL. The online evaluation is more optimistic towards the trained policy since it allows perfect evaluation beforehand, thus is unrealistic to utilize in the production environment. Analogous to supervised learning, it is necessary to evaluate the trained RL agent before deployment, rather than directly running it in the real environment. Current benchmarks may use OPE methods on the training data, as in Figure \ref{diag} (b). It will be more reasonable to conduct OPE on unseen test data or unseen cheap test environment.

To tackle the above issues, we propose NeoRL, a suite of \textbf{n}ear r\textbf{e}al-world benchmarks for \textbf{o}ffline RL and provide a unified interface to access the data. The datasets include robotics, industrial control, finance trading, and city management tasks with real-world nature. Currently, we provide different sizes from low to high amounts, and different quality of data collected from corresponding simulators, and benchmark recently proposed model-free and model-based offline RL methods as a reference. The online and offline evaluations are both performed for selecting the hyper-parameters for each algorithm. Moreover, the running system commonly involves a working deterministic policy (we perturbed this policy to collect data from simulators), and the performance of the perturbed behavior policy, i.e., the reward on the dataset decreases. So the performances of these methods are compared with the deterministic behavior policy, and it appears competitive to baseline methods. The comparison results suggest that many of the current offline RL methods do not exceed this deterministic behavior policy significantly. Although offline evaluation before deployment is crucial, the current OPE method can hardly help when compared to random selection. We hope these findings will facilitate the design of offline RL algorithms for real-world applications.

\section{Offline Reinforcement Learning}
A reinforcement learning problem is typically formulated as a Markov decision process (MDP), with a 5-tuple $(S,A,P, r,\gamma)$ \cite{SuttonRL}. $S$ denotes the state space and $A$ is the action space. $P$ is the environment transition function, defined as $P(s^\prime|s,a): S\times A\times S \rightarrow [0,1]$. $r(s,a)$ is the reward function, and $\gamma\in(0,1 )$ is the discount factor. The return is $R=\sum_{t=0}^\infty \gamma^t r(s_t,a_t)$. The goal of RL is to learn a policy $\pi(a_t|s_t)$ for the agent, through interacting with the environment (also known as simulator), where the policy maximizes the expected return, i.e., the state-action value function $Q^\pi:=\mathbf{E}_{(s_t,a_t)\sim(\pi, P)}[\sum_{t=0}^\infty \gamma^t r(s_t, a_t)]$.

Traditional RL algorithms need to interact with the environment to collect trajectories with the current policy and update it, where the environment is treated as a black-box function. If the transition function is known, one can perform dynamic programming (DP) to obtain the optimal policy. Thus, the traditional model-based RL methods first build an environment model \cite{SuttonRL}. Another approach is model-free (e.g., Q-learning), and often outperforms the model-based methods in terms of asymptotic performance in practice. To improve the sample efficiency, current model-free off-policy methods, such as DQN \cite{DQN13}, need to store past trajectories in a replay buffer and use an off-policy algorithm to update the policy model.

In the offline RL setting, the environment is no longer available during training, and only a batch of static data is accessible. The data can be gathered by suboptimal expert policies and with noise. For simplicity, we denote the policy that collected the data as the behavior policy $\pi_b$. Although off-policy algorithms can be readily applied to a static replay buffer, running an off-policy RL algorithm on a static buffer can sometimes diverge, due to issues like the distribution shift \cite{BEAR}. To learn a robust policy, recent offline RL algorithms explicitly or implicitly prevent the training policy from being too disjoint with $\pi_b$ \cite{BCQ,CQL,MOPO,MORel}. Because the data is static and the policy has no access to the environment during training, real-world tasks also involve issues such as action delays and non-stationarities \cite{RWRL}. The absence of a cheap environment makes the evaluation of a training policy untamed. Offline policy evaluation (OPE) is subtly different from off-policy policy evaluation \cite{OffPolicyEval} based on importance sampling, thus novel techniques should be proposed to meet offline evaluation needs.

\section{Previous Benchmarks}
Recently, offline RL benchmarks have been proposed to facilitate the research and the evaluation of offline RL algorithms. These benchmarks include multiple aspects of offline tasks and datasets, and also the performance of prior offline algorithms on these tasks \cite{RLunplugged,D4RL,BenchBatchDRL}. The celebrated Atrai 57 games and MuJoCo tasks (or DeepMind Control Suite \cite{DMControlSuite}) have been widely used to benchmark online and offline RL methods. Besides these two kinds of tasks, D4RL \cite{D4RL} also releases offline datasets of maze, FrankaKitchen \cite{Replay_Learning}, and offline CARLA \cite{CARLA}, etc. These datasets in D4RL are designed to cover a range of challenging properties in real-world scenarios, including narrow and biased data distributions, multitask data, sparse rewards, suboptimal data, non-representable behavior policies, partial observability, and so on. RL Unplugged \cite{RLunplugged} included datasets from Atari and DM control suite, where the properties of these data range from different action spaces, observation spaces, partial observability, the difficulty of exploration, and real-world challenges. D4RL and RL unplugged both propose evaluation protocols. Some recent works utilize the training data or sample from the training data to make the offline datasets \cite{behavior_reg,BEAR,RLunplugged}. The performance of the trained policy is compared with the episode return of $\pi_b$, we find that for the purpose of exploration or increase the diversity, the performance of $\pi_b$ is worse than the deterministic version of $\pi_b$. So we compare some SOTA offline RL algorithms with the deterministic $\pi_b$, to verify the effectiveness of current algorithms. Although previous benchmarks provide diverse data and useful tools for evaluating the performance of offline RL algorithms, the reality gap still exists between them and real-world scenarios.

\section{The Reality Gap}
In real-world applications, the building of a high-fidelity simulator is unrealistic, e.g., the industrial control tasks. The data are directly gathered from the production environment, thus are often conservative and limited. The \emph{reality gap} exists in such scenarios for the following reasons:
\begin{itemize}
	\item\textbf{Conservative data}: Because of the cost and potential risks of random exploration, the operators in the production environment usually take conservative actions that stick to their knowledge passed from generation to generation. This will result in a less diverse dataset than current benchmarks. These datasets may come with different quality.
	
	\item\textbf{Limited available data}: Although previous works often assume that a large amount of logged data are easily obtained in real-world systems, it only holds for large-scale or streaming applications like recommendation systems. It is more challenging with a small batch of data to learn from.
	
	\item\textbf{Highly stochastic environments}: One of the difficulties of real-world environments is their stochastic (or non-stationary) nature. The environments may constantly evolve themselves or generate aleatoric uncertainty that makes credit assignment more difficult.
	
	\item\textbf{Offline evaluation before deployment}: The production environment is risk-sensitive and the policy must be evaluated before deployment. In supervised learning, the trained policy should be evaluated on an unseen test set before the final deployment. It should be the same for RL policy in real-world systems. However, current offline RL benchmarks use full data to train the model and propose protocols, such as OPE methods to conduct offline evaluation or conducting online evaluation through a different simulated environment that has similar dynamics \cite{D4RL,RLunplugged}. However, the different simulated environment evaluation approach somewhat contradicts offline property. If we have a cheap simulator that has similar dynamics, we can benefit more from this simulator, e.g., to pre-train a policy. Besides, it is unrealistic to conduct such validation providing only the production environment is available.
\end{itemize}

Thus, we provide various datasets and tasks to mimic these reality gap challenges in offline RL.

\section{Near Real-World Benchmarks}
To alleviate the above issues, we construct datasets with near real-world application properties. Compared to previous works, the included tasks take into account the above reality gap, and we simulate various complex situations that are likely to encounter in the real world, to build offline datasets with different nature.

\subsection{Near Real-World Environments}
Compared to existing environments, such as MuJoCo, the state and action space can be relatively large in real-world environments and the transition functions are more complex, with stronger stochasticity. Hence, we select tasks that are both high dimensional and with high stochasticity.  
In real scenarios, the historical datasets do not contain rewards. Rewards are calculated based on predefined quantifiable goals, e.g., a function of two successive states. Therefore, we encapsulate the reward function for each environment and provide an interface to use it. The interface is easy for researchers to customize the reward function. By using tasks that capture the nature of real-world environments, it could help offline RL step further towards the real world.

\subsection{Multi-Level Policy and Flexible Data Sizes}
The historical interaction data collected from the real world are often produced by domain experts with suboptimal policies, rather than from a random policy.
To simulate this gap, for each environment, we use SAC \cite{SAC} to train on the environment until convergence and record a policy at every epoch. We denote the policy with the highest episode return during the whole training as the expert policy. Another three levels of policies with around $25\%,50\%,75\%$ expert returns are stored to simulate multi-level suboptimal policies, denoted by low, medium, and high respectively. For each level, 4 policies with similar returns are selected, among which three policies are randomly selected to collect the training data used for offline RL policy training, and the left one produces the test data. The size of the test data is $1/10$ of the training data for each task. The extra test dataset can be used to design the offline evaluation method for the model selection during training and hyper-parameter selection. Because data are noisy demonstrations in general, to reproduce this phenomenon, with probability $20\%$, we explicitly add noise to the actions of the policy output.

To help verify the impact of different amounts of data on the learning ability of the algorithm, for each task, we provide training data with a maximum of $9999$ trajectories and three different sizes of $99$, $999$, and $9999$ trajectories by default. An interface is available to slice and shuffle the data set arbitrarily to meet specific demands.

\subsection{Unified Interface and More Comparisons}
Despite the tasks vary a lot, we provide a unified API on our datasets. Each item in one dataset consists of a tuple of state $s_t$, action $a_t$, reward $r_t$, next state $s_{t+1}$, and a unified interface for calling the reward calculation function for each task. The users can define their reward function if needed. We also benchmark current state-of-the-art offline RL algorithms on our datasets, and especially provide comparisons with the deterministic version of $\pi_b$.

\subsection{Tasks and Datasets}
\textbf{MuJoCo}  The MuJoCo \cite{MuJoCo} continuous control tasks are the standard testbeds for online reinforcement learning algorithms. We select three environments and construct the offline RL task, i.e., HalfCheetah-v3, Walker2d-v3, and Hopper-v3. The subtle difference is that we include the first dimension of the position. Because part of the reward function of these three environments is the distance moved forward, so adding the location information simplifies the reward calculation for the current step.

\textbf{IB} The industrial benchmark (IB) \cite{hein2017a} is an RL benchmark environment motivated to simulate the characteristics presented in various industrial control tasks, such as wind or gas turbines, chemical reactors, etc. It includes problems commonly encountered in real-world industrial environments, such as high-dimensional continuous state spaces, delayed rewards, complex noise patterns, and high stochasticity of multiple reactive targets. In the original IB, the instant reward is calculated with two dummy dimensions of the system state, and the left dimensions represent the observation. We use the system state as the observation. Since the industrial benchmark environment itself is a high-dimensional and highly stochastic environment, no explicit noise is added to the actions when collecting data on this environment, so the behavior policy is deterministic.

\textbf{FinRL} The FinRL environment \cite{liu2020finrl} provides a way to build a trading simulator that replicates the real stock market and supports backtesting with important market frictions such as transaction costs, market liquidity, and investor risk aversion, among other factors. In the FinRL environment, one trade can be made per trading day for the stocks in the pool (30 stocks). The reward function is the difference in the total asset value between the end of the day and the day before. The environment may evolve itself as time elapsed. Because the dataset of 9999 trajectories is too large, we only provide 99 and 999 trajectories for FinRL.

\textbf{CityLearn} The CityLearn (CL) environment \cite{canteli2019citylearn} reshapes the aggregation curve of electricity demand by controlling energy storage in different types of buildings.
The goal of optimization is to coordinate the control of domestic hot water and chilled water storage by the electricity consumers (i.e., buildings) to reshape the overall curve of electricity demand. This environment is highly stochastic and with high-dimensional space.

We defer the detailed feature of IB, FinRL and CityLearn environment to the appendix.

\section{Experiments}
To make fair comparisons for all the offline RL algorithms, the code reproducibility is the first to consider. However, publicly available codes are usually implemented with specific frameworks for some purpose, and these algorithms are highly coupled with their project codes. To focus on the algorithms themselves and easy to call them by a unified interface, we re-implement several algorithms (codes can be found in supplementary materials). The re-implementation has been verified on some of D4RL dataset and matches the result. We roughly divide these algorithms into two categories. Since offline RL algorithms are sensitive to the choice of hyper-parameters, we conduct model selection during training to choose the best policy. Details of the hyper-parameters settings are deferred to the appendix.

\subsection{Comparing Methods}

\subsubsection{Baselines}

\textbf{Expert} We run SAC \cite{SAC} to the convergence in each environment to obtain a policy and call it \emph{expert}. Expert is used as a reference of a good policy. However, it does not imply that the expert is optimal.

\textbf{Deterministic Policy} In real-world scenarios, it is common that the running system involves a working deterministic policy. This is the deterministic policy in our experiments.

\textbf{Behavior Policy} The behavior policy is used to collect the data. If the offline data collection process has no randomness injected, the behavior policy equals the deterministic policy. However, in many situations we would like to randomize the deterministic policy to collect a wider range of data. 

\subsubsection{Model-Free Methods}

Most algorithms in current offline RL favor a model-free fashion, especially, by extending from off-policy algorithms. Since offline RL is learning from a fixed static dataset, directly utilizing off-policy algorithms will suffer from distribution shift \cite{levine2020offline} or extrapolation error \cite{BCQ}, where the training policies try to reach out-of-data states and actions. For this reason, model-free algorithms usually explicitly or implicitly constrain the learned policy to be close to the offline data \cite{BCQ,CQL} (or the behavior policy).

\textbf{BC} Behavioral cloning trains a policy to mimic the behavior policy from the data. BC is straightforward and does not require any interactions with the environment. We treat BC as a baseline of learning method.

\textbf{BCQ} BCQ \cite{BCQ} learns a state-conditioned generative model $G_\omega(s)$, i.e., VAE, to mimic the behavior policy of the dataset, and a perturbation network $\xi_\phi(s,a,\Phi)$ to generate actions $\{a_i=a_i+\xi_\phi(s^\prime,a_i,\Phi)\}_{i=1}^n$, where $\{a_i\sim G_\omega (s^\prime)\}_{i=1}^n$ and the perturbation $\xi_\phi(s,a,\Phi)$ lies in the range $[-\Phi,\Phi]$. Controlling the perturbation amount by a hyper-parameter $\Phi$, the learned policy is constrained near the original behavior policy.

\textbf{CQL} CQL \cite{CQL} penalizes the value function for states and actions that are not supported by the data to prevent overestimation of the training policy. By introducing a new Q-value maximization term under the offline data distribution ($ \mathbf{E}_{s\sim \mathbf{D}, a \sim \hat{\pi}_b(s,a)} [Q(s,a)] $), CQL learns a \emph{conservative} Q function. The authors have also proved this additional term helps achieve a tighter lower bound on the expected Q-value of the training policy $\pi$.

\textbf{PLAS} PLAS \cite{PLAS} is an extension of BCQ. Instead of learning a perturbation model on the action space, PLAS learns a deterministic policy on the latent space of VAE and assumes that the latent action space implicitly defines a constraint over the action output, thus the policy selects actions within the support of the dataset during training. In PLAS architecture, actions are decoded from latent actions. An optional perturbation layer can be applied in the PLAS architecture to improve the out-of-data generalization, akin to the perturbation model in BCQ.

\begin{table*}
\centering
\caption{The online evaluation results. Policies are measured by their episode return and averaged over 1000 episodes. Bold numbers indicate the best result for each task, while numbers marked by $^*$ indicate results worse than BC. The task name is composed of the specific task, the quality of dataset, and the number of trajectories. L, M, and H stands for low, medium, high respectively. Det. is abbreviation of deterministic. The w/ and w/o refer to with and without.}
\label{benchmark_online}
\scalebox{0.68}{\begin{tabular}{lc|cc|cccccccc}
\toprule
Task & \begin{tabular}[c]{@{}c@{}}Expert \\ Policy\end{tabular} & \begin{tabular}[c]{@{}c@{}}Det. \\ Policy\end{tabular} & \begin{tabular}[c]{@{}c@{}}Behavior \\ Policy\end{tabular} & Random & BC & CQL & PLAS & BCQ & MOPO & \begin{tabular}[c]{@{}c@{}}MB-PPO \\ w/ KL\end{tabular} & \begin{tabular}[c]{@{}c@{}}MB-PPO \\ w/o KL\end{tabular} \\
\midrule
HalfCheetah-v3-L-99 & 12284 & 3195 & 2871 & -298 & 3260 & 3792 & 3451 & 3363 & \textbf{5059} & 3228$^*$ & 52$^*$ \\
HalfCheetah-v3-L-999 & 12284 & 3195 & 2871 & -298 & 3200 & 4457 & 3572 & 3524 & \textbf{4760} & 3261 & 2324$^*$ \\
HalfCheetah-v3-L-9999 & 12284 & 3195 & 2871 & -298 & 3295 & 4761 & 3587 & 3509 & \textbf{4971} & 3257$^*$ & 2190$^*$ \\
HalfCheetah-v3-M-99 & 12284 & 6027 & 5568 & -298 & 5943 & 6106 & 5716$^*$ & 4950$^*$ & \textbf{7182} & 5770$^*$ & 1307$^*$ \\
HalfCheetah-v3-M-999 & 12284 & 6027 & 5568 & -298 & 6077 & 6580 & 5888$^*$ & 5664$^*$ & \textbf{7958} & 6060$^*$ & -9$^*$ \\
HalfCheetah-v3-M-9999 & 12284 & 6027 & 5568 & -298 & 6083 & 6455 & 5960$^*$ & 5318$^*$ & \textbf{6943} & 5853$^*$ & 170$^*$ \\
HalfCheetah-v3-H-99 & 12284 & 9020 & 7836 & -298 & \textbf{8964} & 8643$^*$ & 7057$^*$ & 7312$^*$ & 2740$^*$ & -140$^*$ & -57$^*$ \\
HalfCheetah-v3-H-999 & 12284 & 9020 & 7836 & -298 & 9136 & \textbf{9346} & 8110$^*$ & 8689$^*$ & 2857$^*$ & 9123$^*$ & -89$^*$ \\
HalfCheetah-v3-H-9999 & 12284 & 9020 & 7836 & -298 & \textbf{8969} & 8738$^*$ & 8164$^*$ & 2976$^*$ & 1335$^*$ & 8959$^*$ & 574$^*$ \\
\midrule
Hopper-v3-L-99 & 3294 & 508 & 498 & 5 & 515 & 527 & 527 & 545 & 170$^*$ & 423$^*$ & \textbf{995} \\
Hopper-v3-L-999 & 3294 & 508 & 498 & 5 & 514 & 551 & 677 & 624 & 1017 & \textbf{1055} & 131$^*$ \\
Hopper-v3-L-9999 & 3294 & 508 & 498 & 5 & 511 & 519 & 578 & 583 & 150$^*$ & 855 & \textbf{1047} \\
Hopper-v3-M-99 & 3294 & 1530 & 1410 & 5 & 1734 & \textbf{2155} & 2025 & 1717$^*$ & 16$^*$ & 1895 & 245$^*$ \\
Hopper-v3-M-999 & 3294 & 1530 & 1410 & 5 & 1528 & \textbf{2766} & 1682 & 1630 & 20$^*$ & 2312 & 53$^*$ \\
Hopper-v3-M-9999 & 3294 & 1530 & 1410 & 5 & 1669 & \textbf{3004} & 2518 & 2020 & 180$^*$ & 2957 & 1005$^*$ \\
Hopper-v3-H-99 & 3294 & 2294 & 1551 & 5 & 1325 & 2207 & 2711 & 2072 & 129$^*$ & \textbf{3184} & 986$^*$ \\
Hopper-v3-H-999 & 3294 & 2294 & 1551 & 5 & 2354 & 2874 & 2544 & \textbf{3022} & 261$^*$ & 1894$^*$ & 876$^*$ \\
Hopper-v3-H-9999 & 3294 & 2294 & 1551 & 5 & 2366 & 2860 & \textbf{3042} & 1885$^*$ & 201$^*$ & 2458 & 1002$^*$ \\
\midrule
Walker2d-v3-L-99 & 5143 & 1572 & 1278 & 1 & 1749 & \textbf{2370} & 42$^*$ & 1407$^*$ & 705$^*$ & 1753 & -25$^*$ \\
Walker2d-v3-L-999 & 5143 & 1572 & 1278 & 1 & 1433 & 2015 & -8$^*$ & 1844 & 381$^*$ & \textbf{2193} & 1044$^*$ \\
Walker2d-v3-L-9999 & 5143 & 1572 & 1278 & 1 & 1470 & \textbf{2375} & 247$^*$ & 1546 & 836$^*$ & 1473 & 0$^*$ \\
Walker2d-v3-M-99 & 5143 & 2547 & 2221 & 1 & 2801 & 2647$^*$ & 201$^*$ & 2745$^*$ & 1760$^*$ & \textbf{2996} & 303$^*$ \\
Walker2d-v3-M-999 & 5143 & 2547 & 2221 & 1 & 2507 & \textbf{2977} & 2601 & 2921 & 731$^*$ & 2814 & 1047$^*$ \\
Walker2d-v3-M-9999 & 5143 & 2547 & 2221 & 1 & 2566 & \textbf{3195} & 1877$^*$ & 2928 & 686$^*$ & 2494$^*$ & 915$^*$ \\
Walker2d-v3-H-99 & 5143 & 3550 & 2936 & 1 & 3406 & \textbf{3875} & 3327$^*$ & 3712 & 979$^*$ & 3326$^*$ & -7$^*$ \\
Walker2d-v3-H-999 & 5143 & 3550 & 2936 & 1 & 3593 & \textbf{3950} & 2327$^*$ & 3884 & 994$^*$ & 3884 & 523$^*$ \\
Walker2d-v3-H-9999 & 5143 & 3550 & 2936 & 1 & 3704 & 3832 & 2394$^*$ & 3693$^*$ & 99$^*$ & \textbf{4004} & 1044$^*$ \\
\midrule
IB-L-99 & -180240 & -344311 & -344311 & -317624 & -333400 & -298161 & -341327$^*$ & -410860$^*$ & -379405$^*$ & \textbf{-230455} & -278524 \\
IB-L-999 & -180240 & -344311 & -344311 & -317624 & -339959 & -341099$^*$ & -338732 & -410531$^*$ & -372254$^*$ & \textbf{-220832} & -329973 \\
IB-L-9999 & -180240 & -344311 & -344311 & -317624 & -340716 & -323374 & -322796 & -407141$^*$ & -409110$^*$ & -339899 & \textbf{-310140} \\
IB-M-99 & -180240 & -283121 & -283121 & -317624 & -281225 & -277511 & -410918$^*$ & -410196$^*$ & -345350$^*$ & -234696 & \textbf{-217471} \\
IB-M-999 & -180240 & -283121 & -283121 & -317624 & -382304 & -279299 & -283242 & -862628$^*$ & -381155 & -283838 & \textbf{-223842} \\
IB-M-9999 & -180240 & -283121 & -283121 & -317624 & -277010 & -282285$^*$ & -410751$^*$ & -410820$^*$ & -406456$^*$ & \textbf{-218091} & -235499 \\
IB-H-99 & -180240 & -220156 & -220156 & -317624 & -217240 & -223178$^*$ & -410869$^*$ & -406571$^*$ & -410431$^*$ & -272607$^*$ & \textbf{-213269} \\
IB-H-999 & -180240 & -220156 & -220156 & -317624 & -220528 & -213588 & -411404$^*$ & -410517$^*$ & -314221$^*$ & \textbf{-207958} & -257114$^*$ \\
IB-H-9999 & -180240 & -220156 & -220156 & -317624 & \textbf{-220370} & -280470$^*$ & -222682$^*$ & -410618$^*$ & -410726$^*$ & -261822$^*$ & -224130$^*$ \\
\midrule
FinRL-L-99 & 631 & 150 & 152 & 206 & 136 & \textbf{487} & 447 & 330 & 369 & 136 & 328 \\
FinRL-L-999 & 631 & 150 & 152 & 206 & 137 & 416 & 396 & 323 & 341 & \textbf{752} & 656 \\
FinRL-M-99 & 631 & 300 & 357 & 206 & 355 & 700 & 388 & 376 & 357 & 593 & \textbf{1213} \\
FinRL-M-999 & 631 & 300 & 357 & 206 & 504 & 621 & 470$^*$ & 356$^*$ & 373$^*$ & 504 & \textbf{698} \\
FinRL-H-99 & 631 & 441 & 419 & 206 & 252 & \textbf{671} & 464 & 426 & 531 & 640 & 484 \\
FinRL-H-999 & 631 & 441 & 419 & 206 & 270 & 444 & 495 & 330 & 373 & 581 & \textbf{787} \\
\midrule
CL-L-99 & 50350 & 28500 & 29514 & 16280 & 29420 & \textbf{30670} & 29918 & 23238$^*$ & 21135$^*$ & 29902 & 23326$^*$ \\
CL-L-999 & 50350 & 28500 & 29514 & 16280 & 30317 & \textbf{31611} & 30141$^*$ & 30451 & 21756$^*$ & 23528$^*$ & 23019$^*$ \\
CL-L-9999 & 50350 & 28500 & 29514 & 16280 & 30231 & \textbf{32285} & 30716 & 25107$^*$ & 21396$^*$ & 23293$^*$ & 23431$^*$ \\
CL-M-99 & 50350 & 37800 & 36900 & 16280 & 27422 & 39551 & \textbf{40957} & 31398 & 21553$^*$ & 18606$^*$ & 23166$^*$ \\
CL-M-999 & 50350 & 37800 & 36900 & 16280 & 39132 & 42737 & \textbf{54742} & 25320$^*$ & 22586$^*$ & 38951$^*$ & 23093$^*$ \\
CL-M-9999 & 50350 & 37800 & 36900 & 16280 & 38331 & \textbf{42917} & 40416 & 37222$^*$ & 21499$^*$ & 23365$^*$ & 23467$^*$ \\
CL-H-99 & 50350 & 48600 & 48818 & 16280 & 53071 & \textbf{55158} & 53402 & 43254$^*$ & 24867$^*$ & 23289$^*$ & 23733$^*$ \\
CL-H-999 & 50350 & 48600 & 48818 & 16280 & 54622 & 43437$^*$ & 54397$^*$ & 37040$^*$ & 22151$^*$ & \textbf{54900} & 23288$^*$ \\
CL-H-9999 & 50350 & 48600 & 48818 & 16280 & 52957 & 54696 & 55166 & 48427$^*$ & 21849$^*$ & \textbf{56874} & 23284$^*$ \\
\midrule
Average Rank & - & 5.06 & 6.04 & 9.04 & 4.63 & 2.45 & 4.61 & 5.63 & 7.08 & 3.90 & 6.29\\
\bottomrule
\end{tabular}
}
\end{table*}

\begin{table*}[!h]
\centering
\caption{Ratio of winning the 3 baselines over the 51 tasks by online evaluation.}
\label{benchmark_baseline}
\scalebox{0.88}{\begin{tabular}{lcccccc}
\toprule
Baseline                  & CQL      & PLAS    & BCQ      & MOPO  & MB-PPO w/ KL & MB-PPO w/o KL \\
\midrule
Behavior Policy        & 94.12\% & 72.55\% & 56.86\% & 21.57\% & 78.43\% & 29.41\% \\
Deterministic Policy & 88.24\%  & 60.78\% & 43.14\% & 23.53\% & 64.71\% & 29.41\% \\
BC                          & 84.31\% & 54.90\% & 45.10\% & 25.49\% & 56.86\% & 29.41\% \\\bottomrule
\end{tabular}
}
\end{table*}

\subsubsection{Model-Based Methods}
Although model-free methods leverage in offline RL algorithms and are easy to use, an over-constrained policy can hinder stronger results, especially when the data is collected by a low-performance behavior policy. On the other hand, model-based methods learn the environment transition dynamics, which depends less on the quality of the behavior policy $\pi_b$. The transition model takes $(s, a)$ pair as input and outputs next state $s^\prime$, thus online RL algorithms can use these models as simulators to rollout or plan.

\textbf{MB-PPO} A naive approach to use these transition models, is to treat them as the environments and run RL algorithms. We implement a simple variant named Model-Based PPO (MB-PPO). A transition model is trained via maximum likelihood and not updated during the policy training process. When generating new trajectories with the learned transition, we sample a mini-batch of states from the dataset, then rollout with the training policy for 10 steps. We choose PPO \cite{schulman2017proximal} for policy improvement, because the training process is more stable. To mitigate the model exploitation, we also enforce the learned policy to stay close to the behavior policy. In practice, we use BC to recover the behavior policy $\hat{\pi}_b$, and use it to initialize. A penalty term $D_{KL}(\hat{\pi}_b\| \pi)$ can be added to the policy training loss.

\textbf{MOPO} To alleviate the effect of erroneous predictions for out-of-distribution data, MOPO \cite{MOPO} uses an ensemble of transition models to estimate the uncertainty of these predictions. When generating rollouts from the transition models, the reward is penalized by the uncertainty term to encourage the policy to explore states that the transition models are certain about.

\subsection{Online Model Selection}
A straightforward way to evaluate an offline RL algorithm is to run the learned policy in the environment and calculate the mean episode return. We firstly follow this online evaluation, and use the trained policy to interact with the environment for 1,000 episodes and report the mean return.

We select the best final policy through online evaluation, among the hyper-parameter range for each task. The results are reported in Table \ref{benchmark_online}. From Table \ref{benchmark_online}, in most tasks, BC matches the performance of the deterministic policy, which indicates that BC recovered the deterministic behavior policy from noisy data. Interestingly, results of BC form very strong baselines: the other four offline RL algorithms fail to outperform BC in 153 out of 306 configurations. Among all the offline RL algorithms, CQL achieves the best performance on 1/3 of the tasks. However, its performance improvement over the deterministic policy is not that significant anymore on MuJoco and IB environments.

For model-based approaches, the overall performance is worse than model-free methods. In many tasks, MB-PPO without the KL constraint achieves the lowest score, some are even lower than the random policy. The learned environment model in MB-PPO is evaluated on the test data, and we choose the one with the lowest mean square error as the environment for policy training. The results of MB-PPO imply the learned model suffers from model exploitation. With the KL constraint, MB-PPO is competitive on most tasks. On the other hand, with ensemble environment models and reward penalty, MOPO alleviates the model exploitation and raises its performance. MOPO achieves significant improvements over the other methods on HalfCheetah-v3-Low and HalfCheetah-v3-Medium, which reveals the potential of model-based offline RL approaches. However, the dataset can be less diverse as the quality improves, which may incur bias in environment learning and lead to a poorer performance on high quality dataset. Besides, as noted by the authors, MOPO is sensitive to the choice of rollout length $h$ and the penalty coefficient $\lambda$, so MOPO may need to be well-tuned to achieve a better performance on other tasks.

The last row calculates the average value of the rank of a method in each of the 51 tasks. Using the Nemenyi test \cite{demvar.jmlr06}, the critical difference of 10 comparing methods over 51 tasks with confidence level 95\% is $1.8970$. Therefore, if we take BC as the reference, only CQL is significantly better than BC, and Random and MOPO are significantly worse, which is the same if we take the deterministic policy as the reference. 

Table \ref{benchmark_baseline} summarizes the winning rates of the offline RL methods to the behavior policy, the deterministic policy, and BC. It can be observed that the winning rates can have a large drop from the comparison with the behavior policy to the comparison with the deterministic policy. Note that a running system commonly has a working deterministic policy, the goal of offline RL should be to exceed the deterministic policy, but not only to exceed the behavior policy (i.e., the dataset reward). Also, we can observe that many methods are not significantly better than BC.

It should be also noted that for almost all the algorithms, the final performance of the trained policy is not strictly monotonous to neither the quality of data nor the amount of data, and none of the algorithms outperform others consistently. We conjecture this is rooted in data distribution from the environment nature and the sensitivity to the hyper-parameters of different algorithms. This result may hinder the choice of proper algorithms in real-world applications.

\subsection{Offline Model Selection}

Online evaluation validates the actual performance of given policies when deployed, yet it is not practical in real-world scenarios. The main drawback is the performance of the trained policy is almost unknown before deployment. Therefore, offline evaluation is crucial for real-world deployment. We then report the results with offline model selection.

Due to the limited data, predicting the approximate accumulated rewards can be challenging. In our settings, we use offline policy evaluation (OPE) to select the best policy among policies trained by different hyper-parameters (hyper-parameter selection) and among different training stages (model selection). To select the best model beforehand, an effective OPE method only needs to tell the relative performance between policies, rather than approximating the ground-truth performance to some extent.

We utilize an offline evaluation method called fitted Q evaluation (FQE) \cite{FQE}. FQE takes a policy as input and performs policy evaluation on the fixed dataset by Bellman backup. After learning the Q function of the policy, the performance is measured by the mean Q values on the states sampled from the dataset and actions by the policy. As shown in the original paper \cite{FQE}, the distributional critic does not help improve the performance, so we implement FQE without distributional critic and use the same hyper-parameters. We first conduct the experiment on Walker2d-v3 task. Following \cite{useFQE}, we use two metrics to measure the OPE methods:
%

\textbf{Rank Correlation Score (RC Score)}: RC score indicates how the OPE produces the same rank as the ground-truth in the online evaluation. It is computed as Spearman correlation coefficient between the two rankings produced by OPE and online evaluation respectively. RC score lies in $[-1, 1]$, and if the rank is uniformly random, the score will be $0$.

\textbf{Top-$K$ Score}: Top-$K$ score represents the relative performance of the chosen $K$ policies via OPE. To compute this score, the real online performance of each policy is first normalized to a score within $[0, 1]$ by the min and max values over the whole candidate policy set of all the algorithms. Let $\pi_{\text{off}}^k$ denote the $k$-th ranked policy by the offline evaluation, then we use $\frac{1}{K}\sum_{k=1}^K \pi_{\text{off}}^k$ and $\max_k\{\pi_{\text{off}}^k\}$ as the mean and max top-$K$ score respectively.

The online model selection can obtain the ground-truth performance of a policy, so it can be treated as an upper bound of the offline policy selection. We use FQE on the Walker2d-v3 tasks and compare the evaluation results with the online evaluation. We store the policies produced by each hyper-parameter profile for every 25 epochs, thus keep hundreds of policies for every task.
Considered that some policies in the early stage of the training process can fluctuate, we compute the scores on the policies from the latter half of the training process. The policy set includes all trained policies from all the offline algorithms.

\begin{table}[!h]
\centering
\caption{FQE performance on the  latter half of policies. L, M, H stands for low, medium and high quality of dataset. Each item is the mean and std over 3 random seeds of FQE runs except that the average item is the mean and std of 27 scores.}
\label{fqe_50}
\scalebox{0.65}{\begin{tabular}{lrccccccc}
\toprule
Task Name & \begin{tabular}[c]{@{}c@{}}RC \\ Score\end{tabular} & \begin{tabular}[c]{@{}c@{}}Top-1 \\ Mean \\ Score\end{tabular} & \begin{tabular}[c]{@{}c@{}}Top-3 \\ Mean \\ Score\end{tabular} & \begin{tabular}[c]{@{}c@{}}Top-5 \\ Mean \\ Score\end{tabular} & \begin{tabular}[c]{@{}c@{}}Top-1 \\ Max \\ Score\end{tabular} & \begin{tabular}[c]{@{}c@{}}Top-3 \\ Max \\ Score\end{tabular} & \begin{tabular}[c]{@{}c@{}}Top-5 \\ Max \\ Score\end{tabular} & \begin{tabular}[c]{@{}c@{}}Policy \\ Mean \\ Score\end{tabular}\\
\midrule
Walker2d-v3-L-99 & $-.282 \pm .062$ & $.182 \pm .131$ & $.172 \pm .019$ & $.165 \pm .022$ & $.182 \pm .131$ & $.366 \pm .000$ & $.384 \pm .012$ & $.335 \pm .000$ \\
Walker2d-v3-L-999 & $.118 \pm .092$ & $.039 \pm .034$ & $.044 \pm .044$ & $.054 \pm .030$ & $.039 \pm .034$ & $.083 \pm .096$ & $.161 \pm .103$ & $.390 \pm .000$ \\
Walker2d-v3-L-9999 & $.341 \pm .025$ & $.009 \pm .001$ & $.035 \pm .015$ & $.047 \pm .009$ & $.009 \pm .001$ & $.054 \pm .019$ & $.084 \pm .002$ & $.427 \pm .000$ \\
Walker2d-v3-M-99 & $-.152 \pm .080$ & $.264 \pm .353$ & $.297 \pm .027$ & $.357 \pm .080$ & $.264 \pm .353$ & $.832 \pm .097$ & $.887 \pm .089$ & $.473 \pm .000$ \\
Walker2d-v3-M-999 & $-.131 \pm .038$ & $.006 \pm .001$ & $.091 \pm .106$ & $.143 \pm .095$ & $.006 \pm .001$ & $.240 \pm .306$ & $.387 \pm .262$ & $.466 \pm .000$ \\
Walker2d-v3-M-9999 & $.101 \pm .030$ & $.031 \pm .015$ & $.025 \pm .007$ & $.022 \pm .004$ & $.031 \pm .015$ & $.043 \pm .001$ & $.043 \pm .001$ & $.459 \pm .000$ \\
Walker2d-v3-H-99 & $-.116 \pm .078$ & $.274 \pm .379$ & $.364 \pm .126$ & $.397 \pm .137$ & $.274 \pm .379$ & $.721 \pm .125$ & $.734 \pm .136$ & $.481 \pm .000$ \\
Walker2d-v3-H-999 & $-.216 \pm .083$ & $.006 \pm .001$ & $.005 \pm .001$ & $.005 \pm .001$ & $.006 \pm .001$ & $.006 \pm .001$ & $.007 \pm .000$ & $.388 \pm .000$ \\
Walker2d-v3-H-9999 & $.296 \pm .015$ & $.005 \pm .000$ & $.007 \pm .004$ & $.068 \pm .086$ & $.005 \pm .000$ & $.013 \pm .011$ & $.315 \pm .420$ & $.448 \pm .000$ \\
\midrule
Average & $-.005 \pm .222$ & $.091 \pm .209$ & $.116 \pm .138$ & $.140 \pm .153$ & $.091 \pm .209$ & $.262 \pm .321$ & $.334 \pm .340$ & $.430 \pm .046$ \\
\bottomrule
\end{tabular}}
\end{table}

The results latter half of candidate policies are reported in Table \@\ref{fqe_50} and we defer the table for whole policies to the appendix (Table \ref{fqe_full}). The mean policy score is the average performance of all these candidates in each task, and it is just the expected score of uniform random selection and is independent of FQE. In the last two rows, we calculate the average scores over the 9 tasks for the latter half and the whole policy set. It is noteworthy the RC score of the FQE method is even lower than the random selection in $5/9$ tasks. For Top-$K$ scores, as we increase $K$ from 1 to 5, the average max score increases, while the average mean Top-$K$ score does not change too much. On the whole policy set, the max scores of Top-$K$ and mean policy scores are lower than the latter half, which implies that FQE works better when the policies have been trained for a period. On the other hand, although the max score of Top-$K$ increases when increasing $K$, the mean score can have a sudden drop. It should be noted that $9/9$ Top-1 scores and $6/9$ Top-3 (and Top-5) max scores are lower than the mean policy scores, which implies FQE runs may be worse than uniform random selection. These results also indicate that we are still far away from reliably choosing the best policy to deploy fully offline.

\section{Conclusion}
\textbf{NeoRL.} In this paper we present NeoRL, a near real-world benchmark for offline RL. 
Since real-world datasets are usually very limited and collected with conservative policies to ensure system safety, for real-world considerations, NeoRL focuses on conservative actions, limited data, stochastic dynamics, and offline policy evaluation, which are ubiquitous and crucial in real-world decision-making scenarios.
So far, NeoRL has included MuJoCo locomotion controlling, industry controlling, finance trading, and city management tasks, where the training and test datasets are collected from these domains with different sizes.

\textbf{Findings.} 
We benchmark some state-of-the-art offline RL algorithms on NeoRL tasks, including model-free and model-based algorithms, in both online and offline policy evaluation manner. Surprisingly, the experimental results demonstrate that these compared offline RL algorithms fail to outperform neither the simplest behavior cloning method nor the deterministic behavior policy, only except CQL. Their performance may be extremely bounded by the data.

Our experiment results further show that model-based offline RL approaches are overall worse than model-free approaches. However, model-based approaches may have better potential to achieve the out-of-data generalization ability. Meanwhile, we have noticed that better model-learning approaches based on adversarial learning \cite{xu.li.yu.nips2020,vdidi.kdd19,vtaobao.aaai19} could help. We will test these approaches in the future.

\textbf{Lessons learned.}
For real-world applications, the trained policy should be evaluated before being deployed. We recommend using offline policy evaluation methods on an unseen test dataset (or a cheap learned simulator) to evaluate the trained policy. Despite the importance of offline evaluation in real-world scenarios, it can be inferred from the experiments that current offline policy evaluation methods can hardly help improve the policy selection compared to uniform random selection. We argue that offline RL algorithms should pay more attention to real-world restrictions and offline evaluation, and recommend using extra test datasets to conduct offline policy evaluation, which leads to a great challenge for existing offline RL methods.

\textbf{Future work.}
In the future, we will step further towards real-world scenarios and investigate more real-world offline RL challenges, by constantly providing new near real-world datasets and tasks. We also hope the NeoRL benchmark will shed some light on future research and draw more attention to real-world RL applications.

\section*{Acknowledgments}

This work is supported by National Key R\&D Program of China (2018AAA0101100), NSFC (61876077), and Collaborative Innovation Center of Novel Software Technology and Industrialization.

\bibliographystyle{named}
\bibliography{offRL_benchmark21}

\newpage
\appendix
\section{Task Description}
\label{task_description}

\begin{table}[!h]
	\centering
	\caption{Configuration of environments.}
	\label{task_configuration}
	\scalebox{0.99}{\begin{tabular}{ccccc}
			\toprule
			Environment & \begin{tabular}[c]{@{}c@{}}Observation \\ Shape\end{tabular} & \begin{tabular}[c]{@{}c@{}}Action \\ Shape\end{tabular} & \begin{tabular}[c]{@{}c@{}}Have \\ Done\end{tabular} & \begin{tabular}[c]{@{}c@{}}Max \\ Timesteps\end{tabular}\\
			\midrule
			HalfCheetah-v3 & 18 & 6 & False & 1000\\
			Hopper-v3 & 12 & 3 & True & 1000\\
			Walker2d-v3 & 18 & 6 & True & 1000\\
			IB & 182 & 3 & False & 1000\\
			FinRL & 181 & 30 & False & 2516\\
			CL & 74 & 14 & False & 1000\\
			\bottomrule
	\end{tabular}}
\end{table}

\textbf{MuJoCo}  We set EXCLUDE\_CURRENT\_POSITIONS\_FROM\_OBSERVATION  to false to include the first dimension of the position in HalfCheetah-v3, Walker2d-v3, and Hopper-v3. We use Gym-like MuJoCo: \url{https://gym.openai.com/envs/#mujoco}

\textbf{IB}: IB simulates the characteristics presented in various industrial control tasks, such as wind or gas turbines, chemical reactors, etc. The raw system output for each time step is a 6-dimensional vector including velocity, gain, shift, setpoint, consumption, and fatigue. To enhance the Markov property, the authors stitch the system outputs of the last $K$ timesteps as observations ($K=30$ by default). We added two-dimensional dummy system information into the observation, which records the instant system reward information. The action space is three-dimensional. Each action can be interpreted as three proposed changes to the three observable state variables called current steerings.
Original codes can be found at: \url{https://github.com/siemens/industrialbenchmark}.

\textbf{FinRL} FinRL contains 30 stocks in the pool and the trading histories over the past 10 years. Each stock is represented as a 6-dimensional feature vector, where one dimension is the number of stocks currently owned, another five dimensions are the factor information of that stock. The observation has one dimension of information representing the current account cash balance. The dimension of the action space is 30, corresponding to the transactions of each of the thirty stocks. Original codes can be found at: \url{https://github.com/AI4Finance-LLC/FinRL-Library}.

\textbf{CityLearn} The CityLearn (CL) environment \cite{canteli2019citylearn} reshapes the aggregation curve of electricity demand by controlling energy storage in different types of buildings. Domestic hot water (DHW) and solar power demands are modeled in the CL environment. High electricity demand raises the price of electricity and the overall cost of the distribution network. Flattening, smoothing, and narrowing the electricity demand curve help to reduce the operating and capital costs of generation, transmission and distribution. The observation encodes the states of buildings, including time, outdoor temperature, indoor temperature, humidity, solar radiation, power consumption, charging status of the cooling and heating storage units, etc. The action is to control each building to increase or decrease the amount of energy stored in its own heat storage and cooling equipment.
Original codes can be found at: \url{https://github.com/intelligent-environments-lab/CityLearn}.

The configuration of environments is summarized in Table \@\ref{task_configuration}.

\section{Re-implementation verification}
\label{verify_reimplementation}
We have verified our re-implementations on D4RL MuJoCo-medium tasks. The hyper-parameters are set to the recommended values in original papers. The results are shown in Table.\@\ref{reimplementation_results}.

\begin{table}[!h]
	\centering
	\caption{D4RL results of the re-implementations. Values in the brackets stand for the reported values in the original papers. The bold number means the result is significantly different from the original paper.}
	\label{reimplementation_results}
	\begin{tabular}{lcccc}
		\toprule
		Task Name & CQL & PLAS & BCQ & MOPO \\
		\midrule
		walker2d-medium & \textbf{3606.9} (2664.2) & 3256.7 (3072.4) & \textbf{3169.1} (2441.0) & \textbf{1268.2} (644.3)\\
		hopper-medium & 2527.7 (2557.3) & 1093.8 (1182.1) & \textbf{1020.0} ({1752.4}) & 692.7 (842.2) \\
		halfcheetah-medium & 4872.3 (5232.1) & 4798.2 (4964.6) & 5084.2 (4767.9) & 4602.8 (4710.7)\\
		\bottomrule
	\end{tabular}
\end{table}

\section{Computation Resources}
\label{computation_resources}
We use 64 NVIDIA Telsa V100 cards and hundreds of CPU cores to run all experiments, and use Ray to enable parallelism. It takes 12 hours to collect datasets, about one week to train all the offline policies, and 12 hours to conduct the offline policy evaluation(to be specific, FQE) for each task per seed.

\section{Choice of Hyper-parameters}
\label{hyperparameter-appendix}
To make a fair comparison, all the policies and value functions are implemented by an MLP with 2 hidden layers and 256 units per layer. The output of the policies is transformed by $\tanh$ function to ensure the actions are within the range. For model-based approaches, the transition model is represented by a local Gaussian distribution, i.e. $s_{t+1} \sim \mathcal{N}(s_t + \Delta_\theta(s_t, a_t), \sigma_\theta(s_t, a_t))$, where $\Delta_\theta$ and $\sigma_\theta$ are implemented by an MLP with 4 hidden layers and 256 units per layer, using two heads. The transition model is trained by Adam optimizer via maximum likelihood until the negative likelihood (NLL) plateaus on test dataset. 

For BC, the policies is trained by Adam optimizer with learning rate of 3e-4 for 100$K$ steps with a batch size of 256, and it is early stopped with the lowest NLL on the test dataset to prevent overfitting. Although the best policy may get from the middle of the training process, except for BC, there does not exist decent criterion to early stop. Thus, we only consider the finally trained policy for online evaluation. 

For MB-PPO, the policy is trained for 5$K$ gradient steps by Adam optimizer with a learning rate of 3e-4. For other methods, we treat 1$K$ steps as an epoch, and then train PLAS, BCQ, MOPO for 200 epochs and train CQL for 300 epochs (The original CQL used 3000 epochs, but it spends too much time).

CQL, PLAS, BCQ and MOPO can be very sensitive to the choice of hyper-parameters. To evaluate the performance of these algorithms, we conduct grid searches for the important hyper-parameters. The search space of these algorithms are summarized in Table.\@ \ref{hyper-parameters_search_space} and the hyper-parameters used in the reported results is summarized in Table.\@ \ref{hyperparameter}. For parameters not mentioned, their values are set as the same of the original paper.

\begin{table}[!h]
	\centering
	\caption{The search space of hyper-parameters.}
	\label{hyper-parameters_search_space}
	\scalebox{0.99}{\begin{tabular}{cc}
		\toprule
		Algorithms & Search Space\\
		\midrule
		CQL & \begin{tabular}[c]{@{}c@{}}variant $\in$ $\{\mathcal{H}, \rho\}$\\$\alpha \in \{5, 10\}$\\$\tau \in \{-1, 2, 5, 10\}$\\approximate-max backup $\in$ \{True, False\}\end{tabular}\\
		\midrule
		BCQ & \begin{tabular}[c]{@{}c@{}}$\Phi \in \{0.05, 0.1, 0.2, 0.5\}$\end{tabular}\\
		\midrule
		PLAS & \begin{tabular}[c]{@{}c@{}}perturbation $\in$ \{True, False\} \\$\Phi \in \{0.05, 0.1, 0.2, 0.5\}$\end{tabular}\\
		\midrule
		MOPO & \begin{tabular}[c]{@{}c@{}}uncertainty type $\in$ \{aleatoric, disagreement\}\\$h \in \{1, 5\}$\\$\lambda \in \{0.5, 1, 2, 5\}$\end{tabular}\\
		\bottomrule
	\end{tabular}}
\end{table}
\begin{table*}
\centering
\caption{Hyper-parameters for reported results.}
\label{hyperparameter}
\scalebox{0.7}{\begin{tabular}{|l|c|c|c|c|c|c|c|c|c|c|}
\hline
\multirow{2}{*}{Task Name} & \multicolumn{4}{|c|}{CQL} &\multicolumn{1}{|c|}{BCQ} & \multicolumn{2}{|c|}{PLAS} & \multicolumn{3}{|c|}{MOPO} \\ \cline{2-11} &Variant & $\alpha$ & $\tau$ & \begin{tabular}[c]{@{}c@{}}Approximate-max \\ backup\end{tabular} & $\Phi$ & Perturbation & $\Phi$ & Uncertainty type & $h$ & $\lambda$ \\
\hline
HalfCheetah-v3-L-99 & $\mathcal{H}$ & - & 2 & False & 0.05 & True & 0.1 & aleatoric & 5 & 0.5\\
HalfCheetah-v3-L-999 & $\mathcal{H}$ & - & 2 & False & 0.05 & True & 0.1 & aleatoric & 5 & 2\\
HalfCheetah-v3-L-9999 & $\mathcal{H}$ & - & 10 & False & 0.1 & True & 0.05 & aleatoric & 5 & 2\\
HalfCheetah-v3-M-99 & $\rho$ & 5 & -1 & False & 0.05 & False & - & aleatoric & 5 & 2\\
HalfCheetah-v3-M-999 & $\rho$ & 5 & -1 & False & 0.05 & False & - & aleatoric & 5 & 1\\
HalfCheetah-v3-M-9999 & $\rho$ & - & 10 & False & 0.05 & False & - & aleatoric & 5 & 0.5\\
HalfCheetah-v3-H-99 & $\rho$ & - & 10 & True & 0.1 & False & - & disagreement & 5 & 5\\
HalfCheetah-v3-H-999 & $\rho$ & - & 10 & False & 0.05 & False & - & aleatoric & 5 & 1\\
HalfCheetah-v3-H-9999 & $\mathcal{H}$ & 5 & -1 & False & 0.2 & False & - & aleatoric & 1 & 1\\
\hline
Hopper-v3-L-99 & $\mathcal{H}$ & - & 10 & False & 0.05 & True & 0.1 & aleatoric & 5 & 1\\
Hopper-v3-L-999 & $\mathcal{H}$ & - & 5 & False & 0.2 & True & 0.5 & aleatoric & 5 & 1\\
Hopper-v3-L-9999 & $\mathcal{H}$ & 5 & -1 & False & 0.2 & True & 0.2 & aleatoric & 5 & 1\\
Hopper-v3-M-99 & $\rho$ & - & 5 & False & 0.1 & False & - & aleatoric & 5 & 2\\
Hopper-v3-M-999 & $\mathcal{H}$ & 5 & -1 & False & 0.1 & False & - & aleatoric & 5 & 1\\
Hopper-v3-M-9999 & $\mathcal{H}$ & - & 10 & False & 0.05 & False & - & aleatoric & 5 & 1\\
Hopper-v3-H-99 & $\rho$ & - & 10 & False & 0.1 & False & - & aleatoric & 1 & 1\\
Hopper-v3-H-999 & $\rho$ & - & 2 & False & 0.05 & False & - & disagreement & 5 & 5\\
Hopper-v3-H-9999 & $\rho$ & 5 & -1 & False & 0.05 & False & - & aleatoric & 5 & 0.5\\
\hline
Walker2d-v3-L-99 & $\mathcal{H}$ & - & 2 & True & 0.1 & True & 0.2 & aleatoric & 5 & 0.5\\
Walker2d-v3-L-999 & $\mathcal{H}$ & - & 10 & False & 0.1 & True & 0.2 & aleatoric & 1 & 5\\
Walker2d-v3-L-9999 & $\mathcal{H}$ & - & 10 & False & 0.05 & True & 0.05 & aleatoric & 1 & 5\\
Walker2d-v3-M-99 & $\mathcal{H}$ & 5 & -1 & False & 0.1 & True & 0.2 & aleatoric & 1 & 0.5\\
Walker2d-v3-M-999 & $\mathcal{H}$ & - & 5 & False & 0.1 & False & - & aleatoric & 1 & 2\\
Walker2d-v3-M-9999 & $\mathcal{H}$ & - & 5 & False & 0.05 & False & - & aleatoric & 1 & 1\\
Walker2d-v3-H-99 & $\mathcal{H}$ & - & 2 & False & 0.1 & False & - & aleatoric & 1 & 1\\
Walker2d-v3-H-999 & $\rho$ & 5 & -1 & False & 0.05 & False & - & aleatoric & 1 & 1\\
Walker2d-v3-H-9999 & $\rho$ & 5 & -1 & False & 0.2 & False & - & aleatoric & 1 & 0.5\\
\hline
IB-L-99 & $\mathcal{H}$ & 5 & -1 & False & 0.2 & False & - & aleatoric & 1 & 1\\
IB-L-999 & $\rho$ & - & 10 & False & 0.1 & False & - & aleatoric & 5 & 5\\
IB-L-9999 & $\rho$ & - & 5 & False & 0.5 & True & 0.5 & aleatoric & 1 & 2\\
IB-M-99 & $\mathcal{H}$ & - & 5 & False & 0.2 & True & 0.1 & aleatoric & 5 & 5\\
IB-M-999 & $\mathcal{H}$ & - & 10 & False & 0.5 & False & - & aleatoric & 1 & 2\\
IB-M-9999 & $\rho$ & - & 10 & False & 0.05 & False & - & aleatoric & 1 & 2\\
IB-H-99 & $\mathcal{H}$ & - & 2 & False & 0.5 & True & 0.1 & aleatoric & 5 & 5\\
IB-H-999 & $\rho$ & - & 5 & False & 0.1 & False & - & aleatoric & 5 & 1\\
IB-H-9999 & $\rho$ & - & 2 & False & 0.1 & True & 0.1 & aleatoric & 1 & 2\\
\hline
FinRL-L-99 & $\mathcal{H}$ & 5 & -1 & True & 0.5 & True & 0.2 & disagreement & 1 & 2\\
FinRL-L-999 & $\mathcal{H}$ & - & 5 & False & 0.05 & False & - & disagreement & 1 & 5\\
FinRL-M-99 & $\rho$ & - & 2 & False & 0.1 & True & 0.2 & aleatoric & 1 & 5\\
FinRL-M-999 & $\mathcal{H}$ & - & 2 & True & 0.5 & False & - & aleatoric & 5 & 1\\
FinRL-H-99 & $\mathcal{H}$ & - & 5 & False & 0.5 & False & - & aleatoric & 5 & 5\\
FinRL-H-999 & $\mathcal{H}$ & - & 5 & False & 0.2 & False & - & aleatoric & 5 & 5\\
\hline
CL-L-99 & $\rho$ & - & 5 & False & 0.2 & False & - & disagreement & 1 & 1\\
CL-L-999 & $\mathcal{H}$ & - & 2 & False & 0.05 & True & 0.2 & aleatoric & 5 & 2\\
CL-L-9999 & $\mathcal{H}$ & - & 2 & False & 0.5 & False & - & aleatoric & 5 & 5\\
CL-M-99 & $\mathcal{H}$ & - & 2 & False & 0.05 & False & - & aleatoric & 1 & 5\\
CL-M-999 & $\mathcal{H}$ & - & 5 & False & 0.5 & True & 0.1 & disagreement & 1 & 5\\
CL-M-9999 & $\mathcal{H}$ & - & 10 & False & 0.2 & False & - & disagreement & 1 & 5\\
CL-H-99 & $\mathcal{H}$ & - & 5 & True & 0.1 & False & - & disagreement & 1 & 1\\
CL-H-999 & $\mathcal{H}$ & - & 5 & True & 0.2 & True & 0.05 & aleatoric & 1 & 5\\
CL-H-9999 & $\mathcal{H}$ & - & 5 & False & 0.05 & True & 0.1 & aleatoric & 5 & 0.5\\
\hline
\end{tabular}
}
\end{table*}

For CQL, we mainly consider 4 parameters mentioned in the original paper:
\begin{itemize}
	\item Variant: The paper proposed two variants of CQL algorithms, i.e. CQL($\mathcal{H}$) and CQL($\rho$). The former one uses entropy as the regularizer, whereas the later one uses KL-divergence as the regularizer.
	\item Q-values penalty parameter $\alpha$: In the formulation of CQL, $\alpha$ stands for how much penalty will be enforced on the Q function. As suggested in the paper, we search for $\alpha \in \{5, 10\}$.
	\item $\tau$: Since $\alpha$ can be hard to tune, the authors also introduce an auto-tuning trick via dual gradient-descent. The trick introduces a threshold $\tau > 0$. When the difference between Q-values are greater than $\tau$, $\alpha$ will be auto-tuned to a greater value to make the penalty more aggressive. As suggested by the paper, we search $\tau \in \{-1, 2, 5, 10\}$. $\tau = -1$ indicates the trick is disable.
	\item Approximate-max backup: By default, the bellman backup is computed with double-Q, i.e. $y = r + \min_{i=1,2}Q_i(s^\prime, a^\prime)$, where $a^\prime \sim \pi(s^\prime)$. In addition, the authors propose a approximate-max backup, which use 10 samples to approximate the max Q-values, where the backup is computed by $y = r + \min_{i=1,2}\max_{a^\prime_1 ... a^\prime_{10} \sim \pi(s^\prime)}Q_i(s^\prime, a^\prime)$.
\end{itemize}

In BCQ, the action is decoded from VAE plus a perturbation, i.e. $a = \hat{a} + \Phi \tanh(\xi_\phi(s, \hat{a}))$. Here, $\Phi$ controls the maximum deviation allowed for the learned policy from the behavior policy. We search for $\Phi \in \{0.05, 0.1, 0.2, 0.5\}$.

For PLAS, the default setting is only learning a deterministic policy in the latent space of VAE. The authors mentioned that a similar perturbation layer as BCQ can be applied on the output action to improve its generalization out of the dataset. Thus, we search for whether the perturbation layer is applied, and if applied, we search for the value of $\Phi \in \{0.05, 0.1, 0.2, 0.5\}$. 

For MOPO, we consider 3 parameters mentioned in the original paper:
\begin{itemize}
	\item Uncertainty type: In the default setting, MOPO uses the max L2-norm of the output standard deviation among ensemble transition models, i.e. $\max_{i=1...N}\|\sigma_\theta^i(s, a)\|_2^2$, as the uncertainty term. Since the learned variance can theoretically recover the true aleatoric uncertainty \cite{MOPO}, we denote this type of uncertainty as aleatoric. We also include another variant that uses the disagreement between ensemble transition models, i.e. $\max_{i=1...N}\|\Delta_\theta^i(s, a) - \frac{1}{N}\sum_i \Delta_\theta^i(s, a)\|_2^2$, as the uncertainty term. We refer to this variant as disagreement.
	\item Rollout horizon $h$: MOPO uses a branch rollout trick that run rollout from states in the dataset with small length. $h$ determines the length of the rollout. As suggested in the paper, we search for $h \in \{1, 5\}$.
	\item Uncertainty penalty weight $\lambda$: The main idea of MOPO is to penalize the reward function with the uncertainty term, i.e. $\hat{r} = r - \lambda u(s, a)$. Here, $\lambda$ control the amplitude of the penalty. As suggested in the original paper, we search for $\lambda \in \{0.5, 1, 2, 5\}$.
\end{itemize}

\section{Details of Offline Policy Evaluation}
We use FQE to conduct offline policy evaluation. The hyper-parameters are set as the default parameters suggested by \cite{useFQE}. In the experiment, we observe that FQE is inclined to explode to large values. Therefore, we use a value clipping trick on the target of bellman backups. The max and min values are computed by the rewards in the validation dataset with $40\%$ enlargement of the interval. That is, $v_\text{max} = (1.2 r_\text{max} - 0.2 r_\text{min}) / (1 - \gamma)$ and $v_\text{min} = (1.2 r_\text{min} - 0.2 r_\text{max}) / (1 - \gamma)$.

\begin{table}
\centering
\caption{FQE performance on the whole set of policies. L, M, H stands for low, medium and high quality of dataset. Each item is the mean and std over 3 random seeds of FQE runs except that the average item is the mean and std of 27 scores.}
\label{fqe_full}
\scalebox{0.65}{\begin{tabular}{lrccccccc}
\toprule
Task Name & \begin{tabular}[c]{@{}c@{}}RC \\ Score\end{tabular} & \begin{tabular}[c]{@{}c@{}}Top-1 \\ Mean \\ Score\end{tabular} & \begin{tabular}[c]{@{}c@{}}Top-3 \\ Mean \\ Score\end{tabular} & \begin{tabular}[c]{@{}c@{}}Top-5 \\ Mean \\ Score\end{tabular} & \begin{tabular}[c]{@{}c@{}}Top-1 \\ Max \\ Score\end{tabular} & \begin{tabular}[c]{@{}c@{}}Top-3 \\ Max \\ Score\end{tabular} & \begin{tabular}[c]{@{}c@{}}Top-5 \\ Max \\ Score\end{tabular} & \begin{tabular}[c]{@{}c@{}}Policy \\ Mean \\ Score\end{tabular}\\
\midrule
Walker2d-v3-L-99 & $-.237 \pm .020$ & $.057 \pm .025$ & $.176 \pm .014$ & $.168 \pm .021$ & $.057 \pm .025$ & $.366 \pm .000$ & $.375 \pm .012$ & $.324 \pm .000$ \\
Walker2d-v3-L-999 & $.005 \pm .036$ & $.020 \pm .001$ & $.044 \pm .002$ & $.047 \pm .018$ & $.020 \pm .001$ & $.093 \pm .005$ & $.136 \pm .059$ & $.362 \pm .000$ \\
Walker2d-v3-L-9999 & $.338 \pm .046$ & $.026 \pm .001$ & $.024 \pm .009$ & $.046 \pm .004$ & $.026 \pm .001$ & $.036 \pm .015$ & $.096 \pm .002$ & $.407 \pm .000$ \\
Walker2d-v3-M-99 & $-.107 \pm .061$ & $.016 \pm .005$ & $.163 \pm .105$ & $.157 \pm .005$ & $.016 \pm .005$ & $.449 \pm .303$ & $.690 \pm .037$ & $.390 \pm .000$ \\
Walker2d-v3-M-999 & $-.070 \pm .034$ & $.017 \pm .001$ & $.015 \pm .003$ & $.022 \pm .006$ & $.017 \pm .001$ & $.018 \pm .000$ & $.048 \pm .025$ & $.412 \pm .000$ \\
Walker2d-v3-M-9999 & $.101 \pm .027$ & $.011 \pm .001$ & $.018 \pm .007$ & $.019 \pm .005$ & $.011 \pm .001$ & $.031 \pm .015$ & $.042 \pm .001$ & $.416 \pm .000$ \\
Walker2d-v3-H-99 & $-.169 \pm .032$ & $.274 \pm .379$ & $.316 \pm .127$ & $.263 \pm .118$ & $.274 \pm .379$ & $.721 \pm .125$ & $.721 \pm .125$ & $.412 \pm .000$ \\
Walker2d-v3-H-999 & $-.128 \pm .058$ & $.007 \pm .002$ & $.006 \pm .001$ & $.005 \pm .001$ & $.007 \pm .002$ & $.007 \pm .002$ & $.007 \pm .001$ & $.343 \pm .000$ \\
Walker2d-v3-H-9999 & $.334 \pm .006$ & $.006 \pm .000$ & $.012 \pm .005$ & $.011 \pm .001$ & $.006 \pm .000$ & $.023 \pm .013$ & $.030 \pm .005$ & $.392 \pm .000$ \\
\midrule
Average & $.007 \pm .202$ & $.048 \pm .150$ & $.086 \pm .116$ & $.082 \pm .095$ & $.048 \pm .150$ & $.194 \pm .266$ & $.238 \pm .275$ & $.384 \pm .032$ \\
\bottomrule
\end{tabular}}
\end{table}

We show additional results figures and tables below. The scatter plots compare the estimated values against the ground truth values for every policy. The ground truth is estimated by the online performance, i.e., $v_{gt} = \frac{R_{\text{online}}}{(1 - \gamma)  h_{\text{max}}}$, where $h_{\text{max}}$ denotes the maximum horizon of the environment.

\begin{figure*}[ht]
	\centering
	\caption{FQE results on Walker2d-v3 tasks. $r$ stands for the correlation coefficient.}
	\label{Walker2d-fqe-figure}
	\begin{minipage}[h]{0.3\linewidth}
		\includegraphics[width=\linewidth]{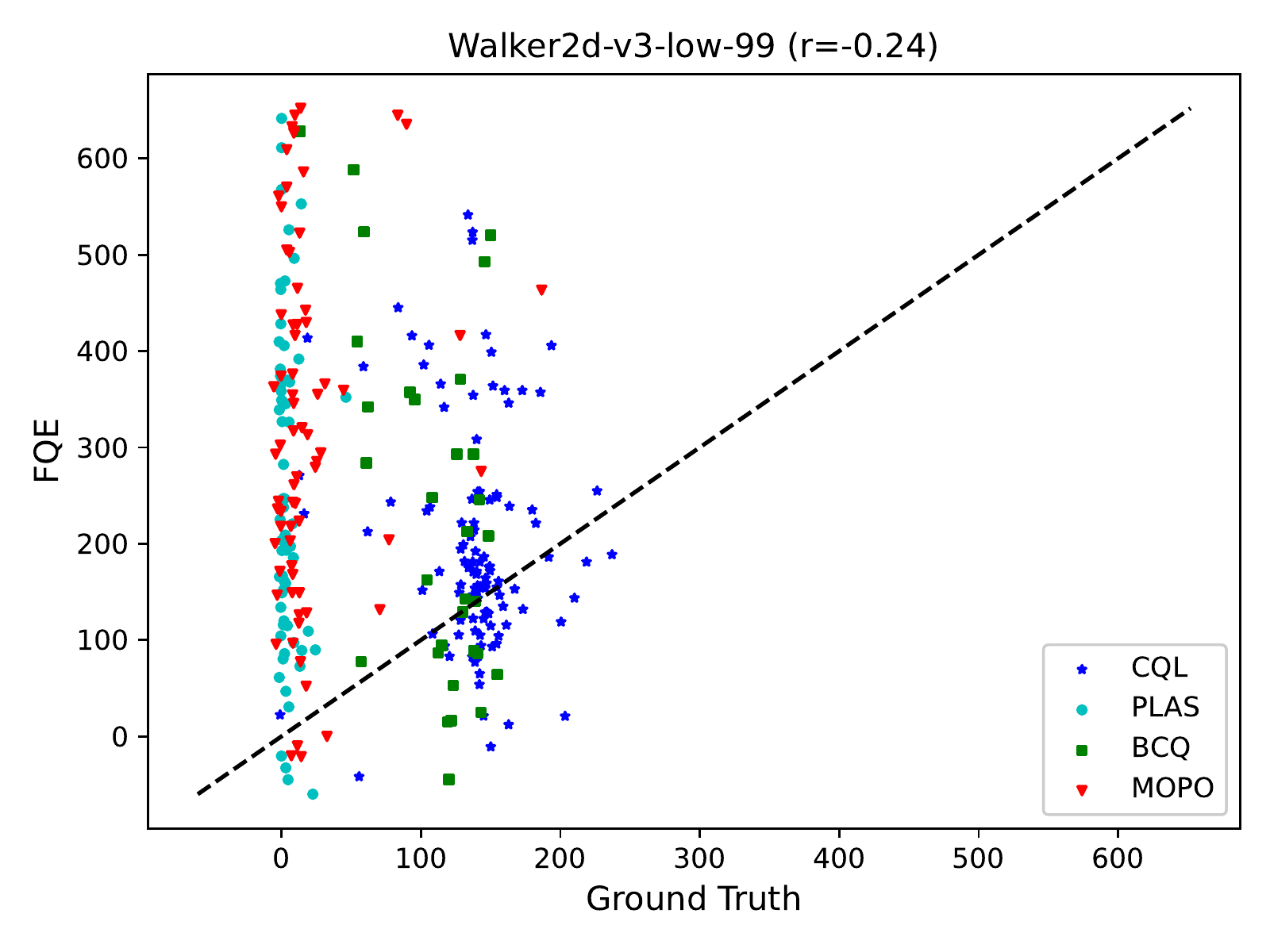}
	\end{minipage}
	\begin{minipage}[h]{0.3\linewidth}
		\includegraphics[width=\linewidth]{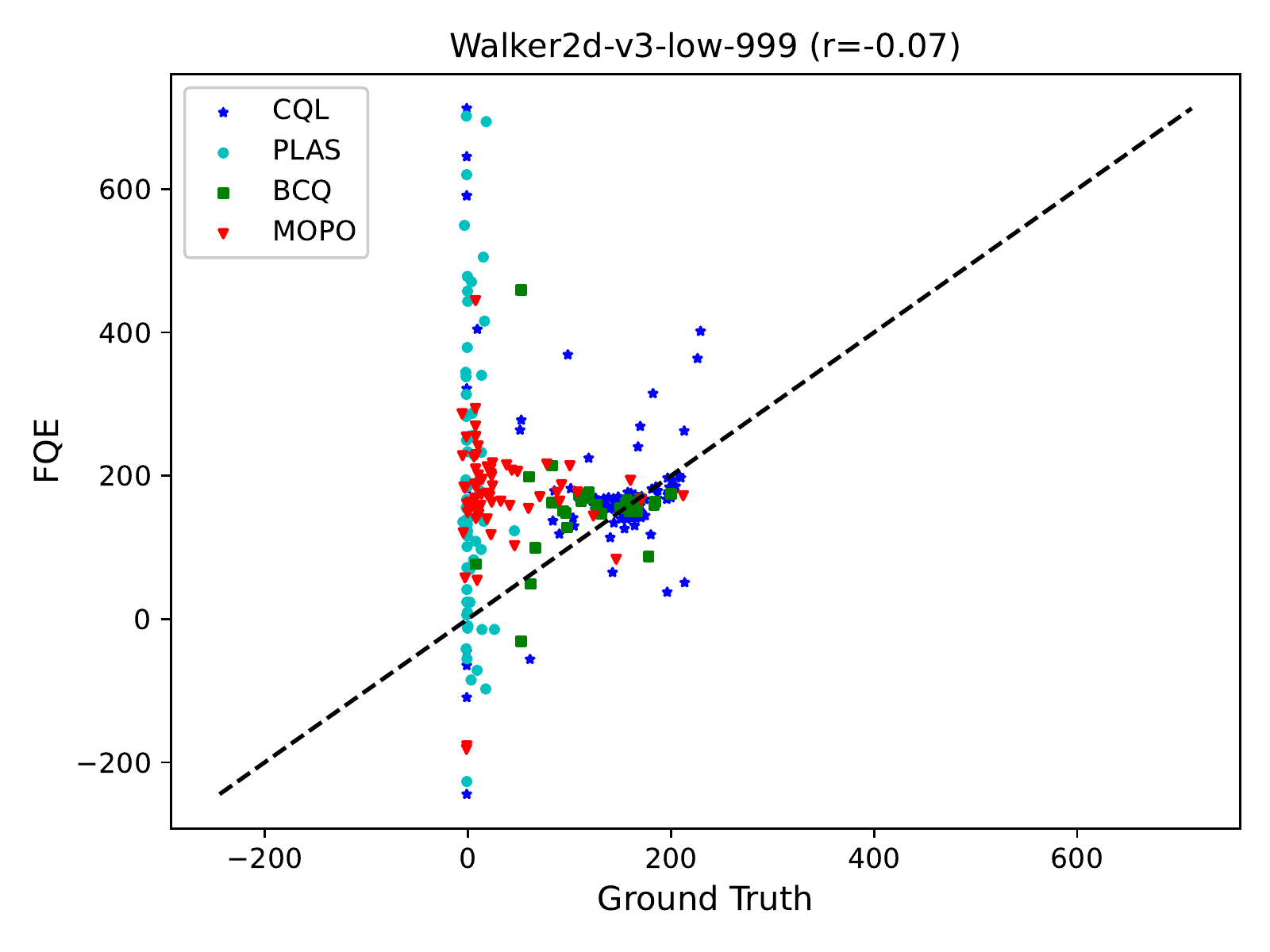}
	\end{minipage}
	\begin{minipage}[h]{0.3\linewidth}
		\includegraphics[width=\linewidth]{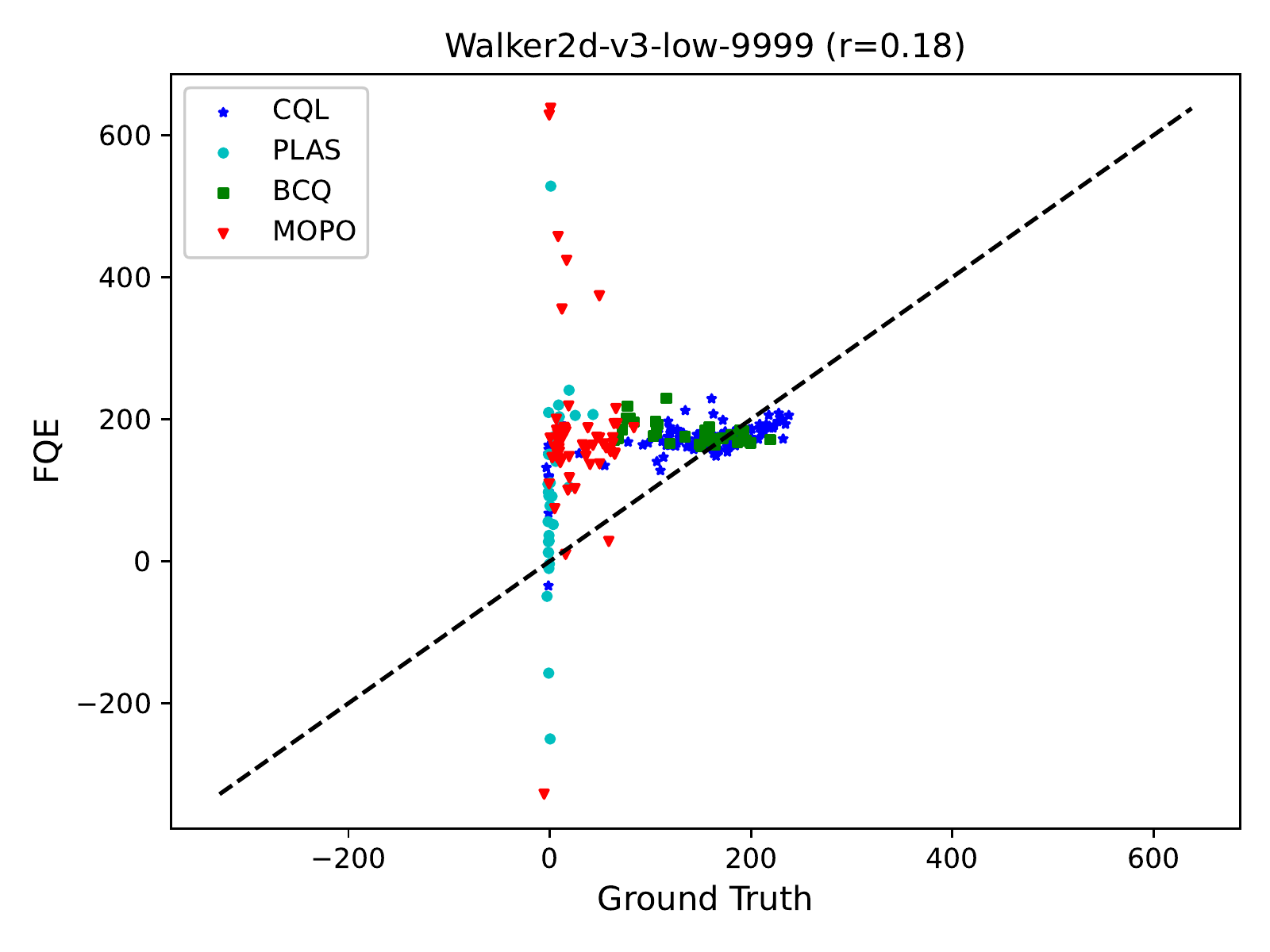}
	\end{minipage}
	
	\begin{minipage}[h]{0.3\linewidth}
		\includegraphics[width=\linewidth]{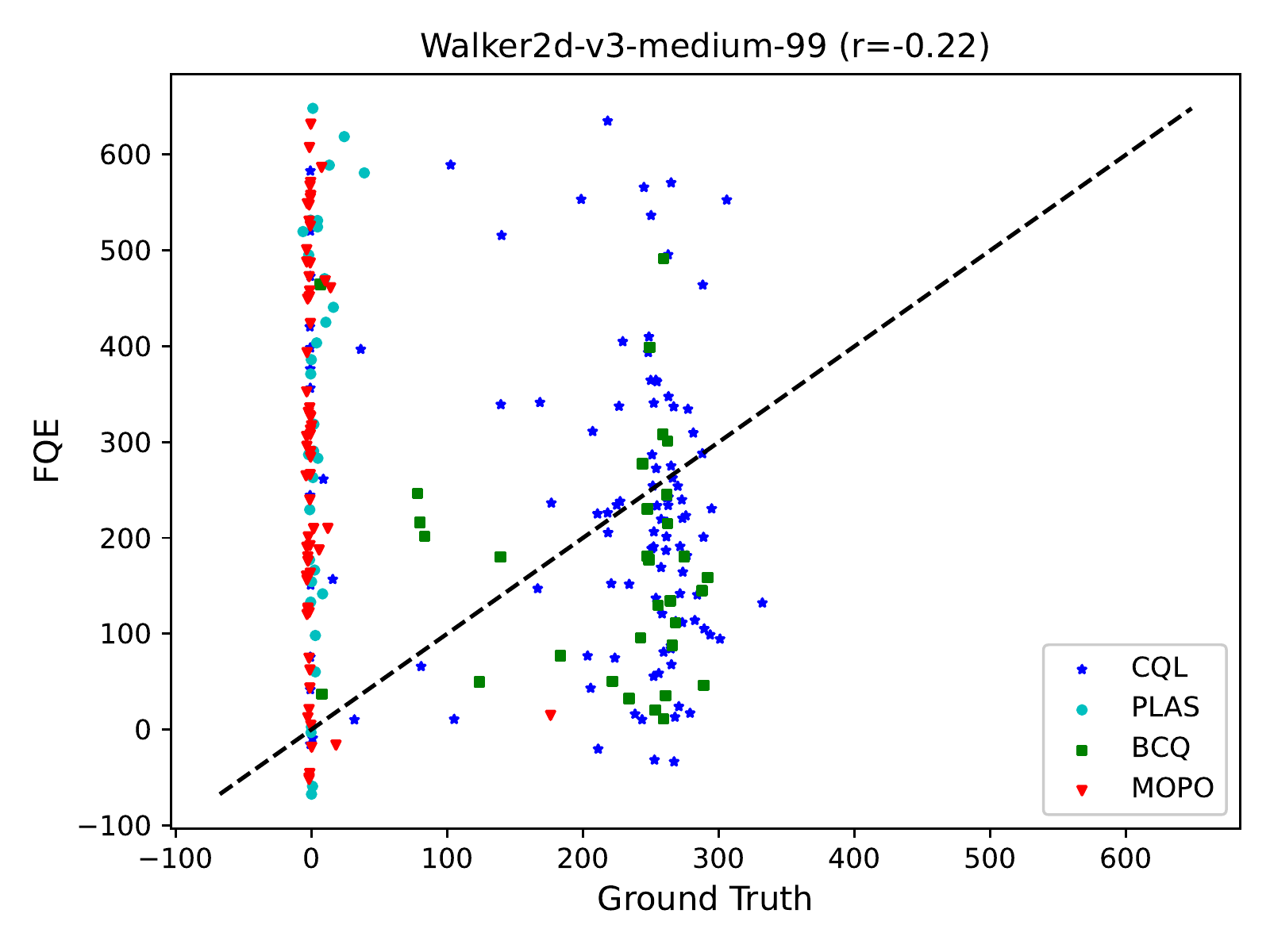}
	\end{minipage}
	\begin{minipage}[h]{0.3\linewidth}
		\includegraphics[width=\linewidth]{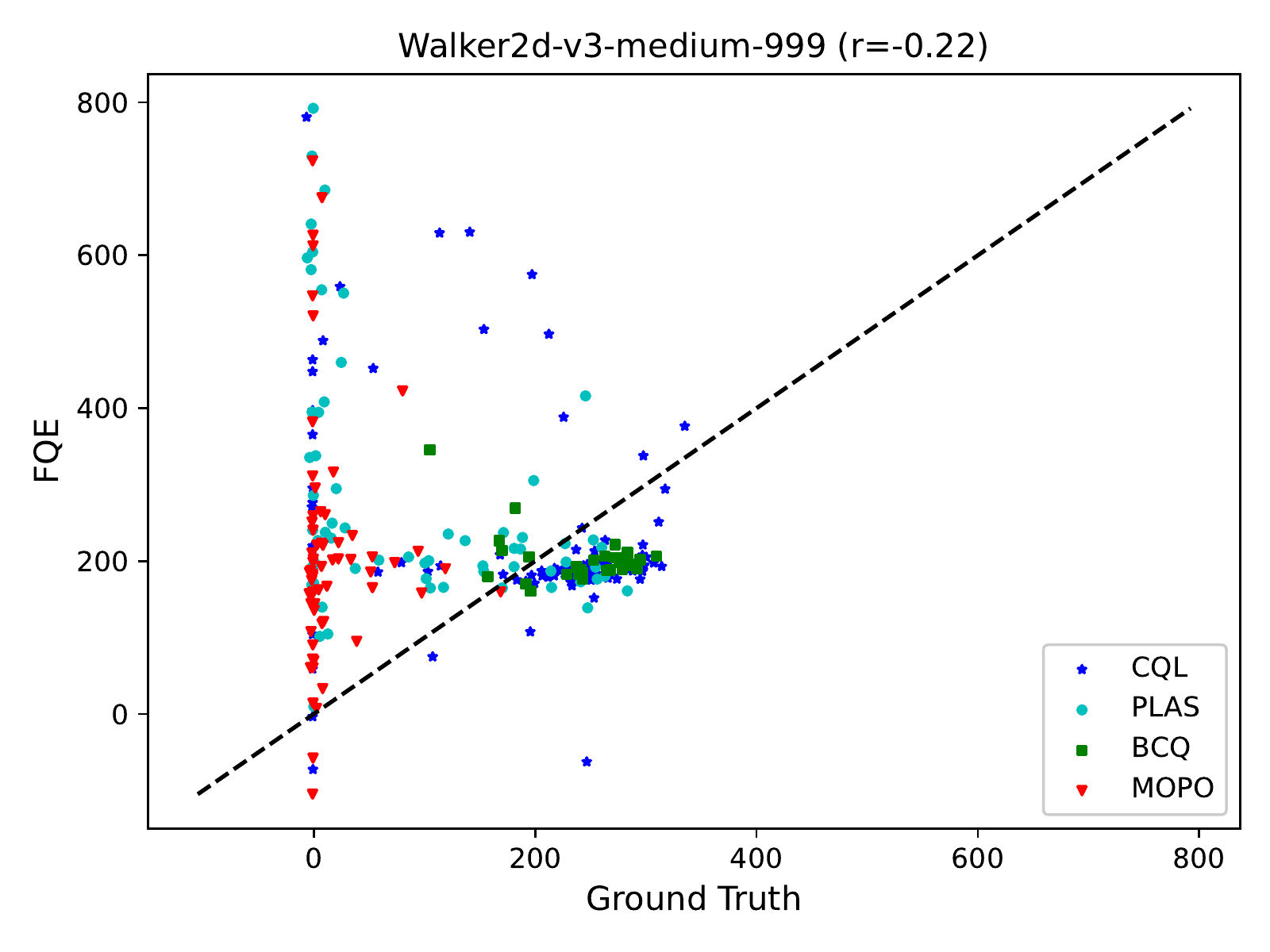}
	\end{minipage}
	\begin{minipage}[h]{0.3\linewidth}
		\includegraphics[width=\linewidth]{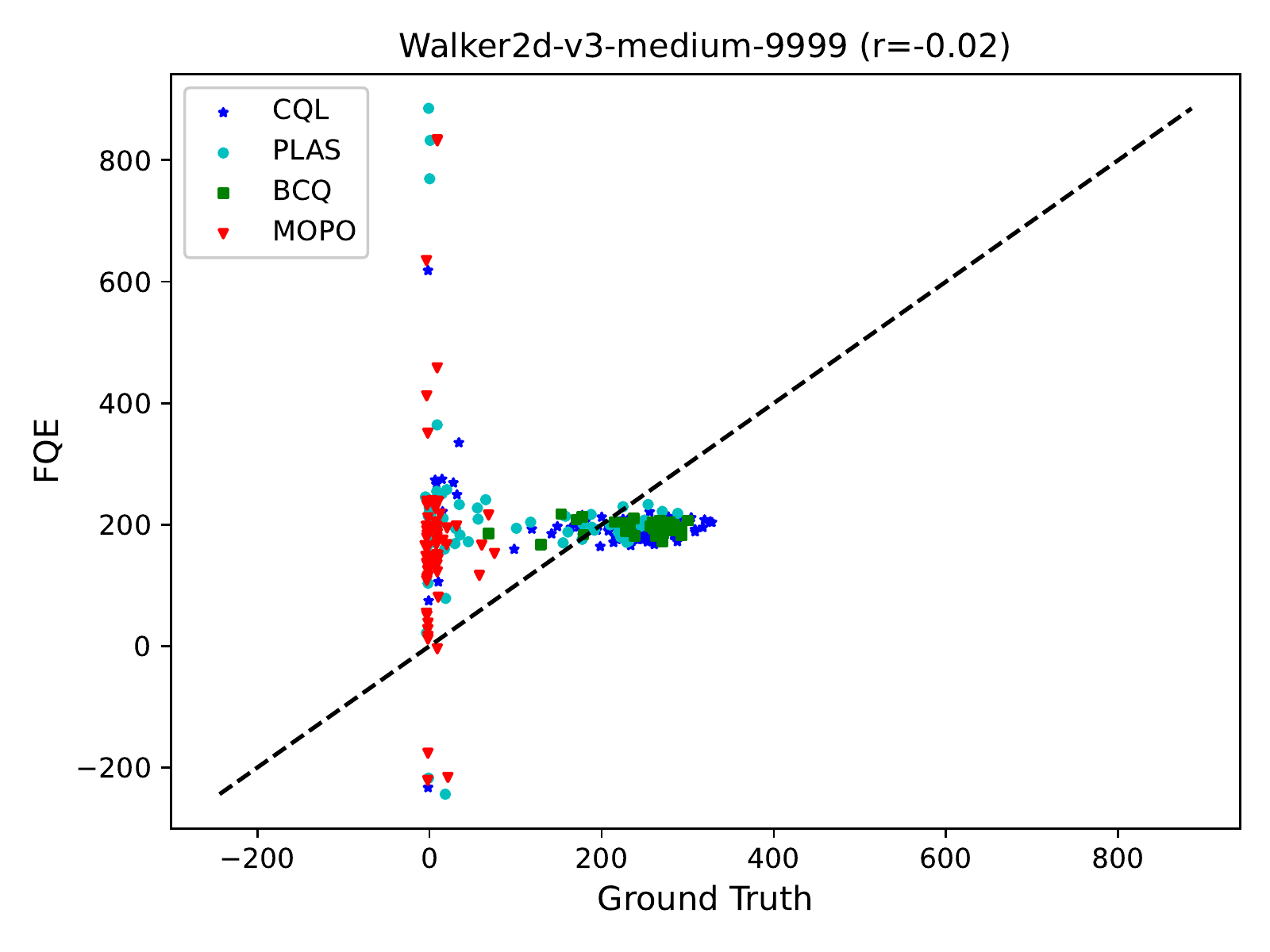}
	\end{minipage}
	
	\begin{minipage}[h]{0.3\linewidth}
		\includegraphics[width=\linewidth]{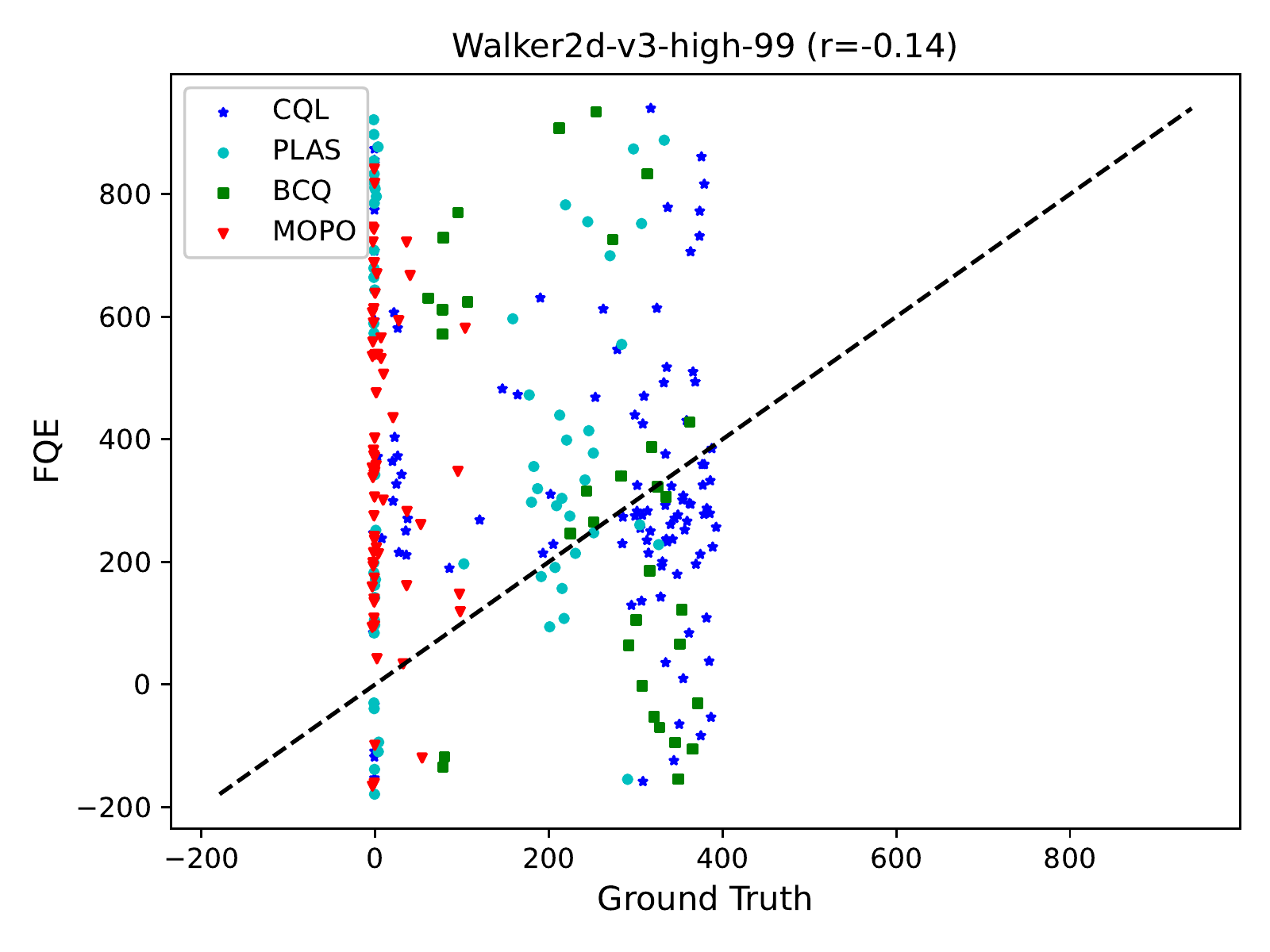}
	\end{minipage}
	\begin{minipage}[h]{0.3\linewidth}
		\includegraphics[width=\linewidth]{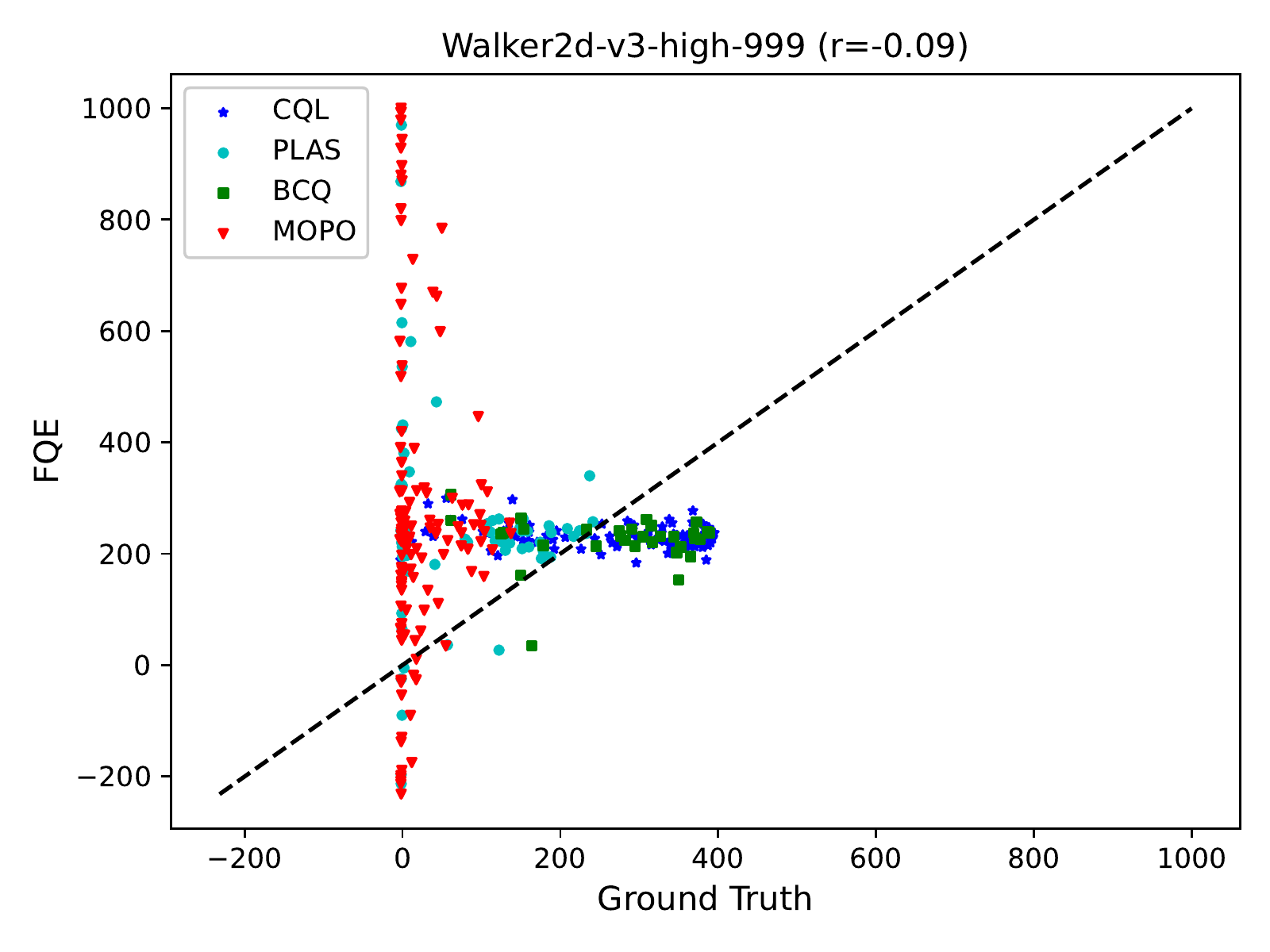}
	\end{minipage}
	\begin{minipage}[h]{0.3\linewidth}
		\includegraphics[width=\linewidth]{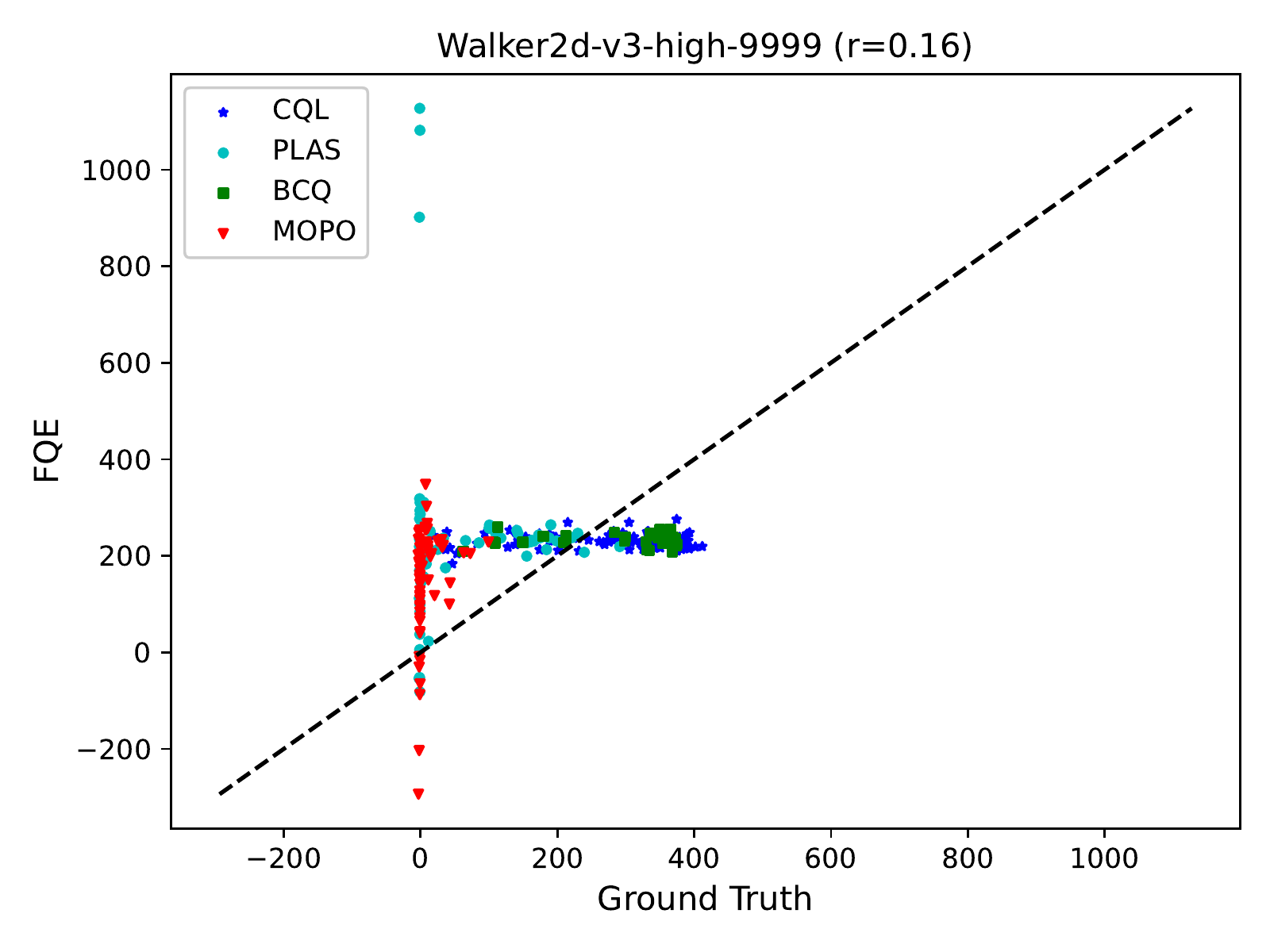}
	\end{minipage}
\end{figure*}

\begin{figure*}[ht]
	\centering
	\caption{FQE results on HalfCheetah-v3 tasks. $r$ stands for the correlation coefficient.}
	\label{HalfCheetah-fqe-figure}
	\begin{minipage}[h]{0.3\linewidth}
		\includegraphics[width=\linewidth]{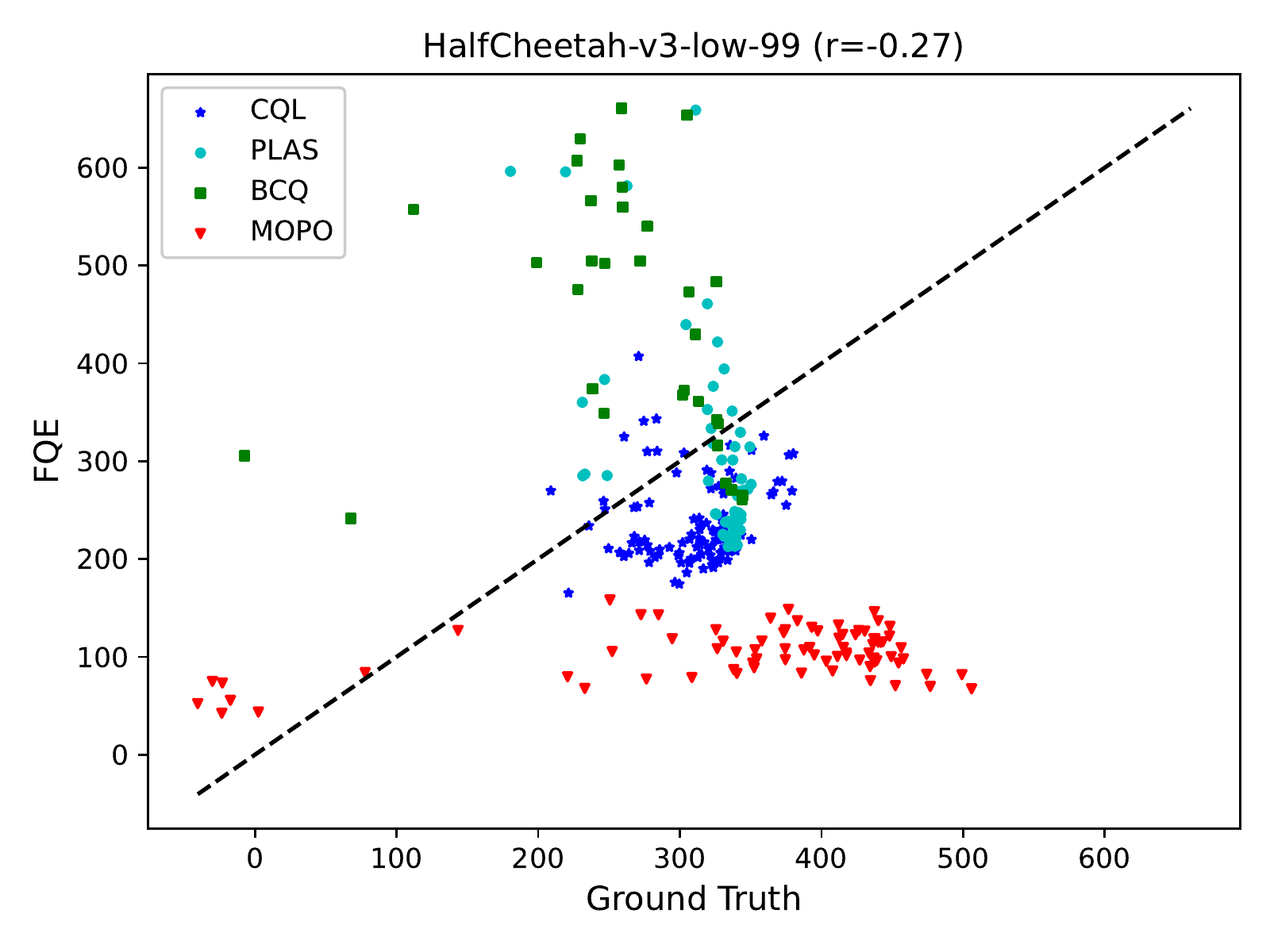}
	\end{minipage}
	\begin{minipage}[h]{0.3\linewidth}
		\includegraphics[width=\linewidth]{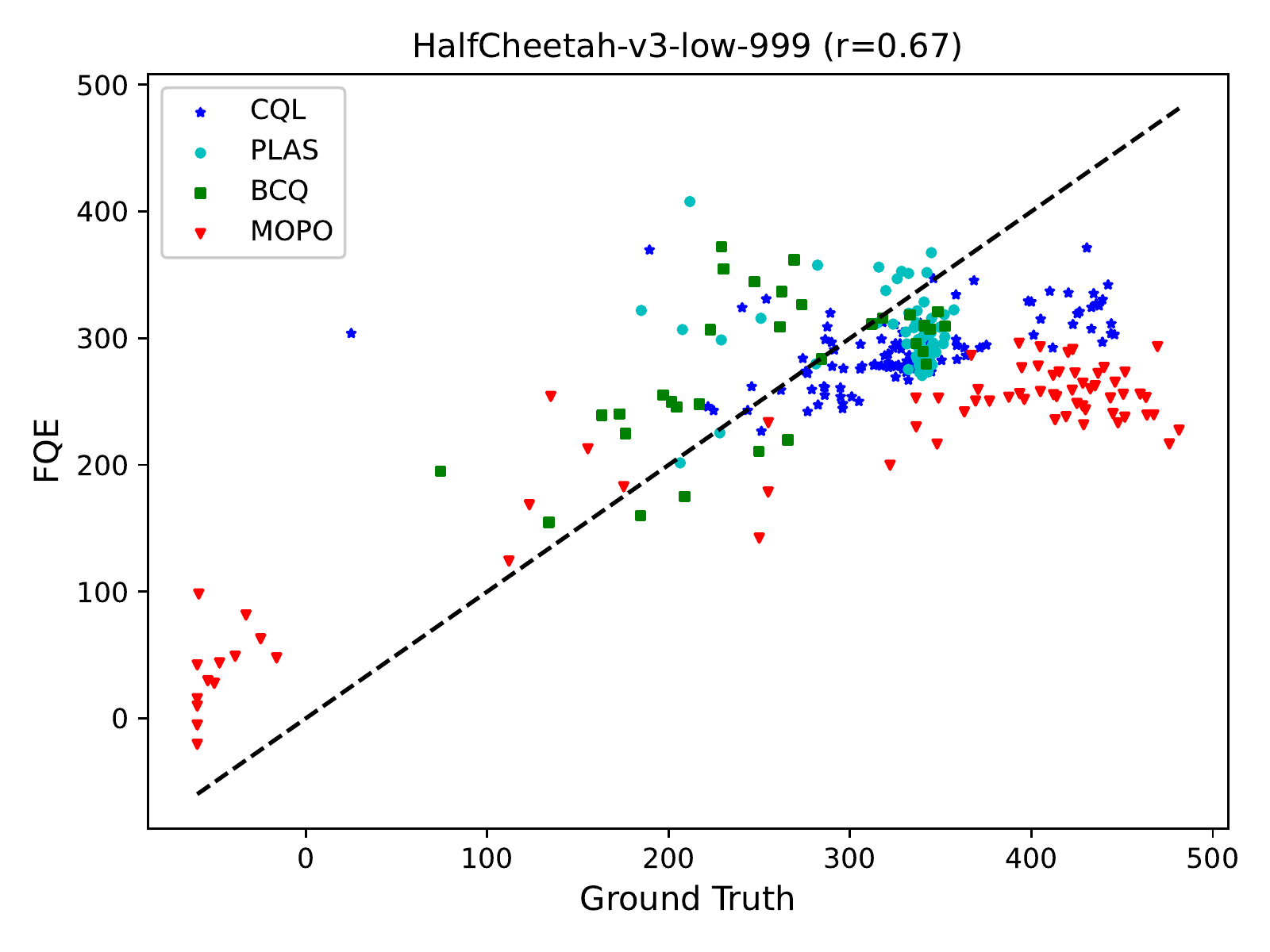}
	\end{minipage}
	\begin{minipage}[h]{0.3\linewidth}
		\includegraphics[width=\linewidth]{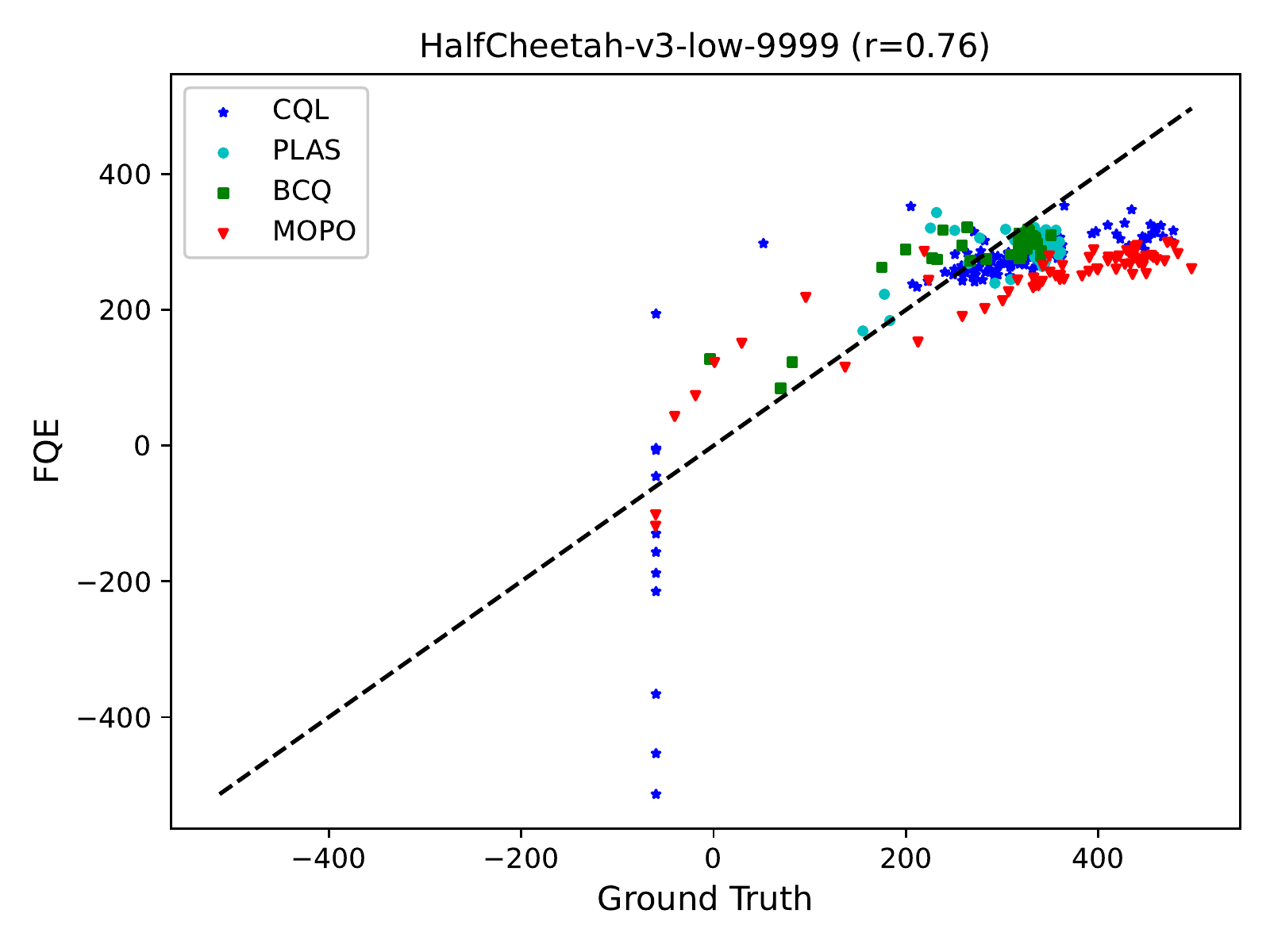}
	\end{minipage}
	
	\begin{minipage}[h]{0.3\linewidth}
		\includegraphics[width=\linewidth]{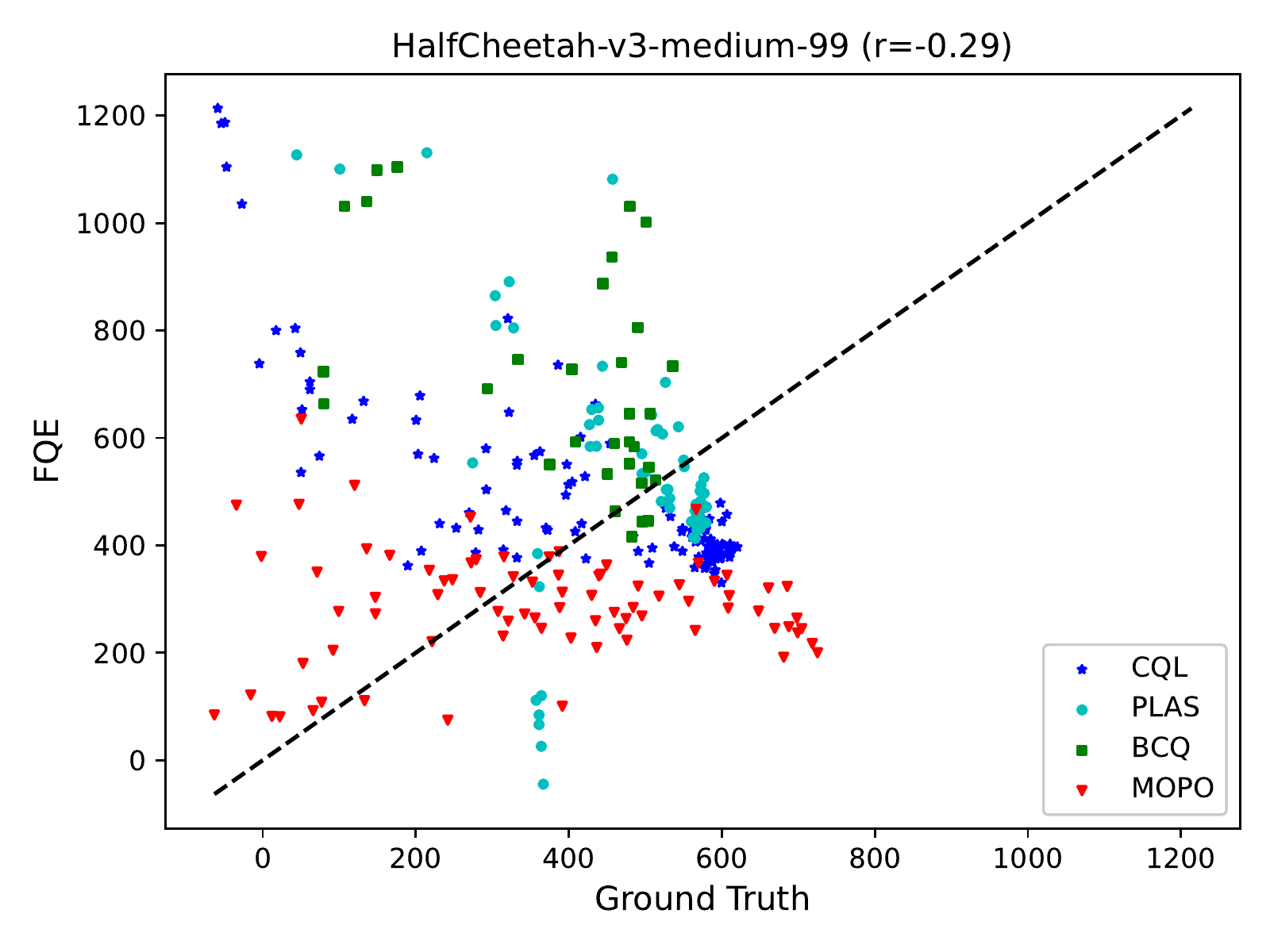}
	\end{minipage}
	\begin{minipage}[h]{0.3\linewidth}
		\includegraphics[width=\linewidth]{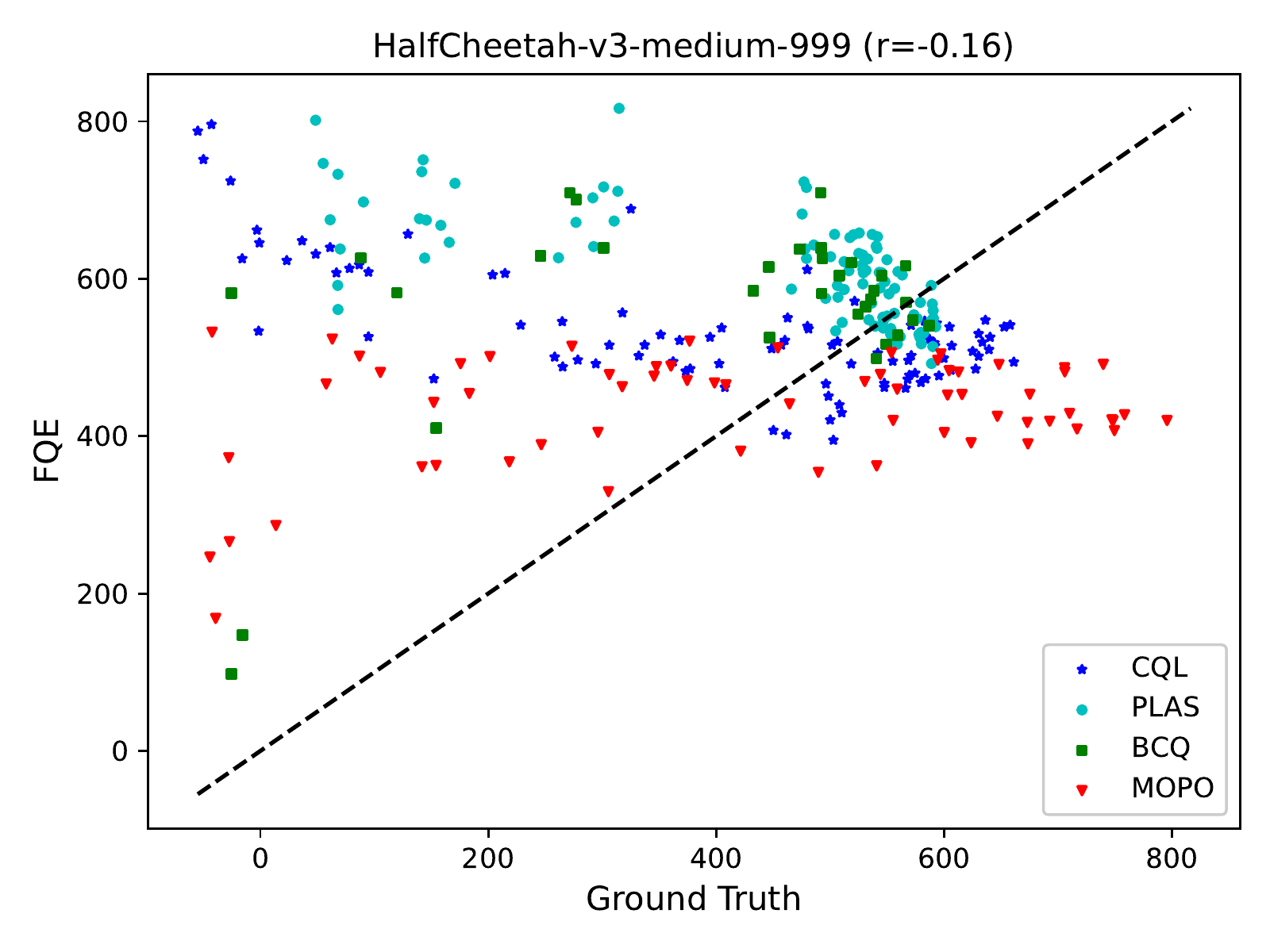}
	\end{minipage}
	\begin{minipage}[h]{0.3\linewidth}
		\includegraphics[width=\linewidth]{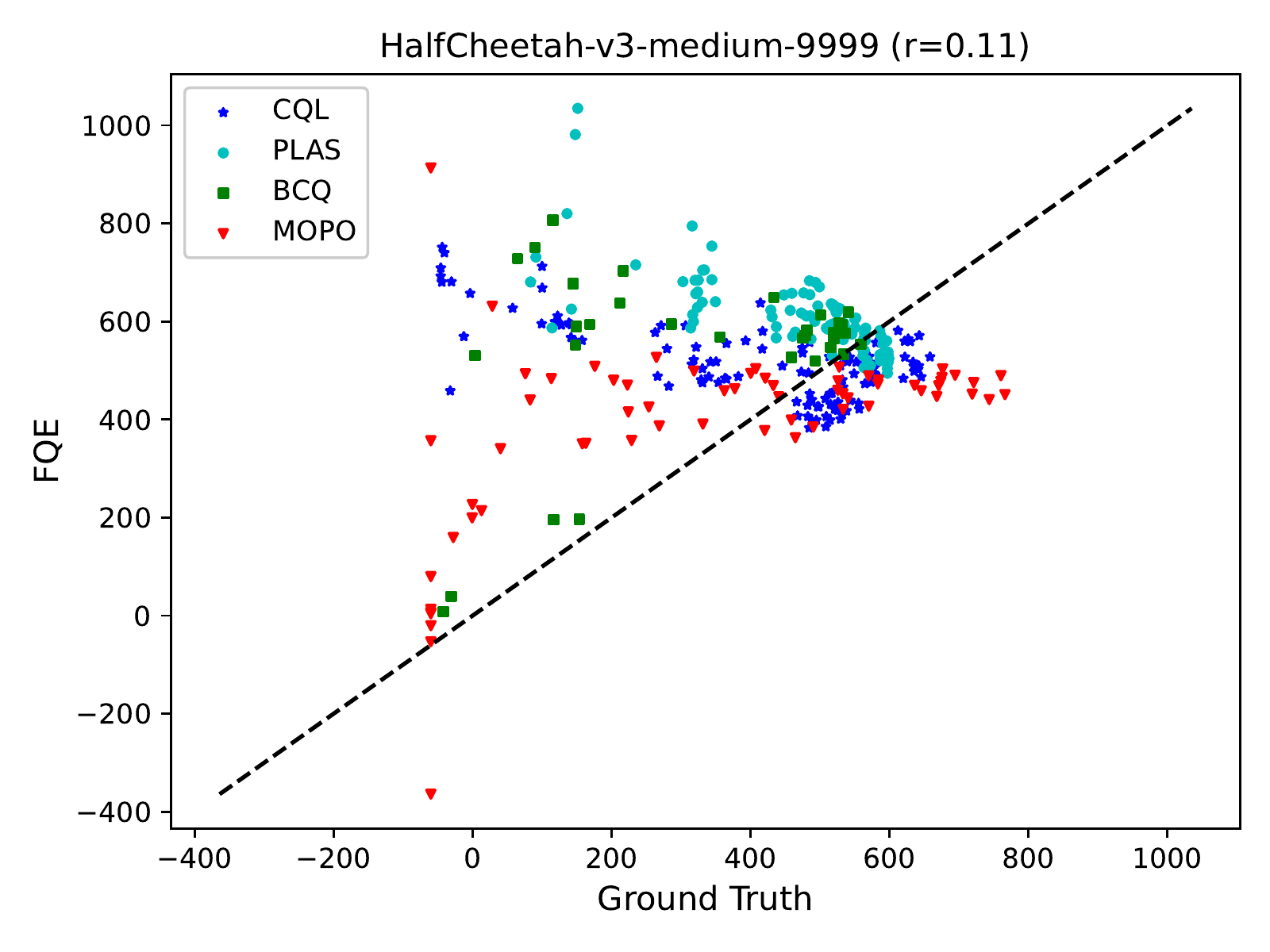}
	\end{minipage}
	
	\begin{minipage}[h]{0.3\linewidth}
		\includegraphics[width=\linewidth]{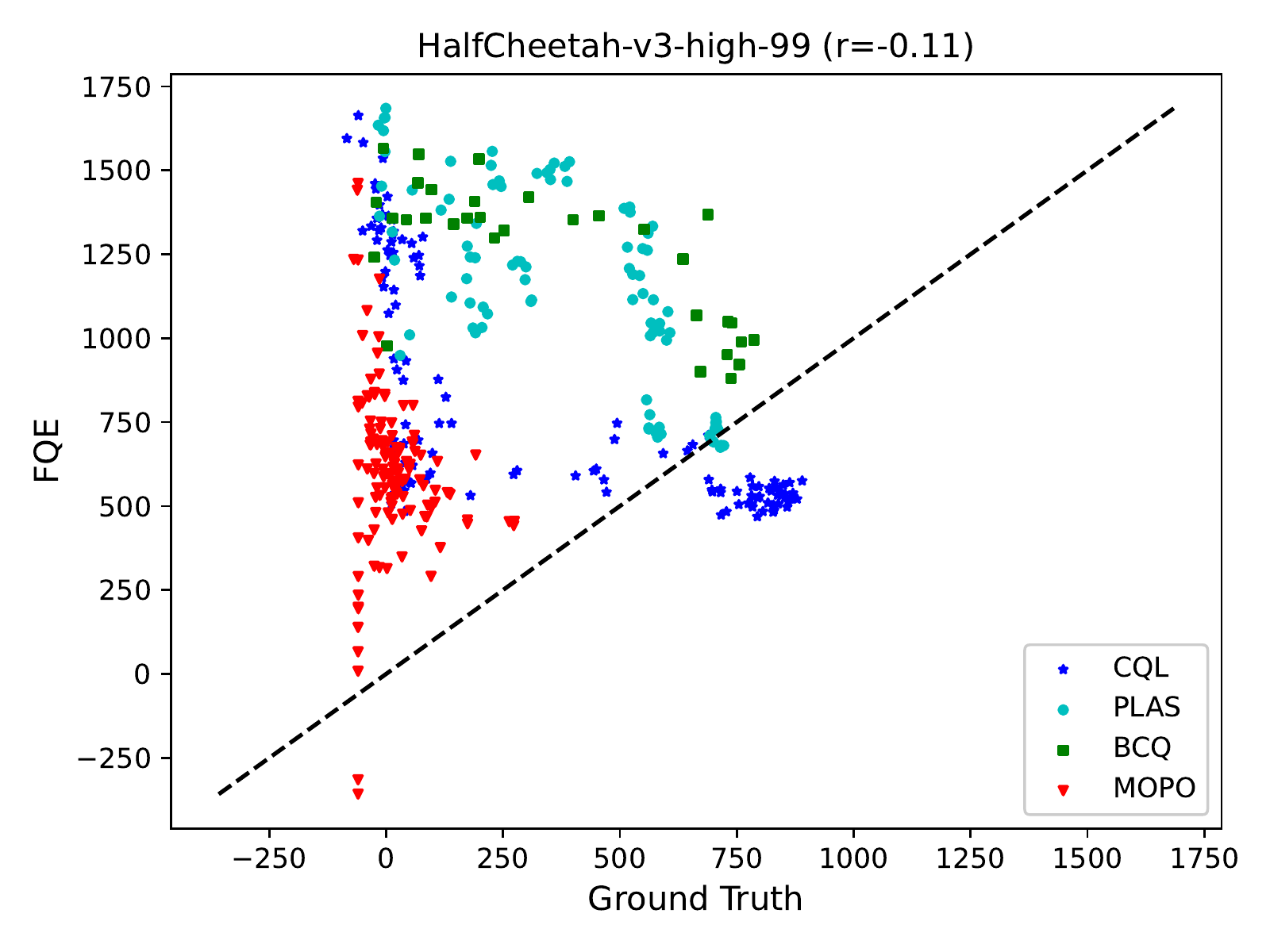}
	\end{minipage}
	\begin{minipage}[h]{0.3\linewidth}
		\includegraphics[width=\linewidth]{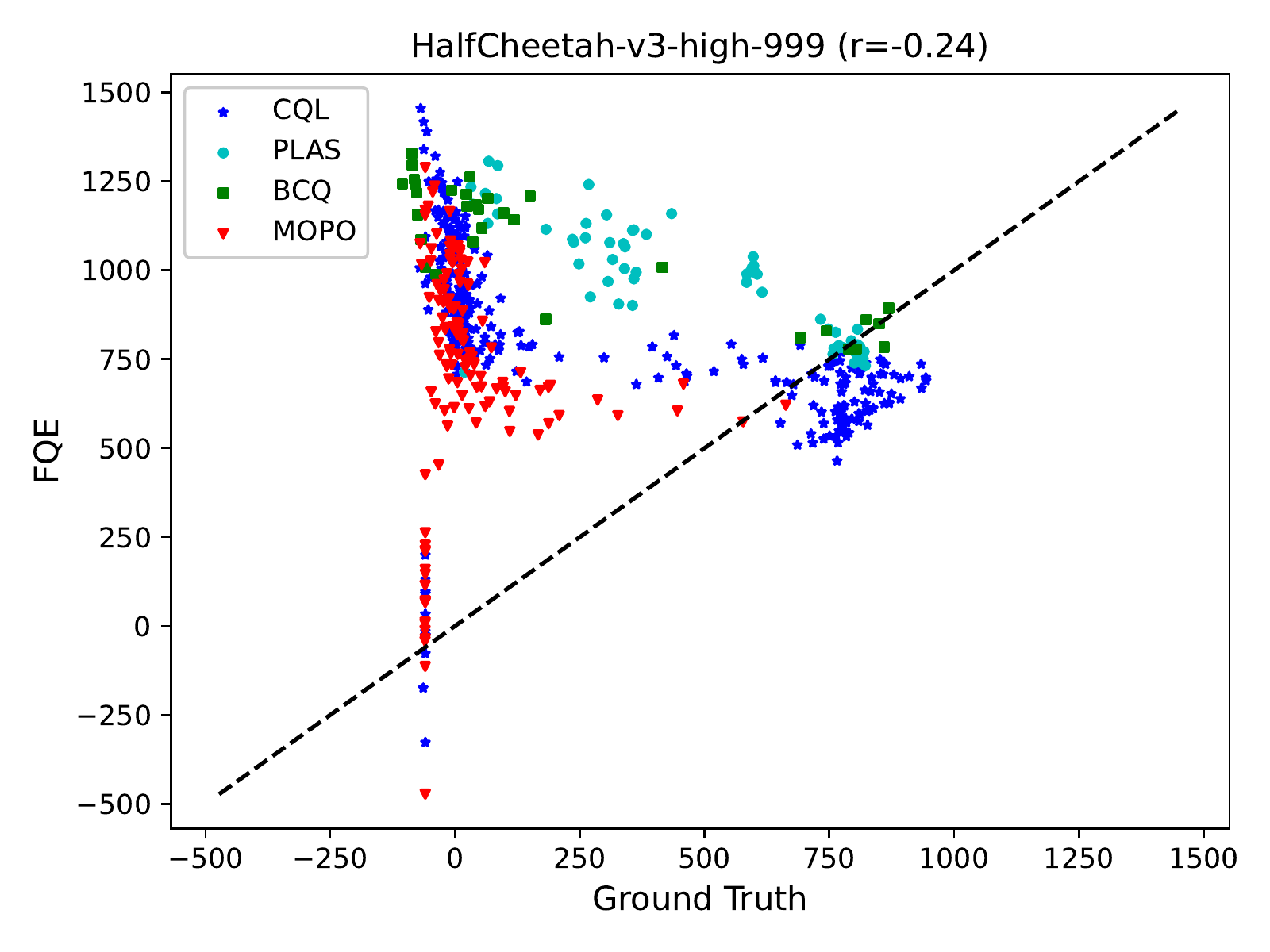}
	\end{minipage}
	\begin{minipage}[h]{0.3\linewidth}
		\includegraphics[width=\linewidth]{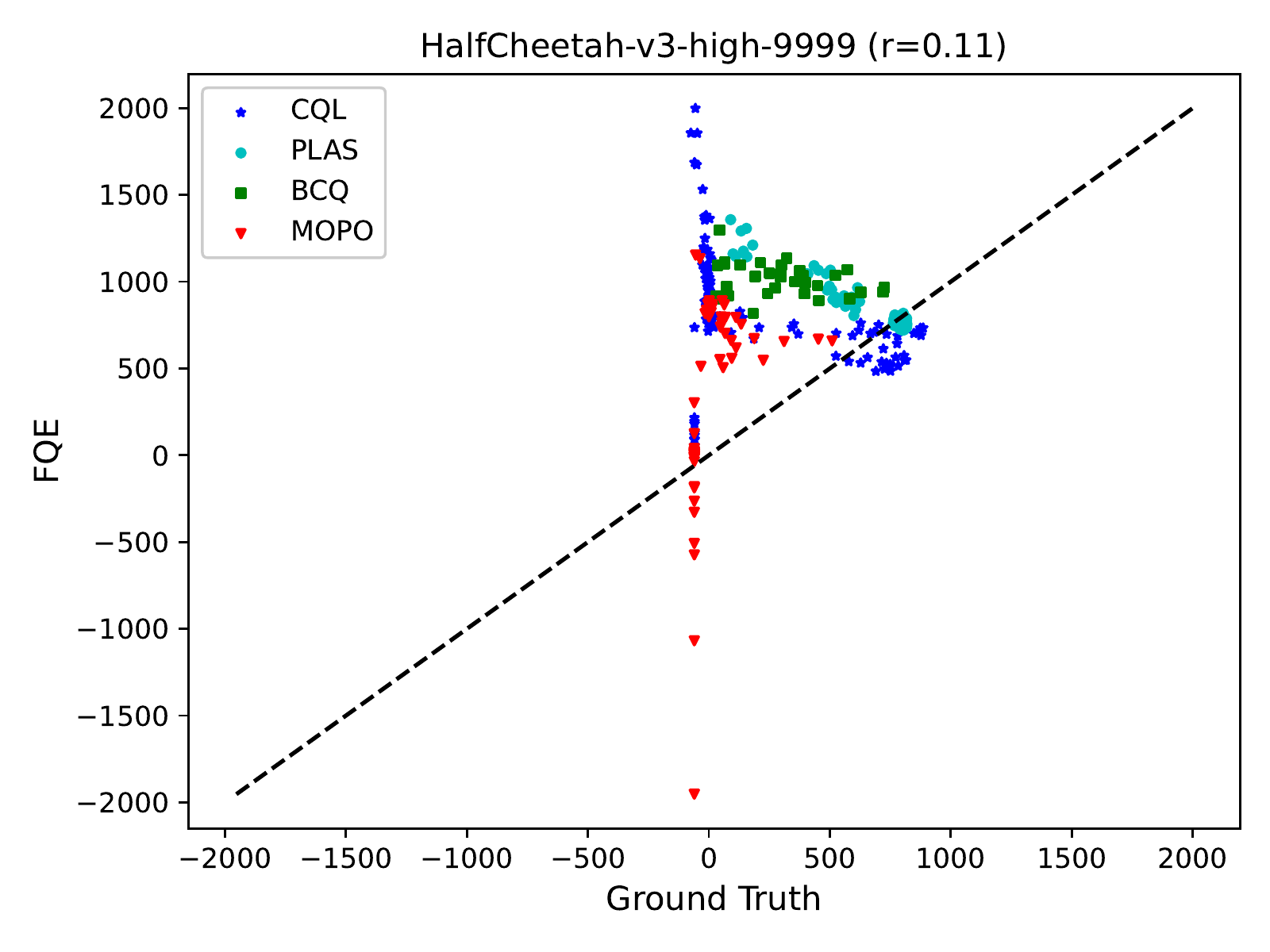}
	\end{minipage}
\end{figure*}

\begin{figure*}[ht]
	\centering
	\caption{FQE results on Hopper-v3 tasks. $r$ stands for the correlation coefficient.}
	\label{Hopper-fqe-figure}
	\begin{minipage}[h]{0.3\linewidth}
		\includegraphics[width=\linewidth]{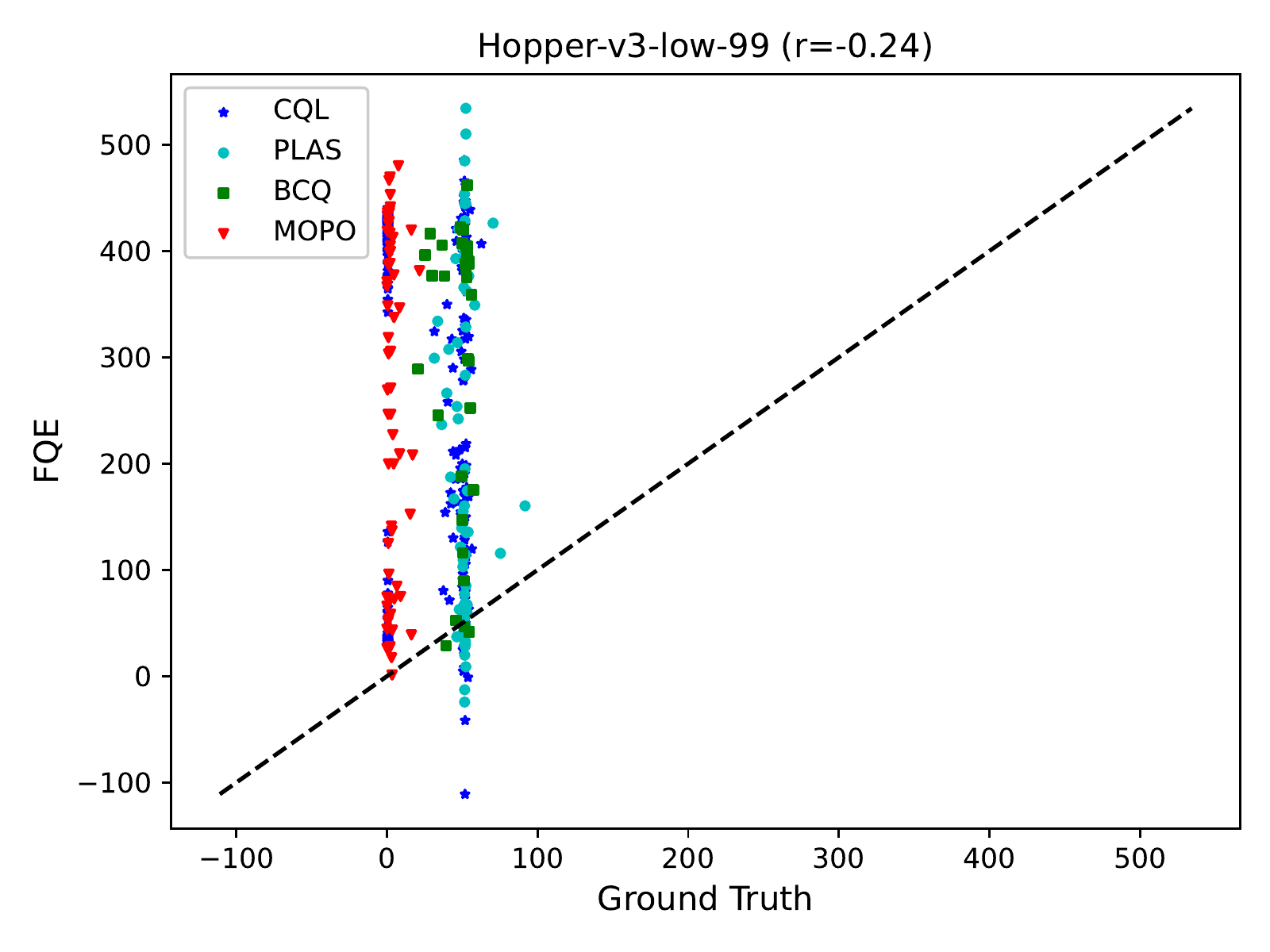}
	\end{minipage}
	\begin{minipage}[h]{0.3\linewidth}
		\includegraphics[width=\linewidth]{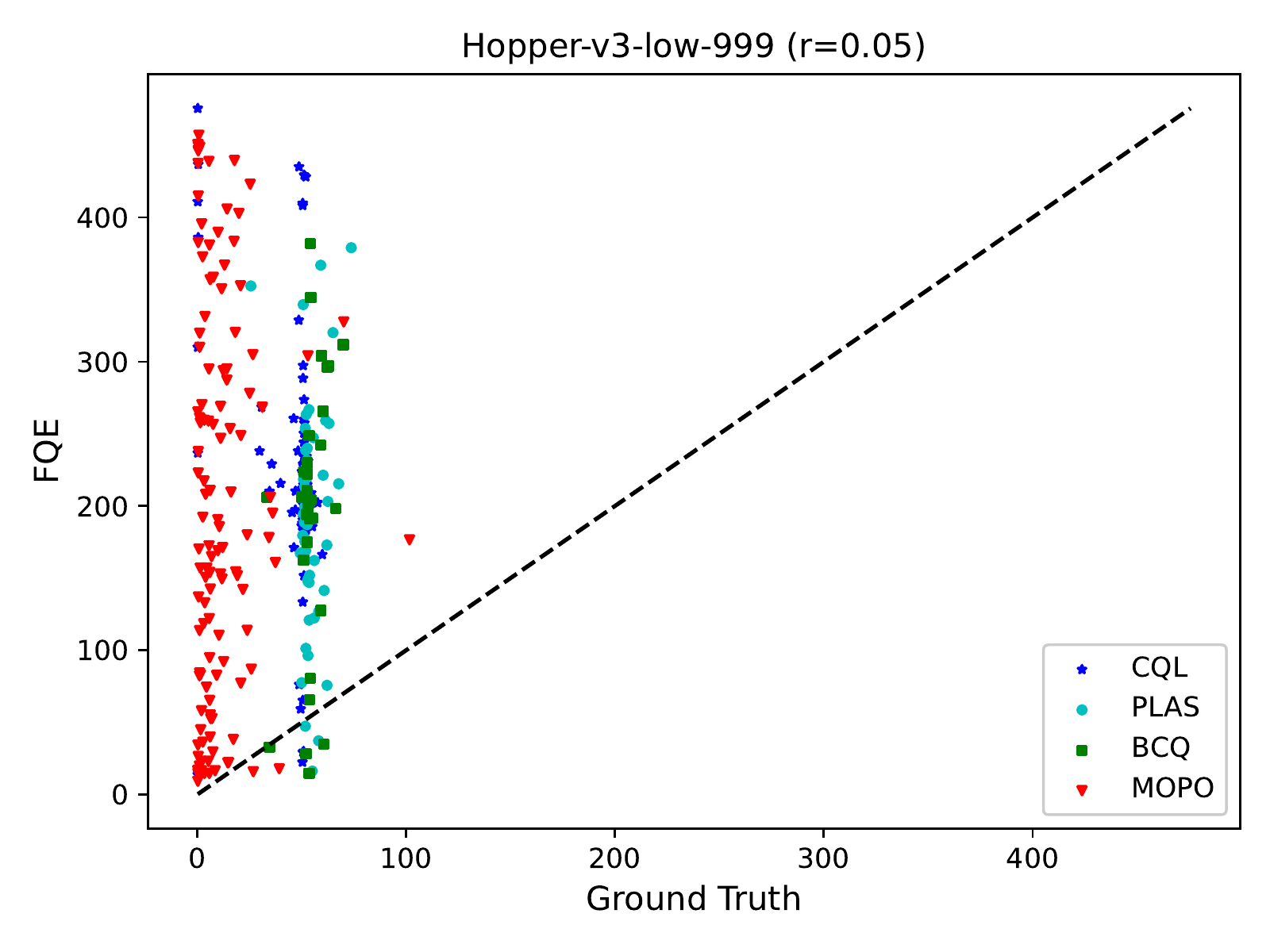}
	\end{minipage}
	\begin{minipage}[h]{0.3\linewidth}
		\includegraphics[width=\linewidth]{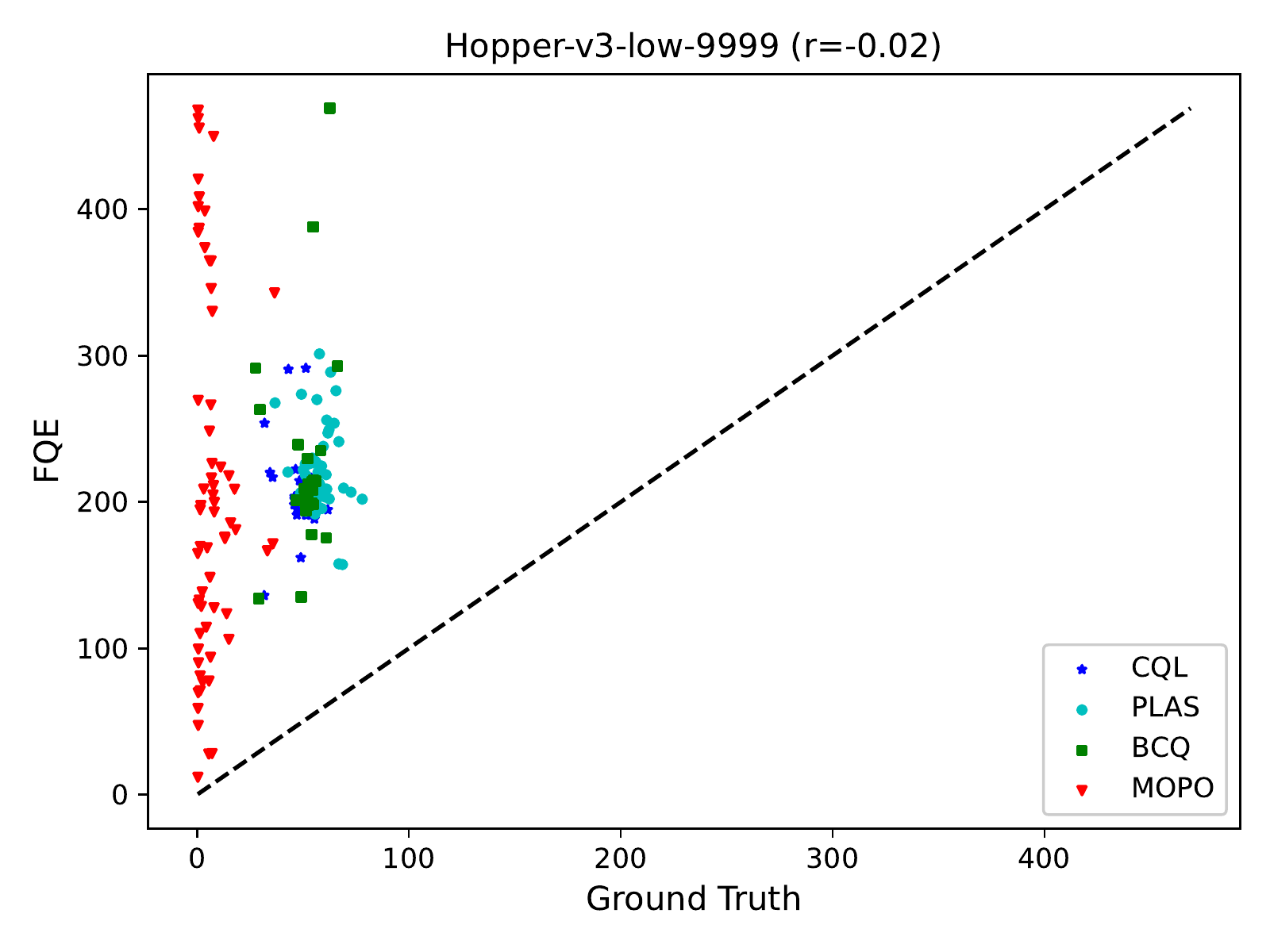}
	\end{minipage}
	
	\begin{minipage}[h]{0.3\linewidth}
		\includegraphics[width=\linewidth]{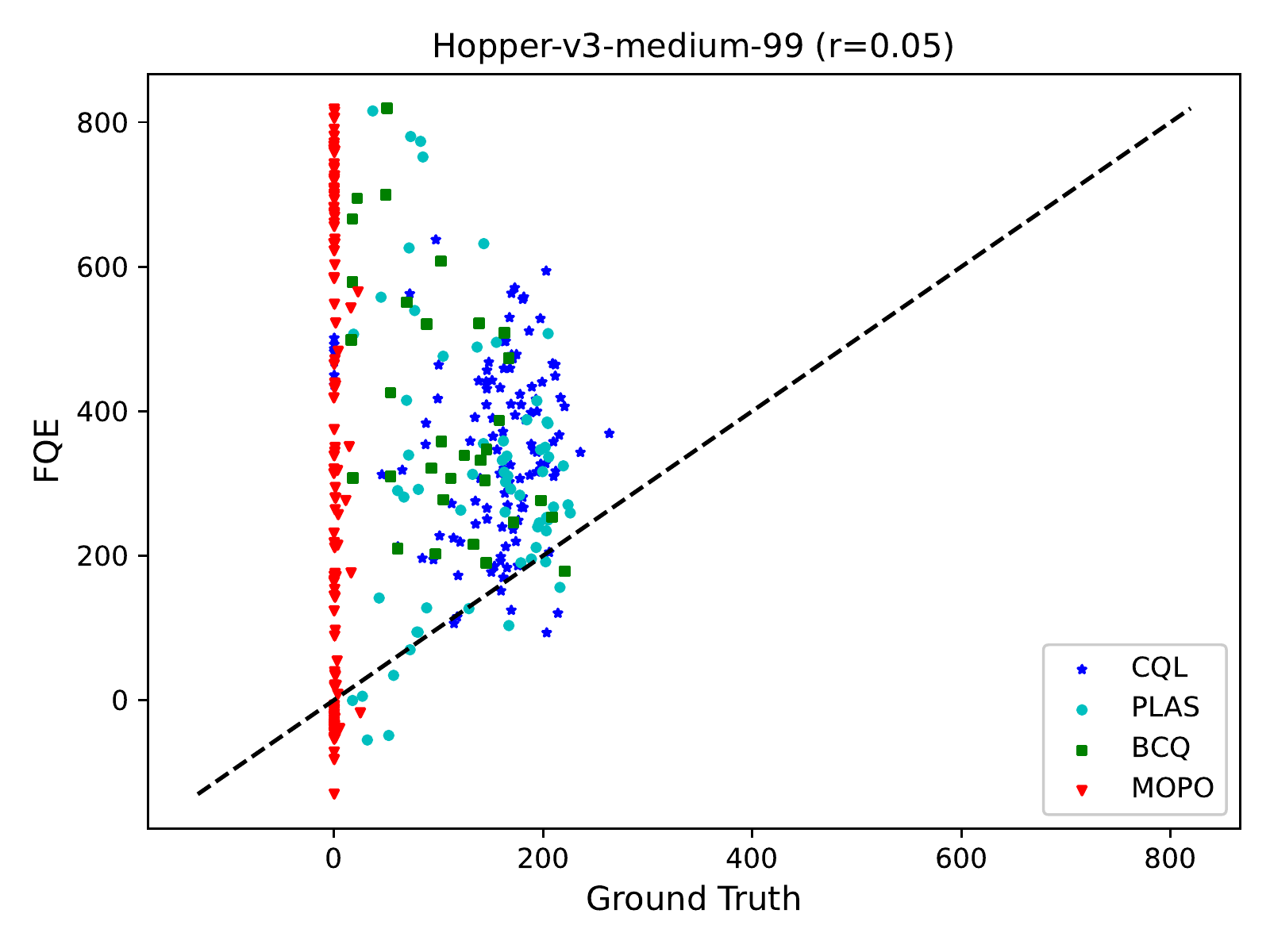}
	\end{minipage}
	\begin{minipage}[h]{0.3\linewidth}
		\includegraphics[width=\linewidth]{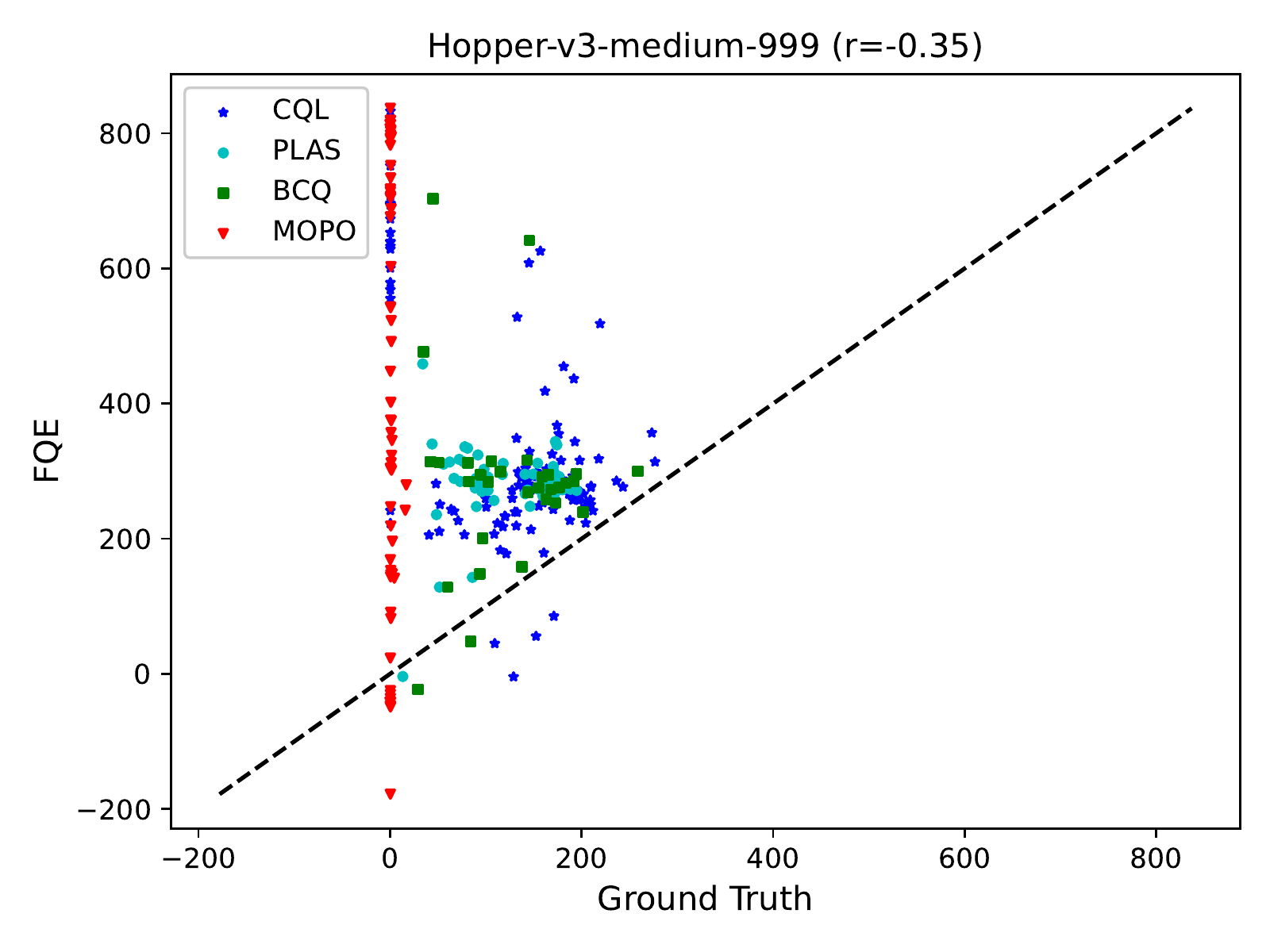}
	\end{minipage}
	\begin{minipage}[h]{0.3\linewidth}
		\includegraphics[width=\linewidth]{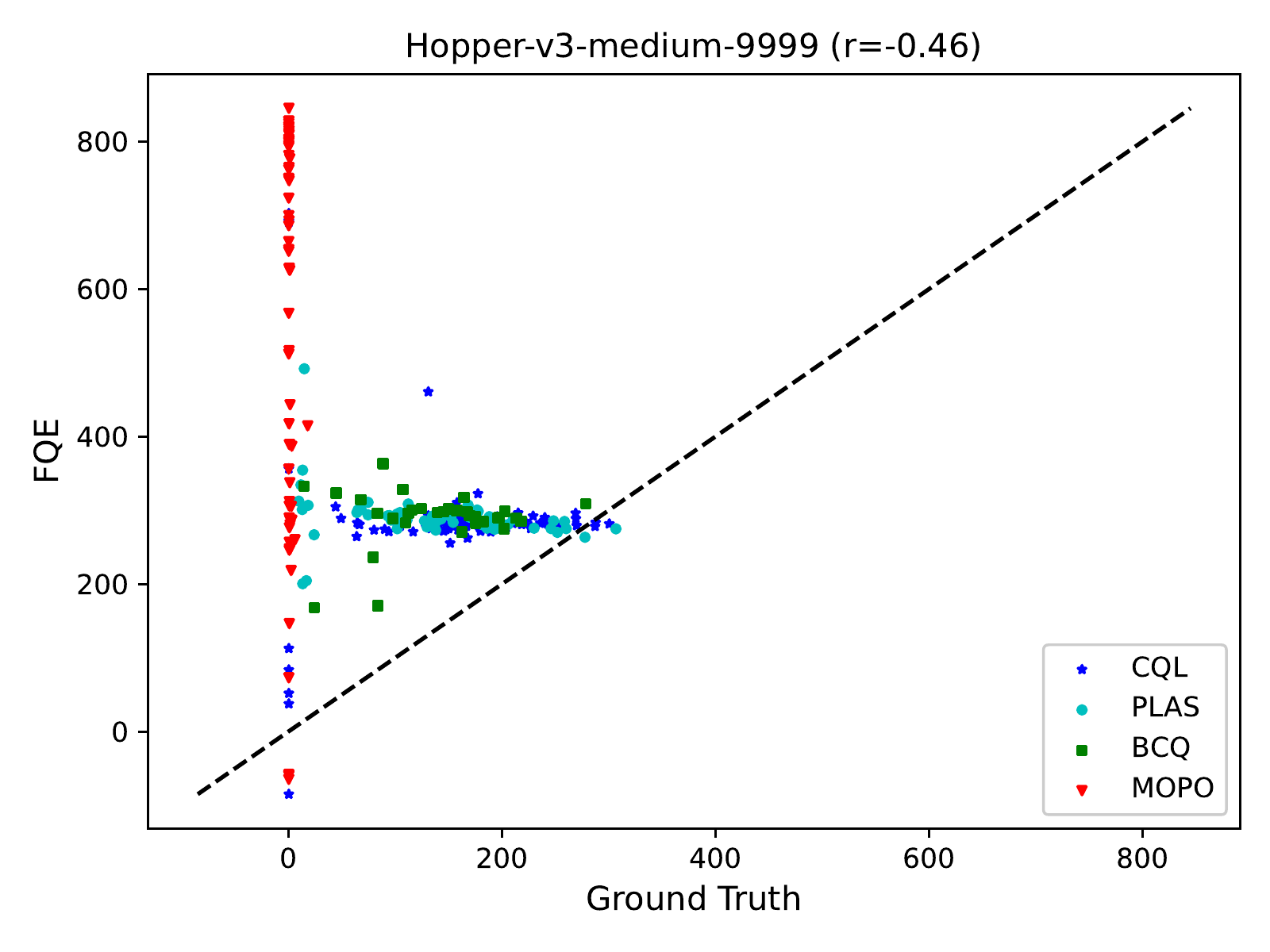}
	\end{minipage}
	
	\begin{minipage}[h]{0.3\linewidth}
		\includegraphics[width=\linewidth]{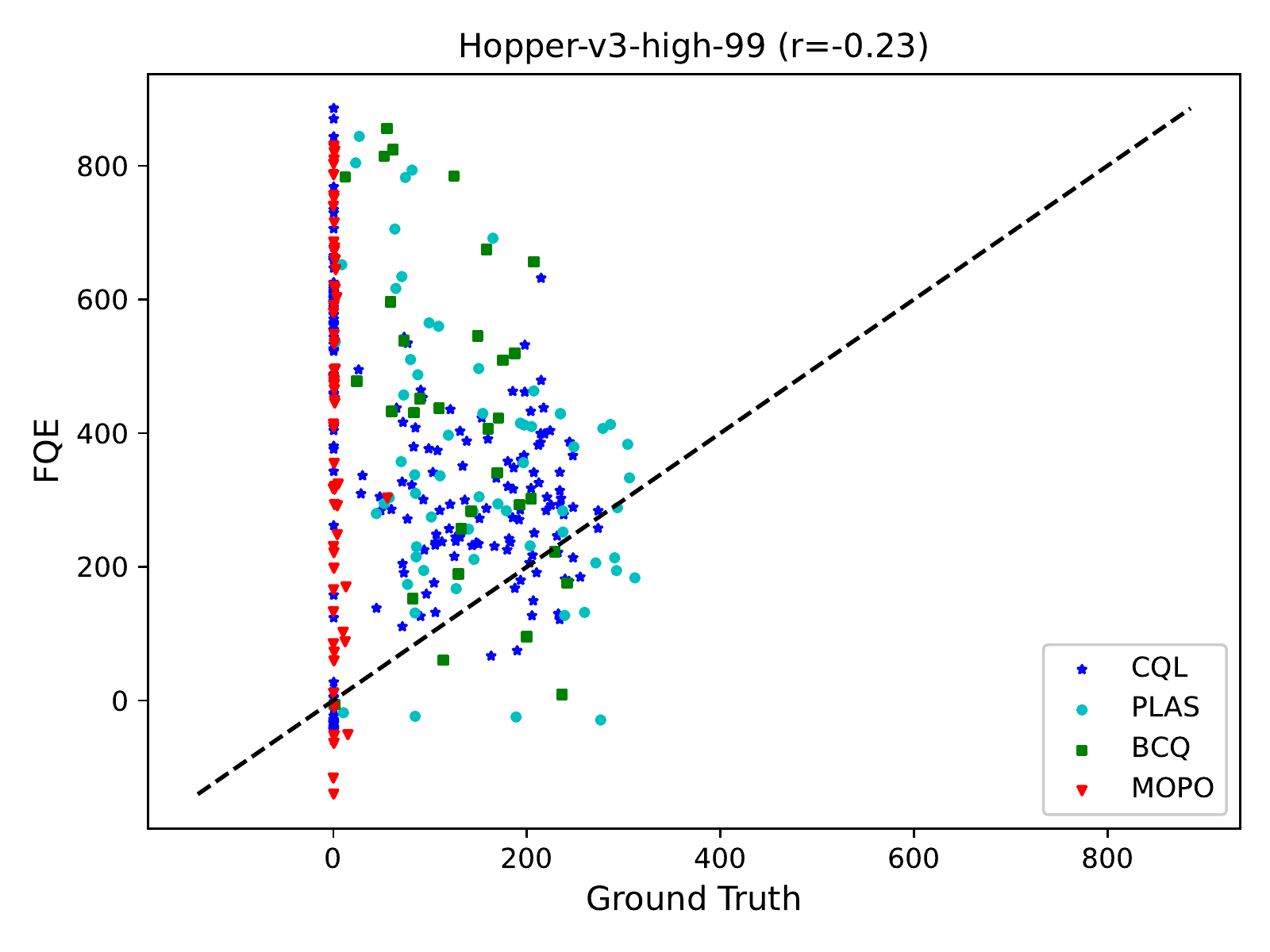}
	\end{minipage}
	\begin{minipage}[h]{0.3\linewidth}
		\includegraphics[width=\linewidth]{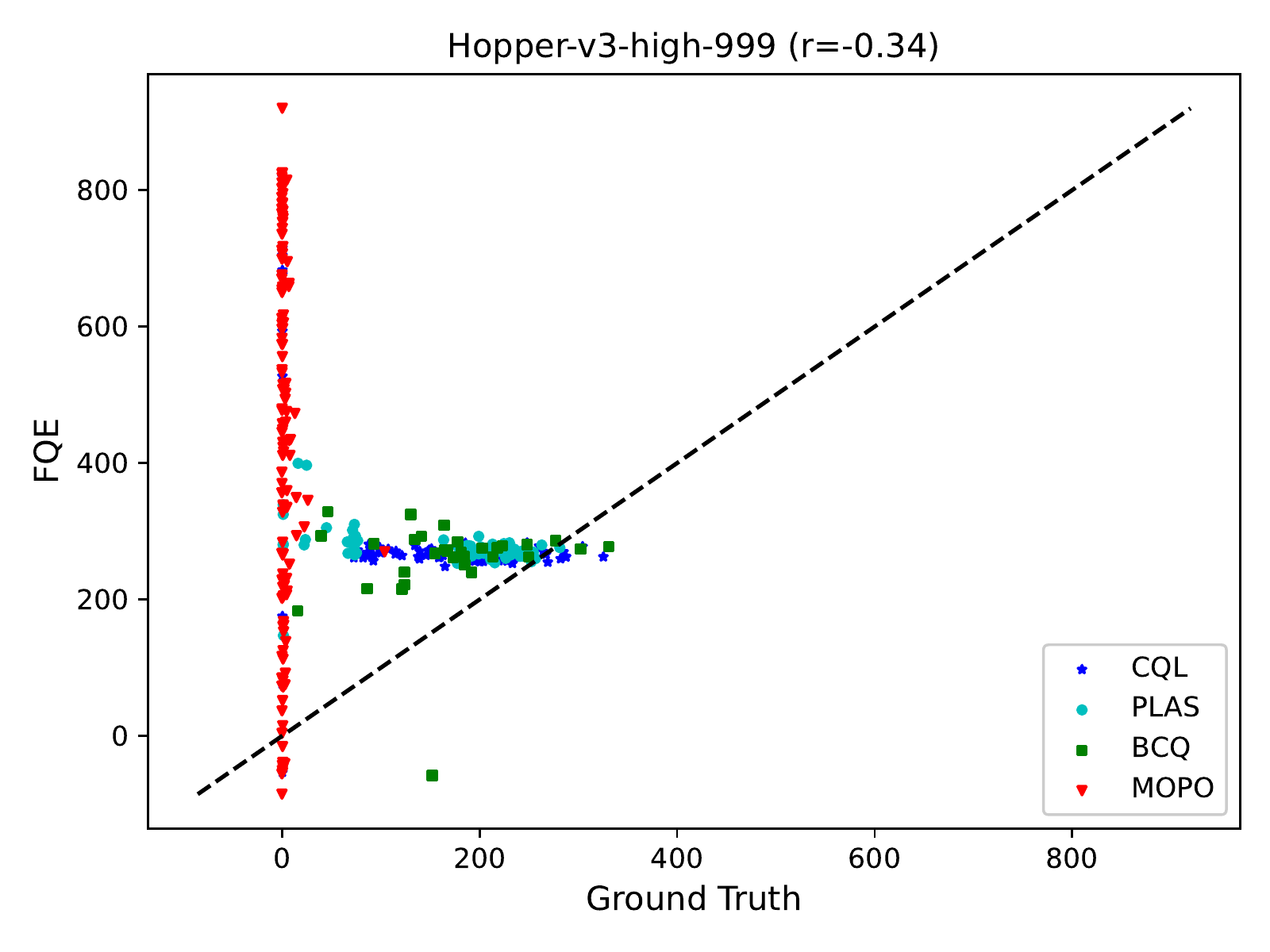}
	\end{minipage}
	\begin{minipage}[h]{0.3\linewidth}
		\includegraphics[width=\linewidth]{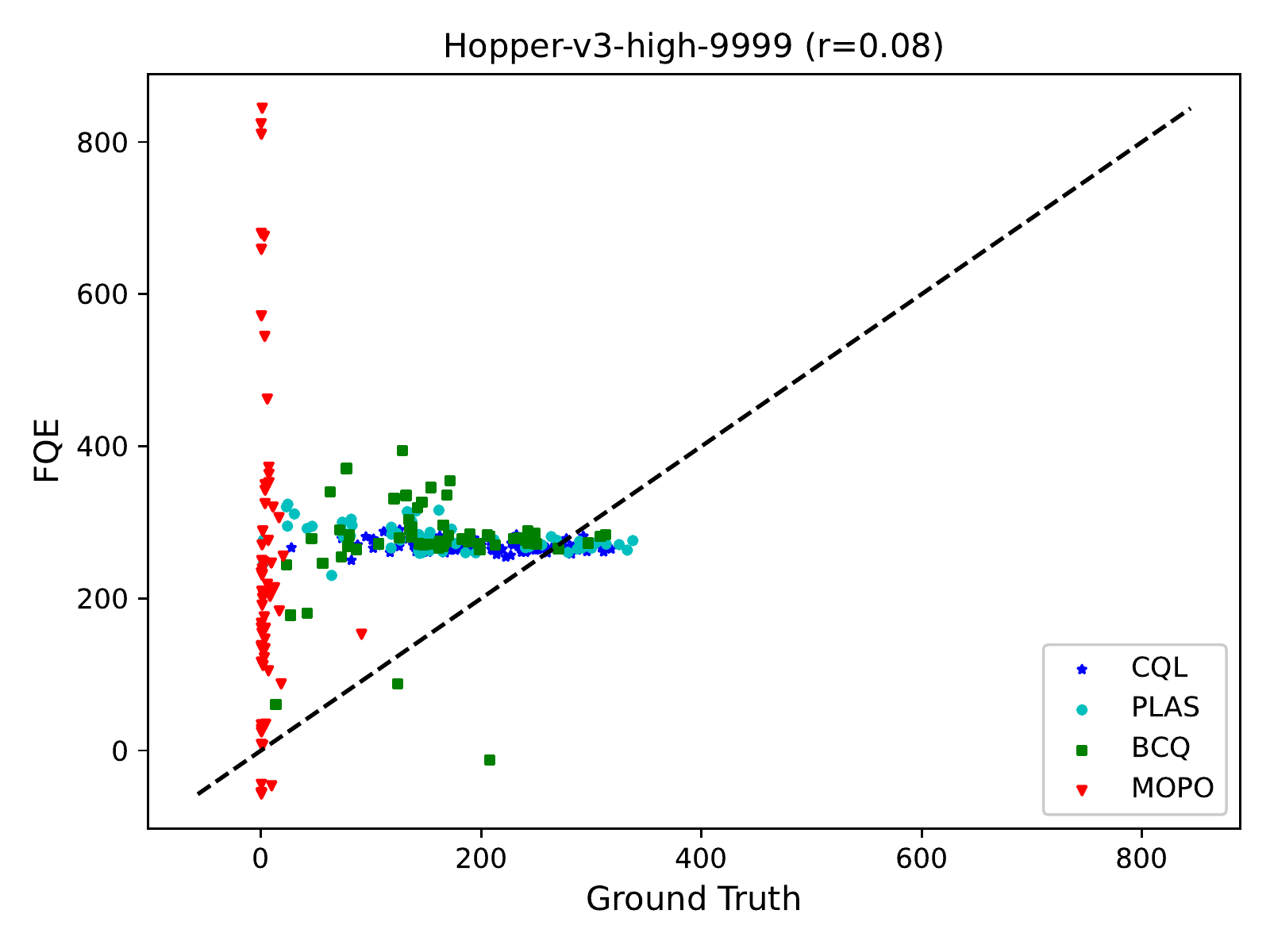}
	\end{minipage}
\end{figure*}

\begin{figure*}[ht]
	\centering
	\caption{FQE results on IB tasks. $r$ stands for the correlation coefficient.}
	\label{IB-fqe-figure}
	\begin{minipage}[h]{0.3\linewidth}
		\includegraphics[width=\linewidth]{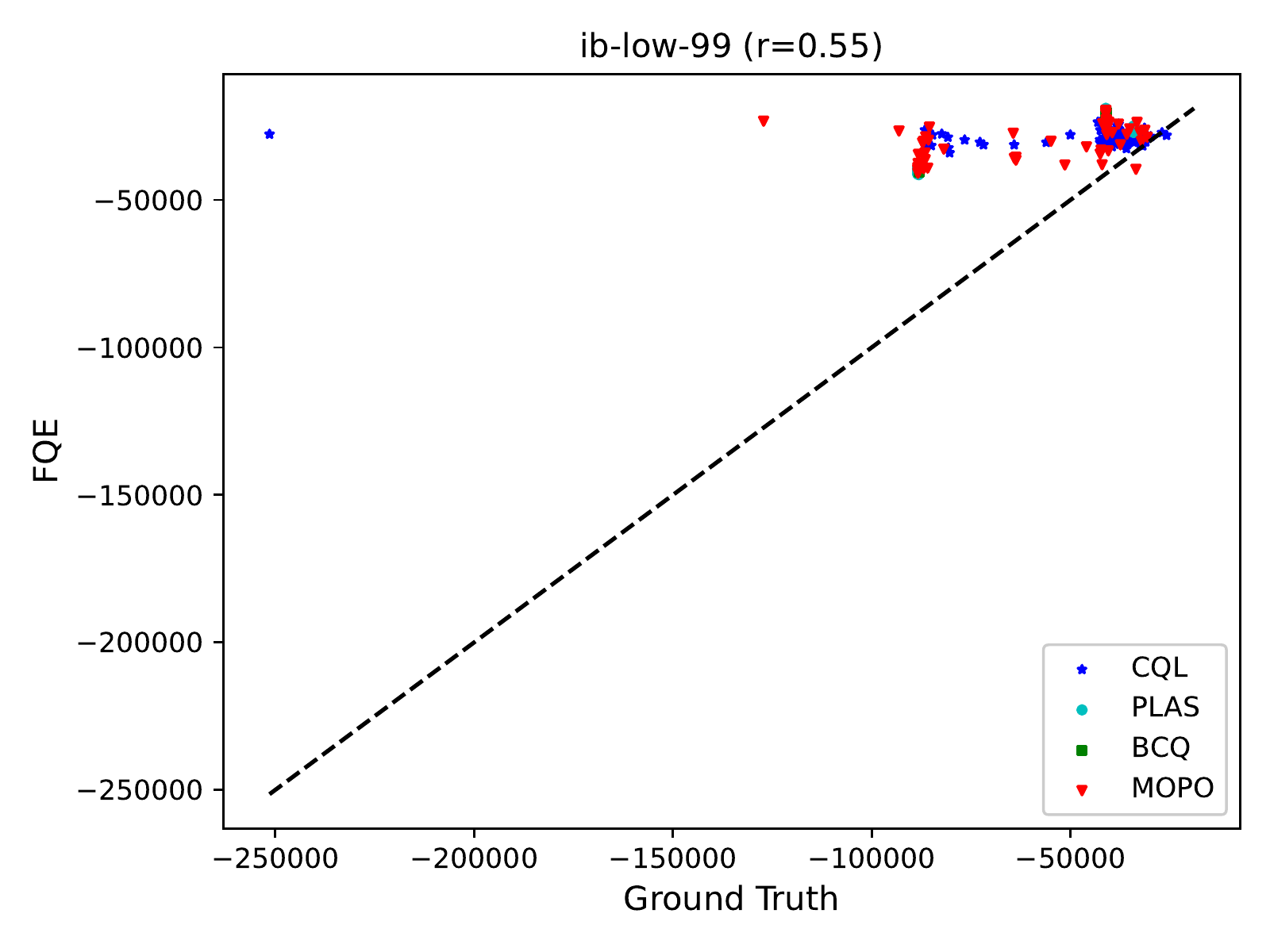}
	\end{minipage}
	\begin{minipage}[h]{0.3\linewidth}
		\includegraphics[width=\linewidth]{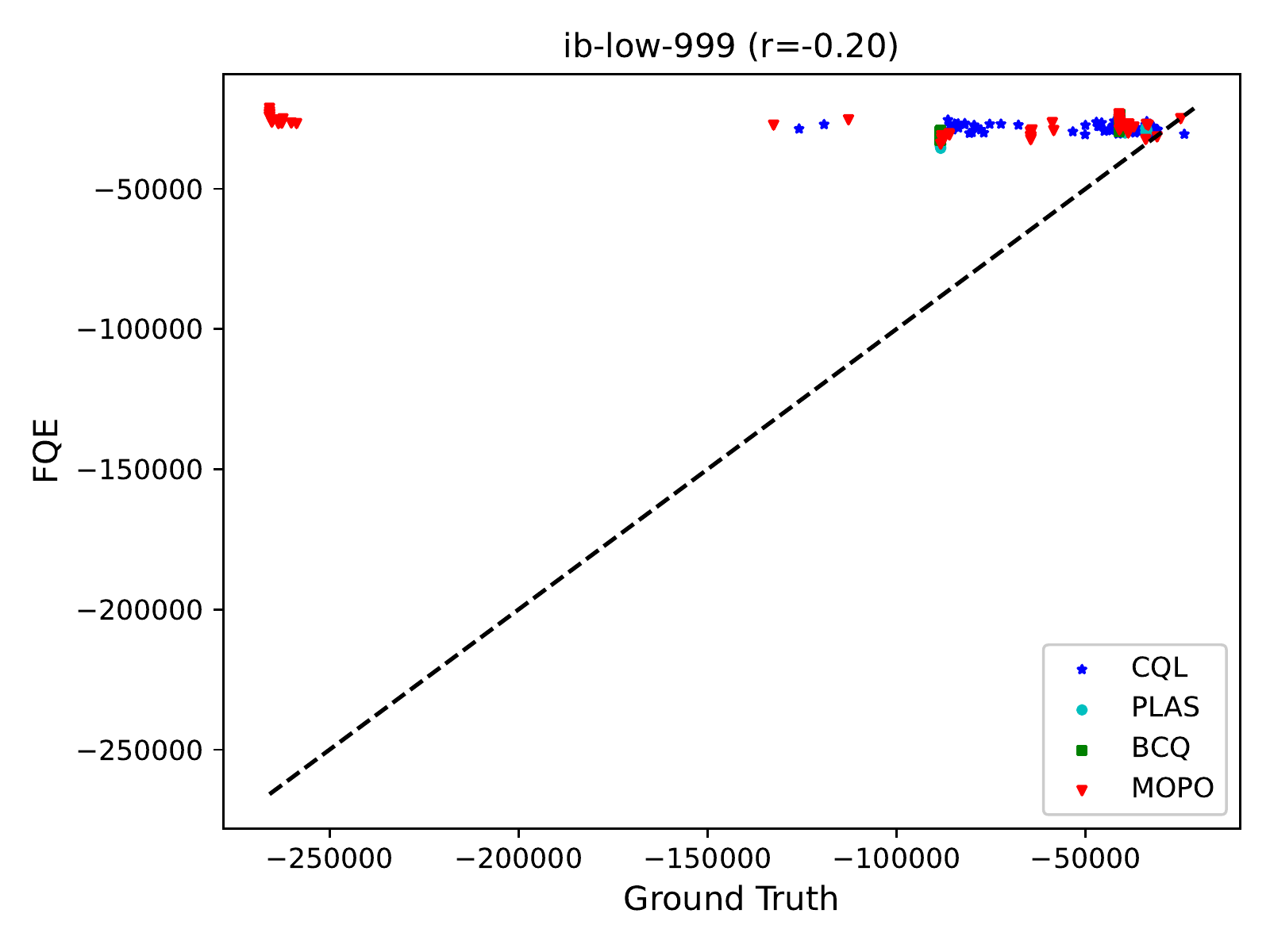}
	\end{minipage}
	\begin{minipage}[h]{0.3\linewidth}
		\includegraphics[width=\linewidth]{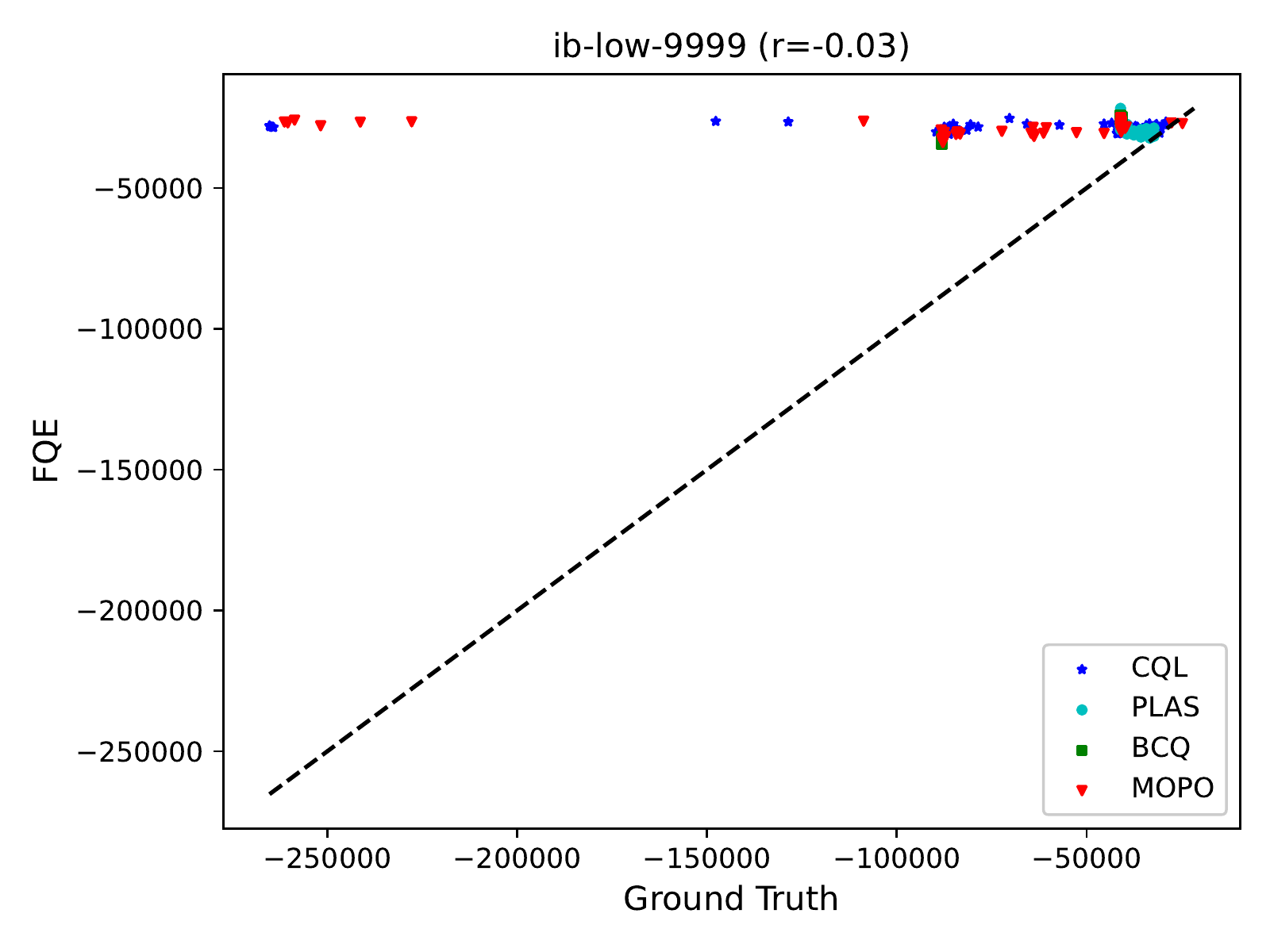}
	\end{minipage}
	
	\begin{minipage}[h]{0.3\linewidth}
		\includegraphics[width=\linewidth]{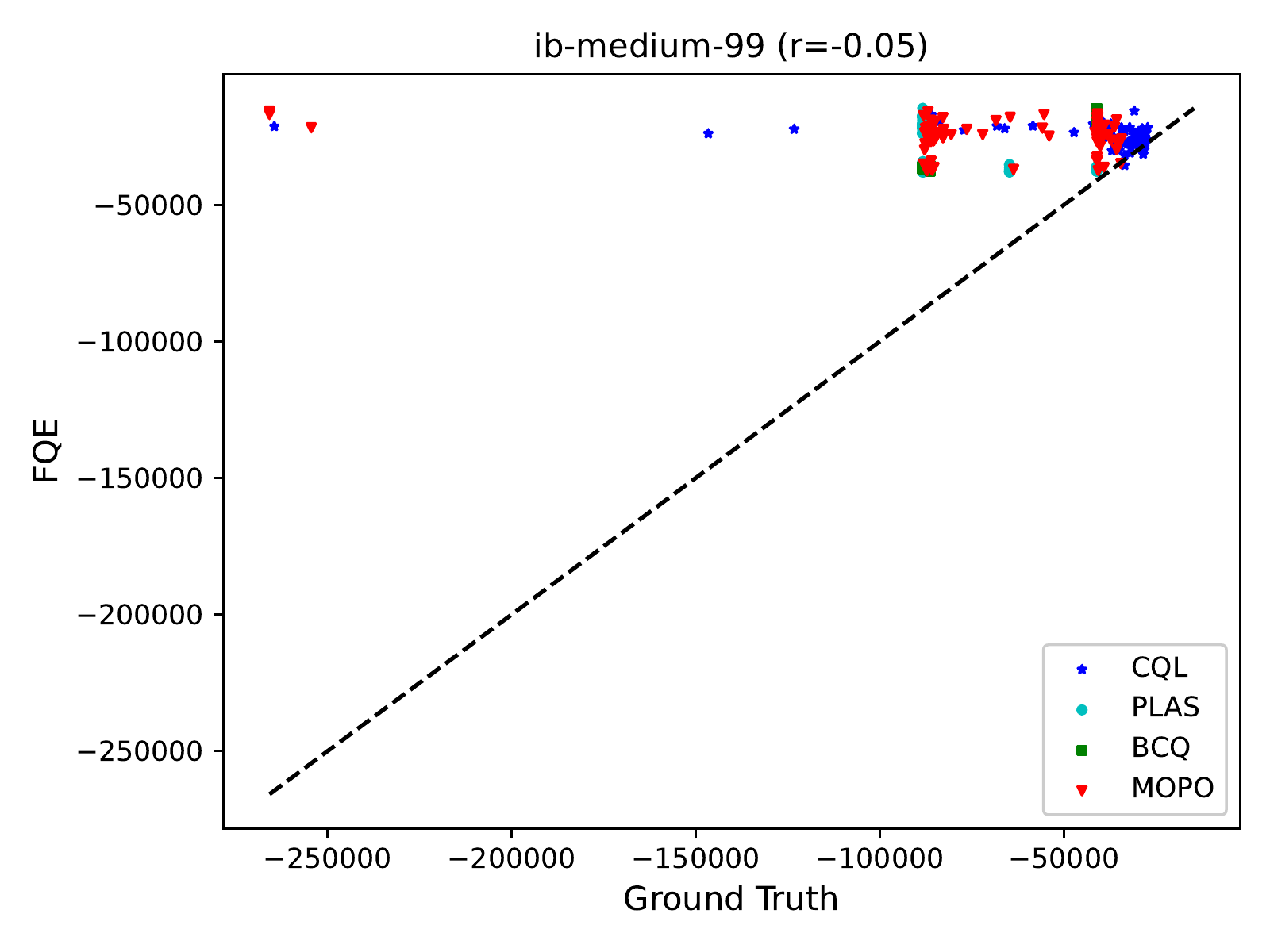}
	\end{minipage}
	\begin{minipage}[h]{0.3\linewidth}
		\includegraphics[width=\linewidth]{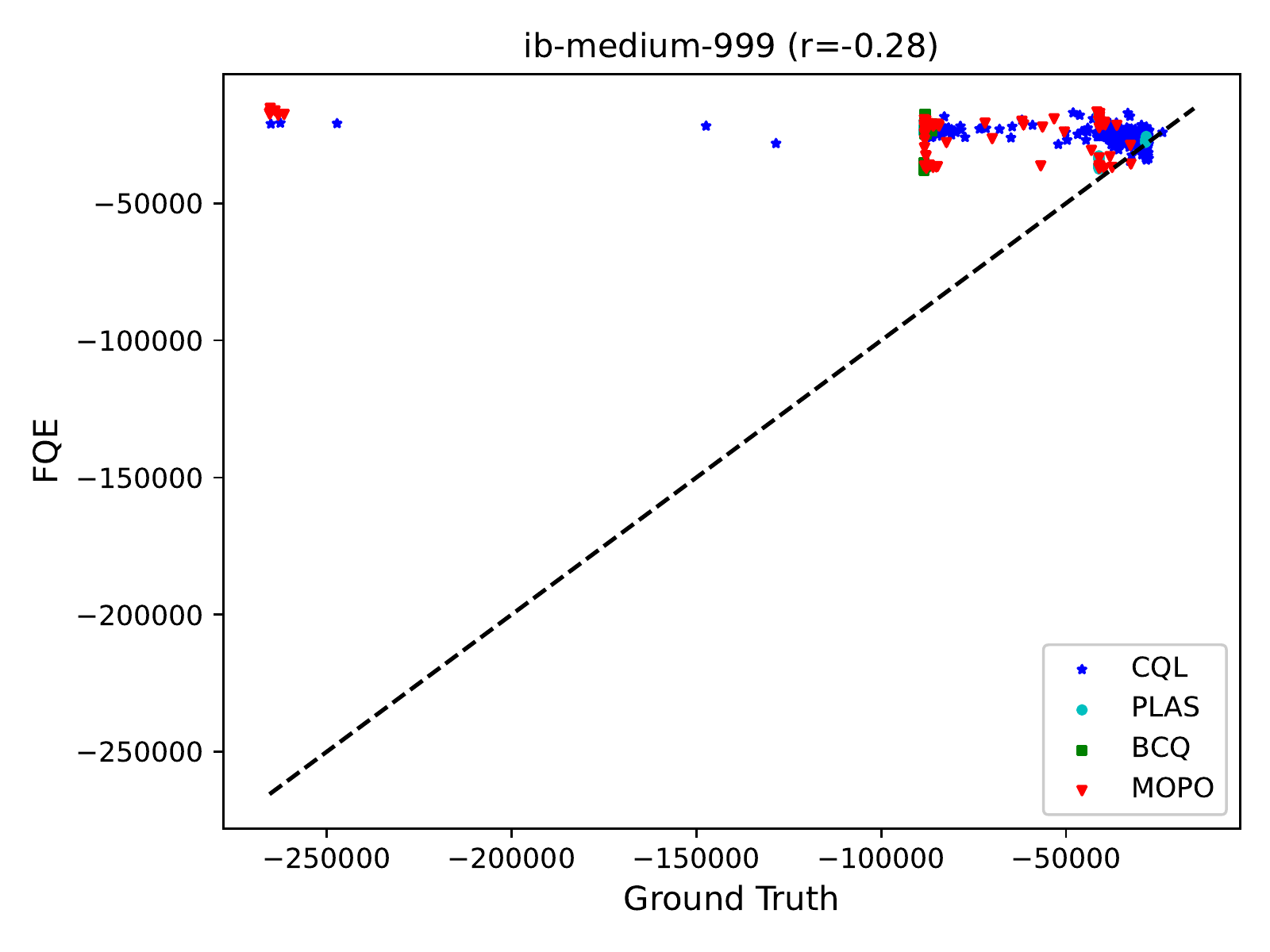}
	\end{minipage}
	\begin{minipage}[h]{0.3\linewidth}
		\includegraphics[width=\linewidth]{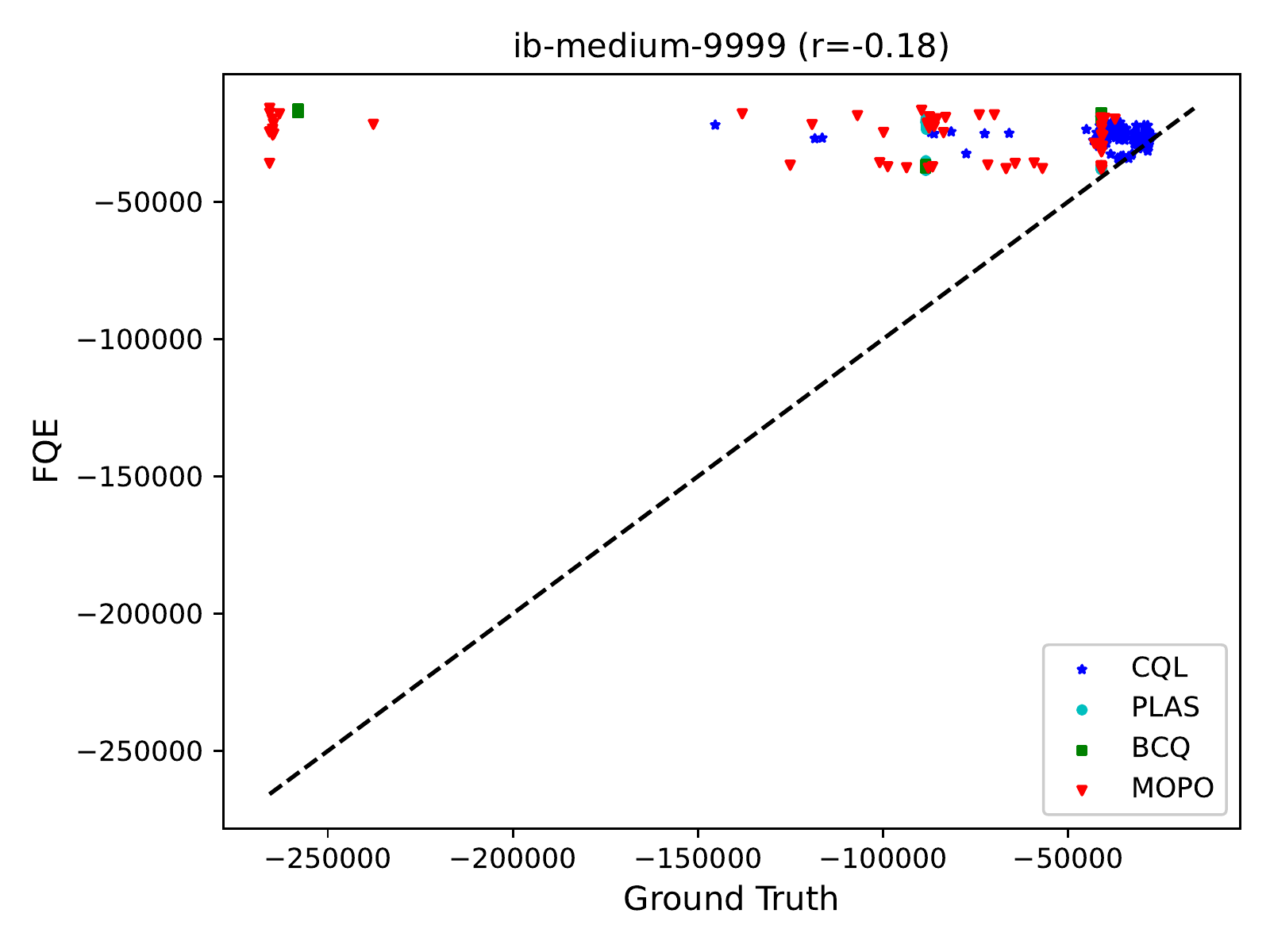}
	\end{minipage}
	
	\begin{minipage}[h]{0.3\linewidth}
		\includegraphics[width=\linewidth]{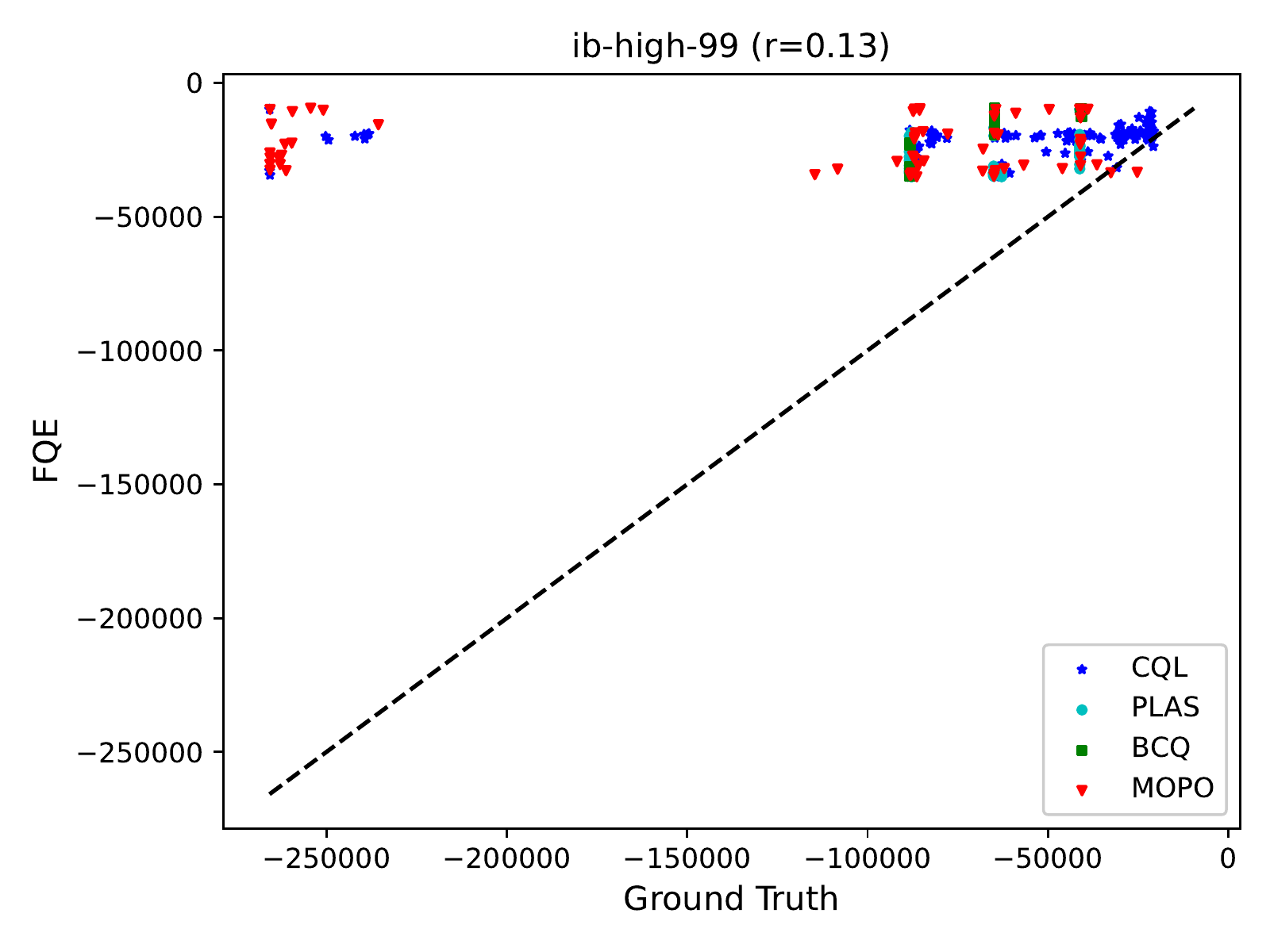}
	\end{minipage}
	\begin{minipage}[h]{0.3\linewidth}
		\includegraphics[width=\linewidth]{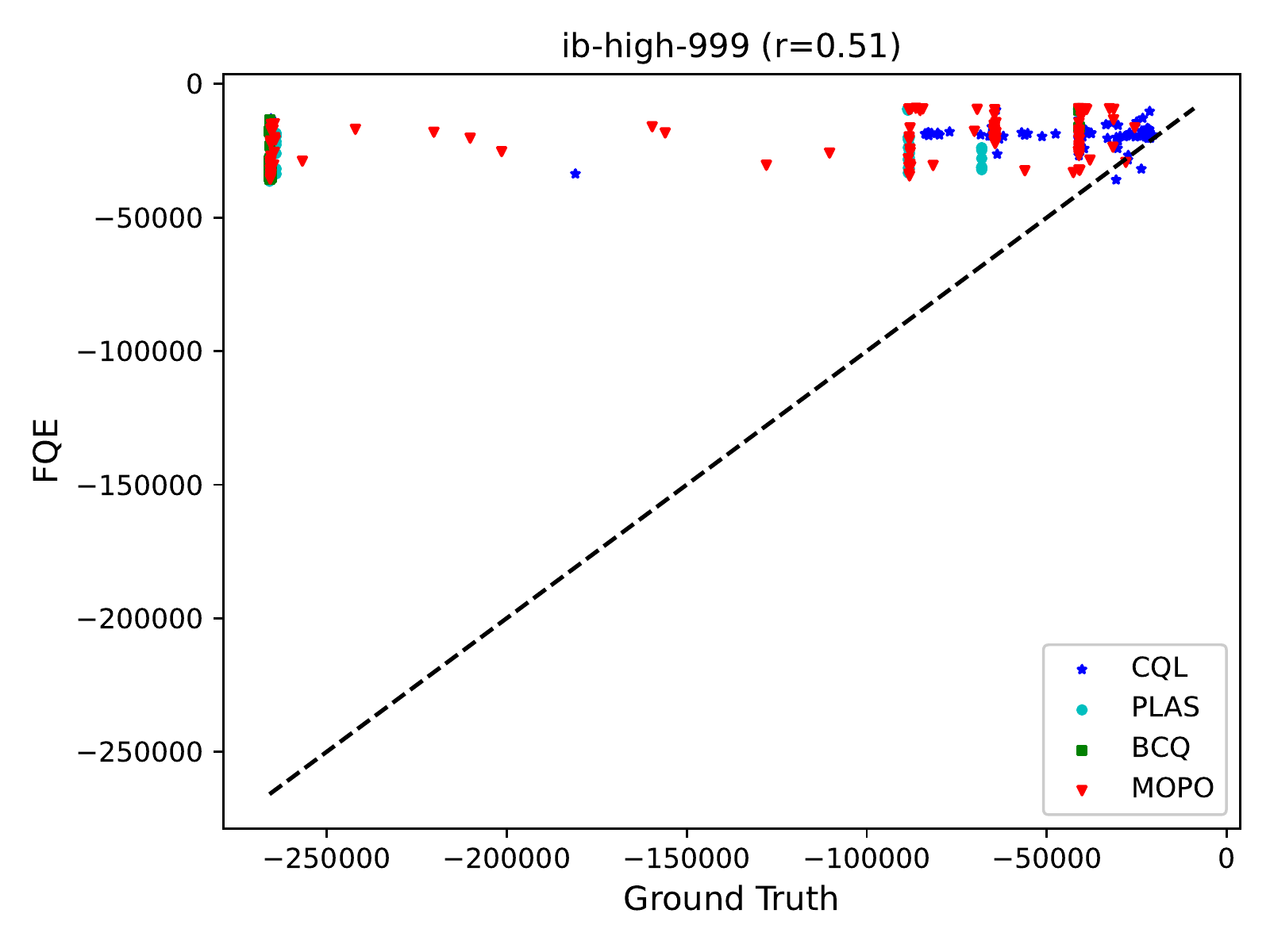}
	\end{minipage}
	\begin{minipage}[h]{0.3\linewidth}
		\includegraphics[width=\linewidth]{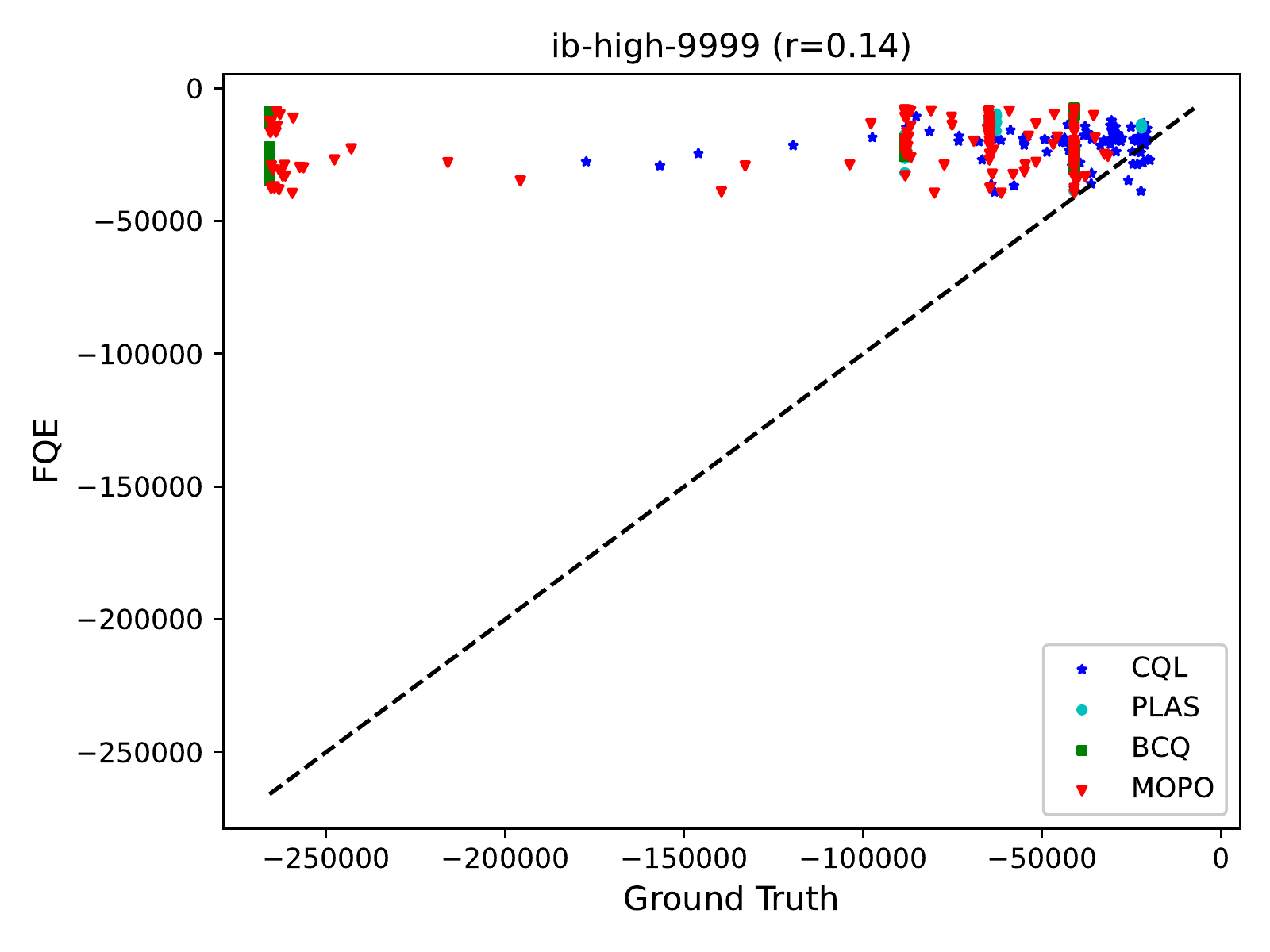}
	\end{minipage}
\end{figure*}

\begin{figure*}[ht]
	\centering
	\caption{FQE results on CityLearn tasks. $r$ stands for the correlation coefficient.}
	\label{CityLearn-fqe-figure}
	\begin{minipage}[h]{0.3\linewidth}
		\includegraphics[width=\linewidth]{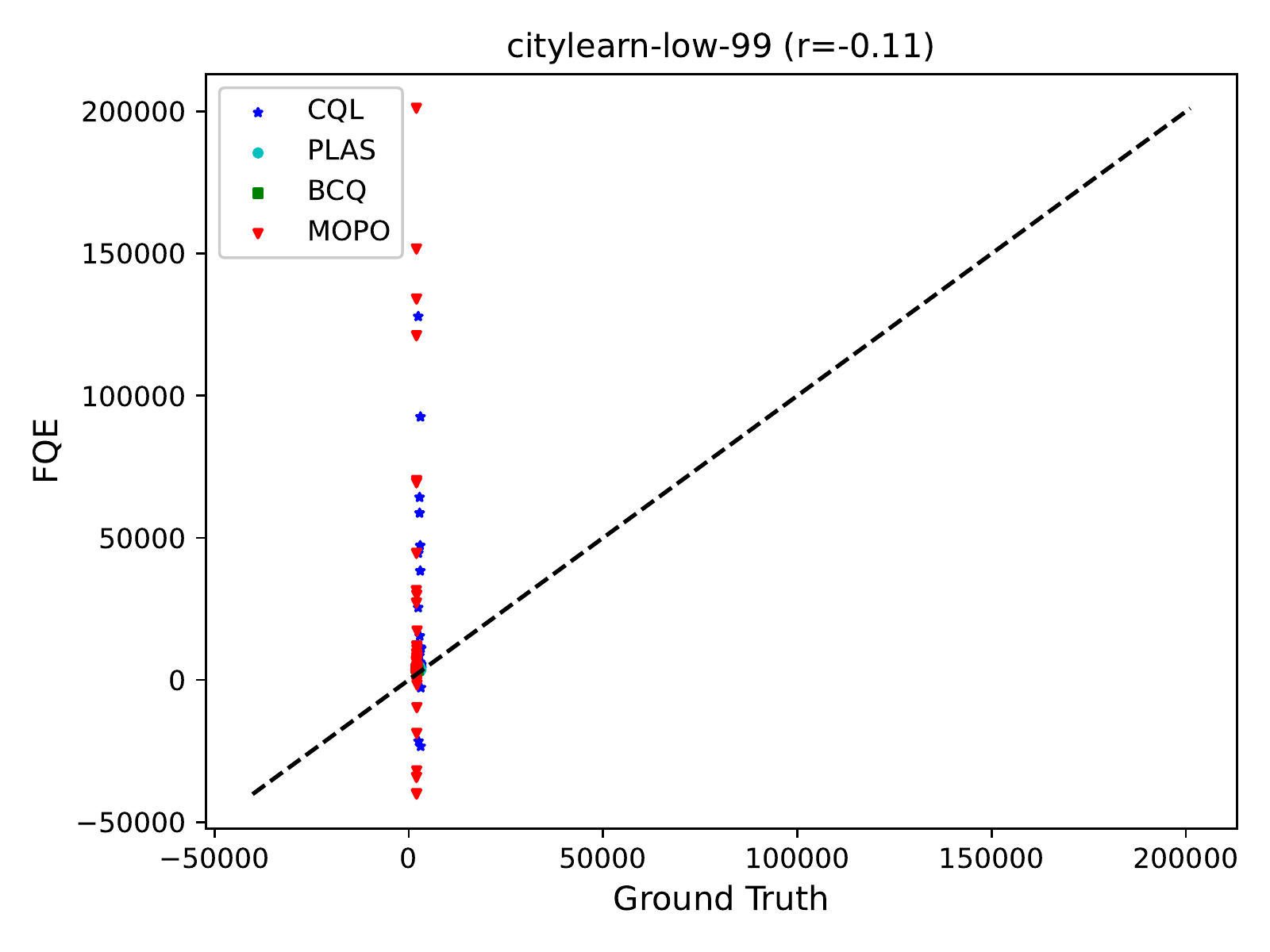}
	\end{minipage}
	\begin{minipage}[h]{0.3\linewidth}
		\includegraphics[width=\linewidth]{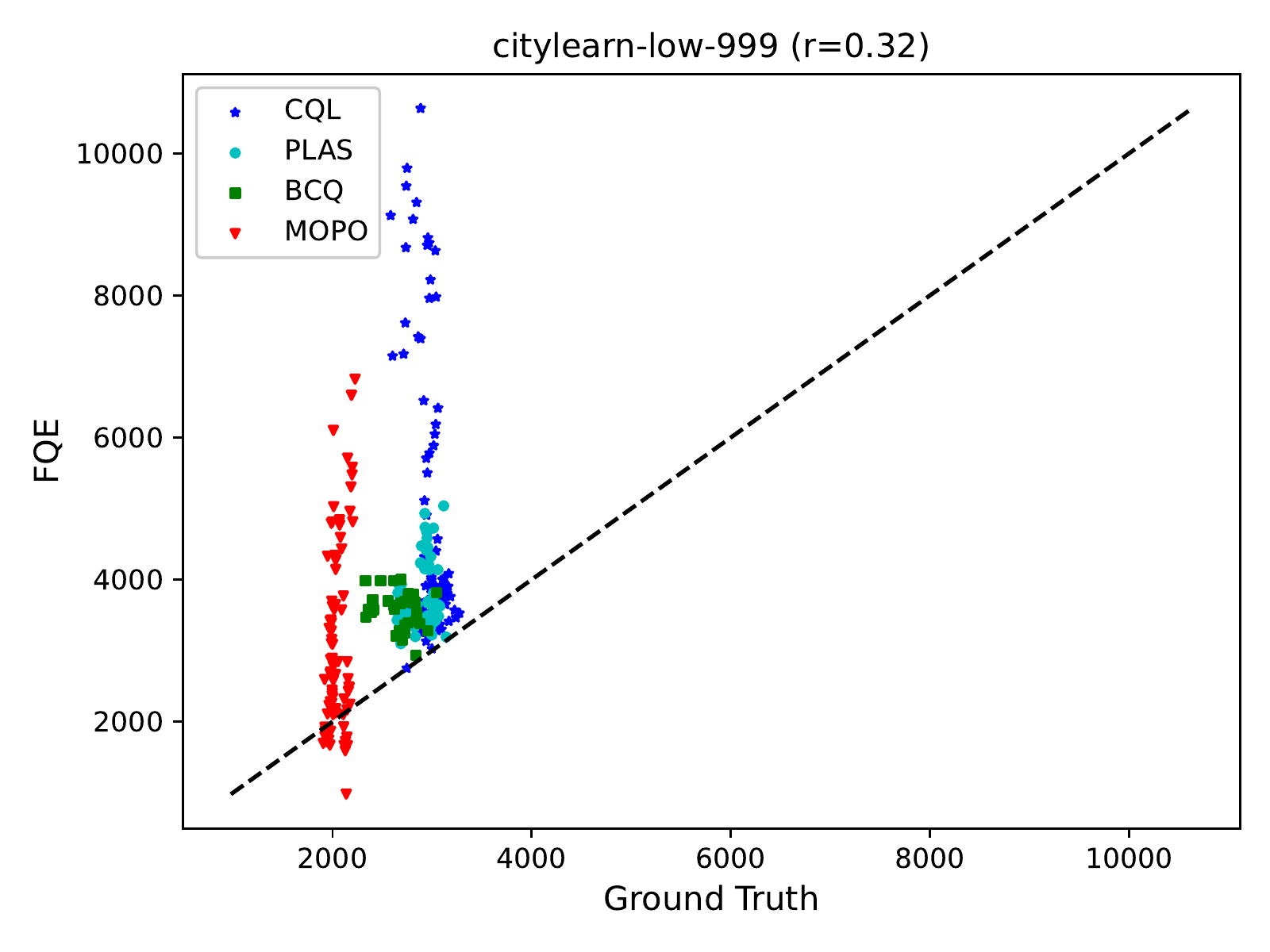}
	\end{minipage}
	\begin{minipage}[h]{0.3\linewidth}
		\includegraphics[width=\linewidth]{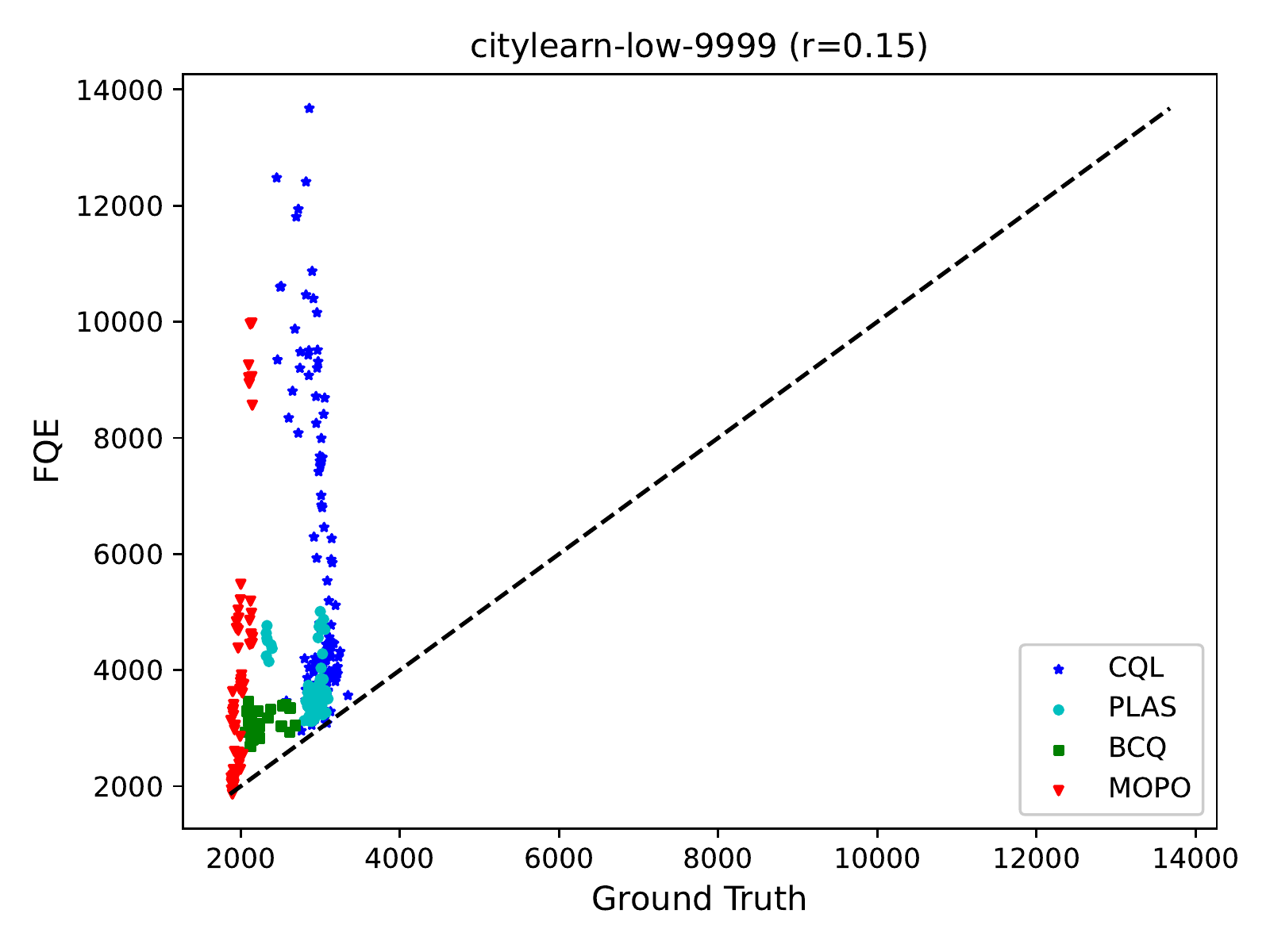}
	\end{minipage}
	
	\begin{minipage}[h]{0.3\linewidth}
		\includegraphics[width=\linewidth]{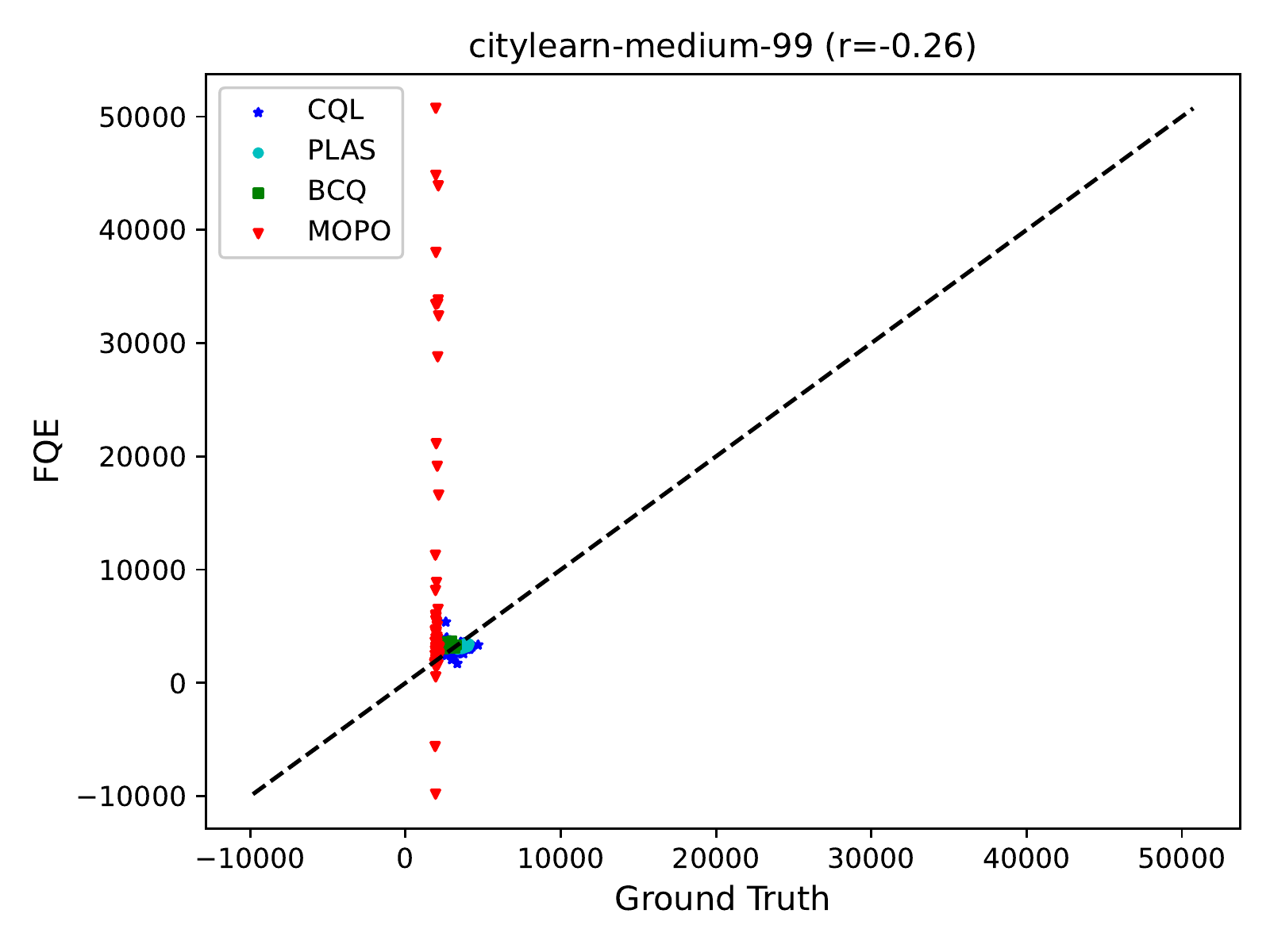}
	\end{minipage}
	\begin{minipage}[h]{0.3\linewidth}
		\includegraphics[width=\linewidth]{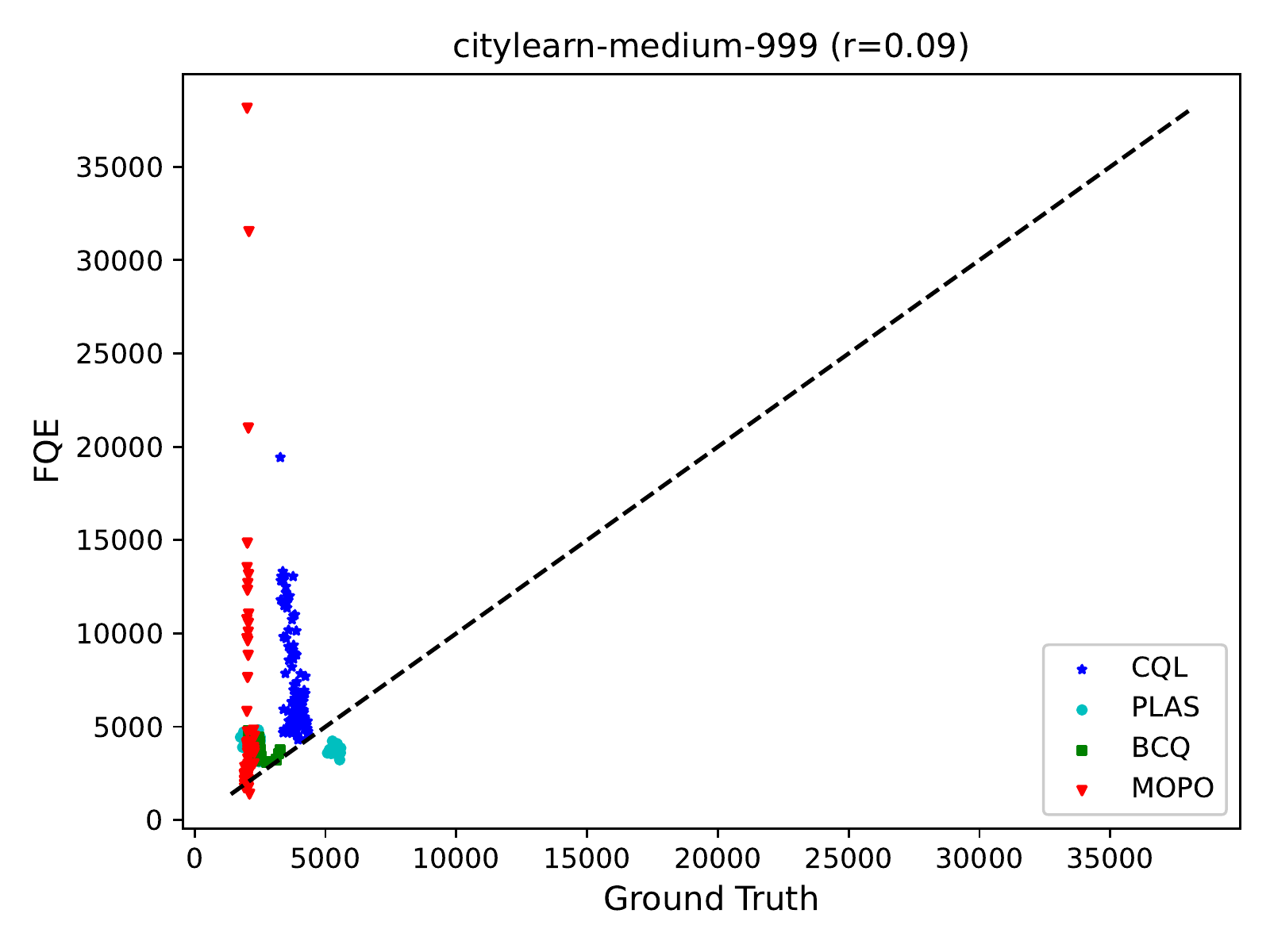}
	\end{minipage}
	\begin{minipage}[h]{0.3\linewidth}
		\includegraphics[width=\linewidth]{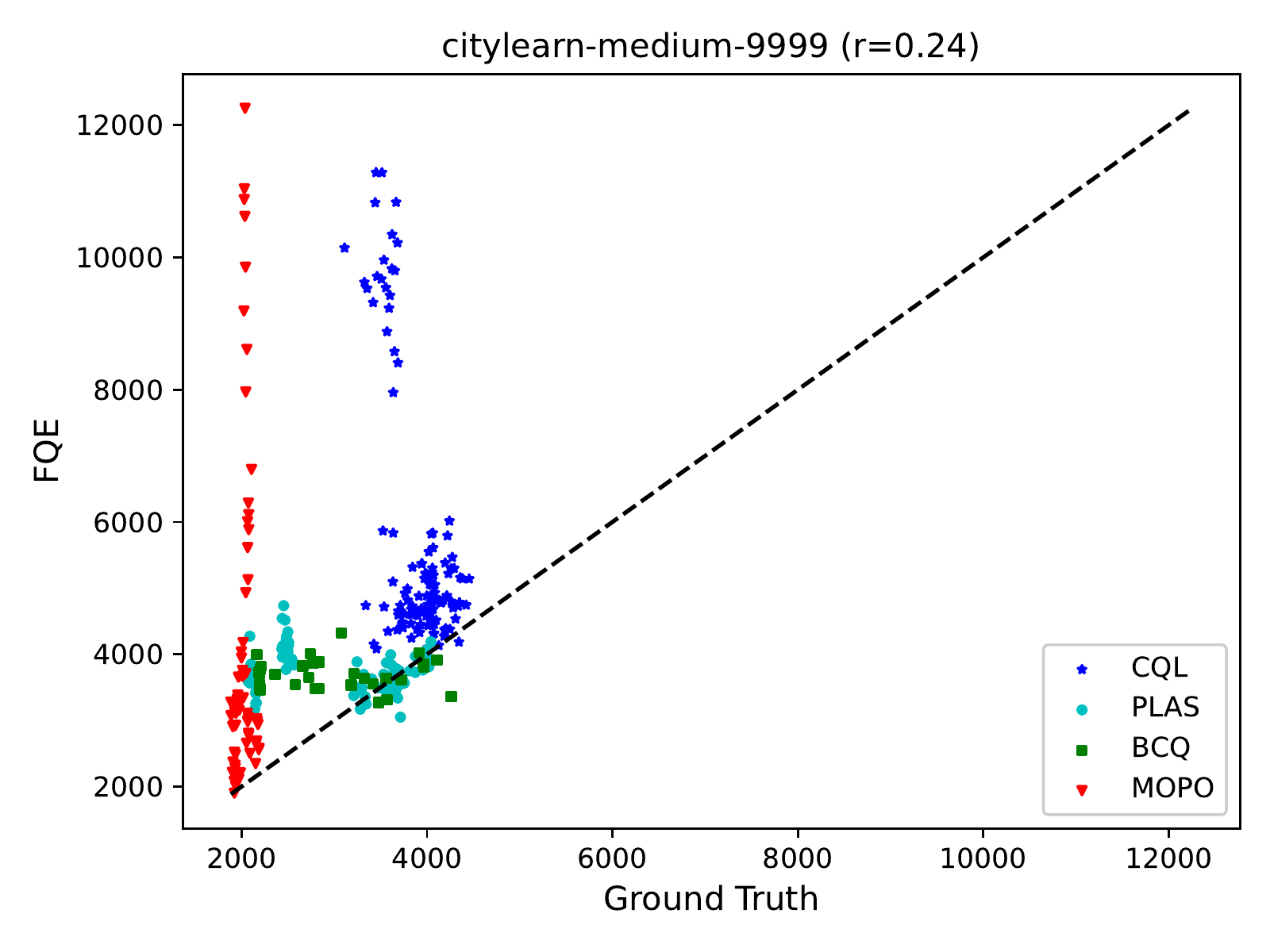}
	\end{minipage}
	
	\begin{minipage}[h]{0.3\linewidth}
		\includegraphics[width=\linewidth]{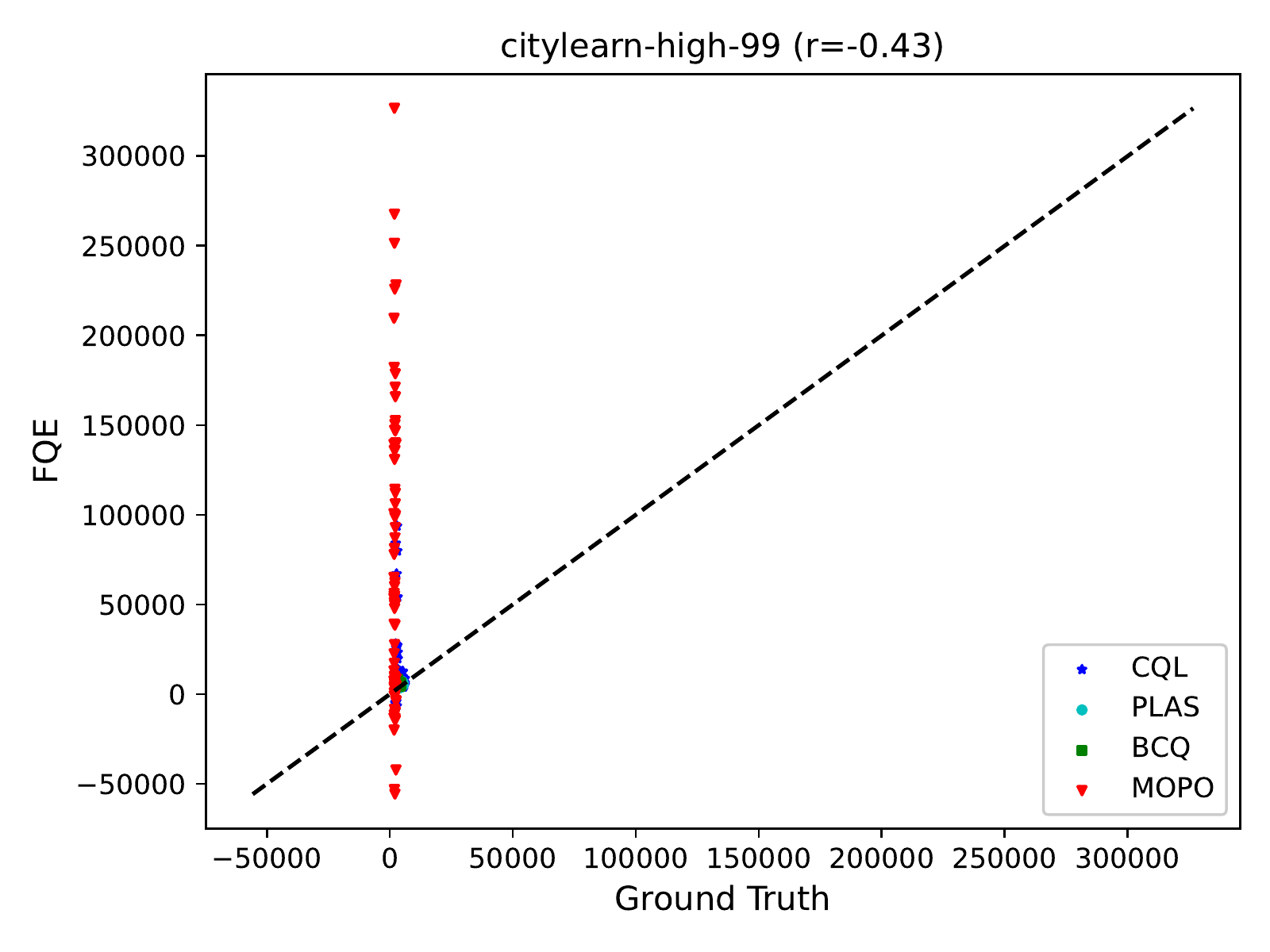}
	\end{minipage}
	\begin{minipage}[h]{0.3\linewidth}
		\includegraphics[width=\linewidth]{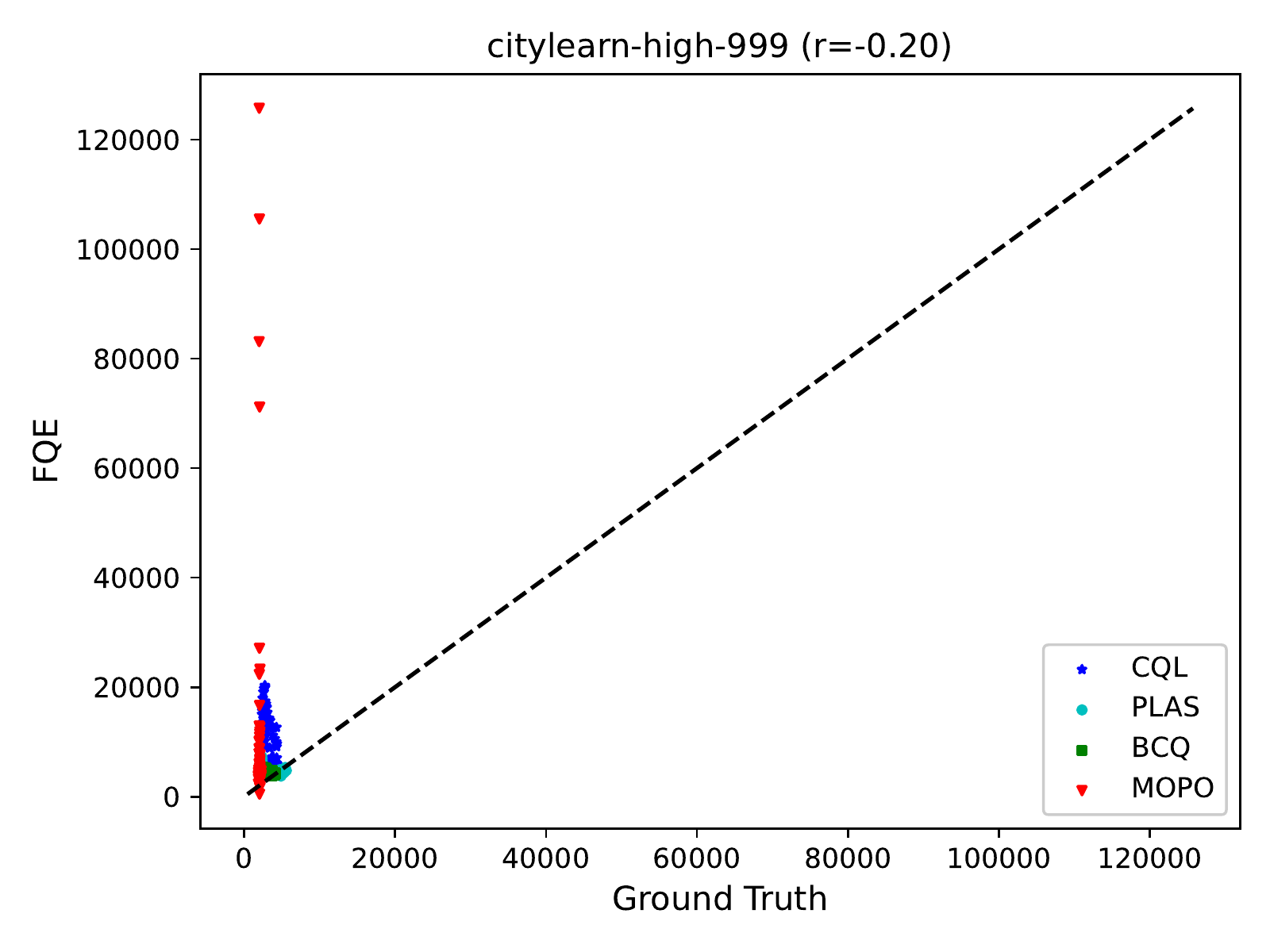}
	\end{minipage}
	\begin{minipage}[h]{0.3\linewidth}
		\includegraphics[width=\linewidth]{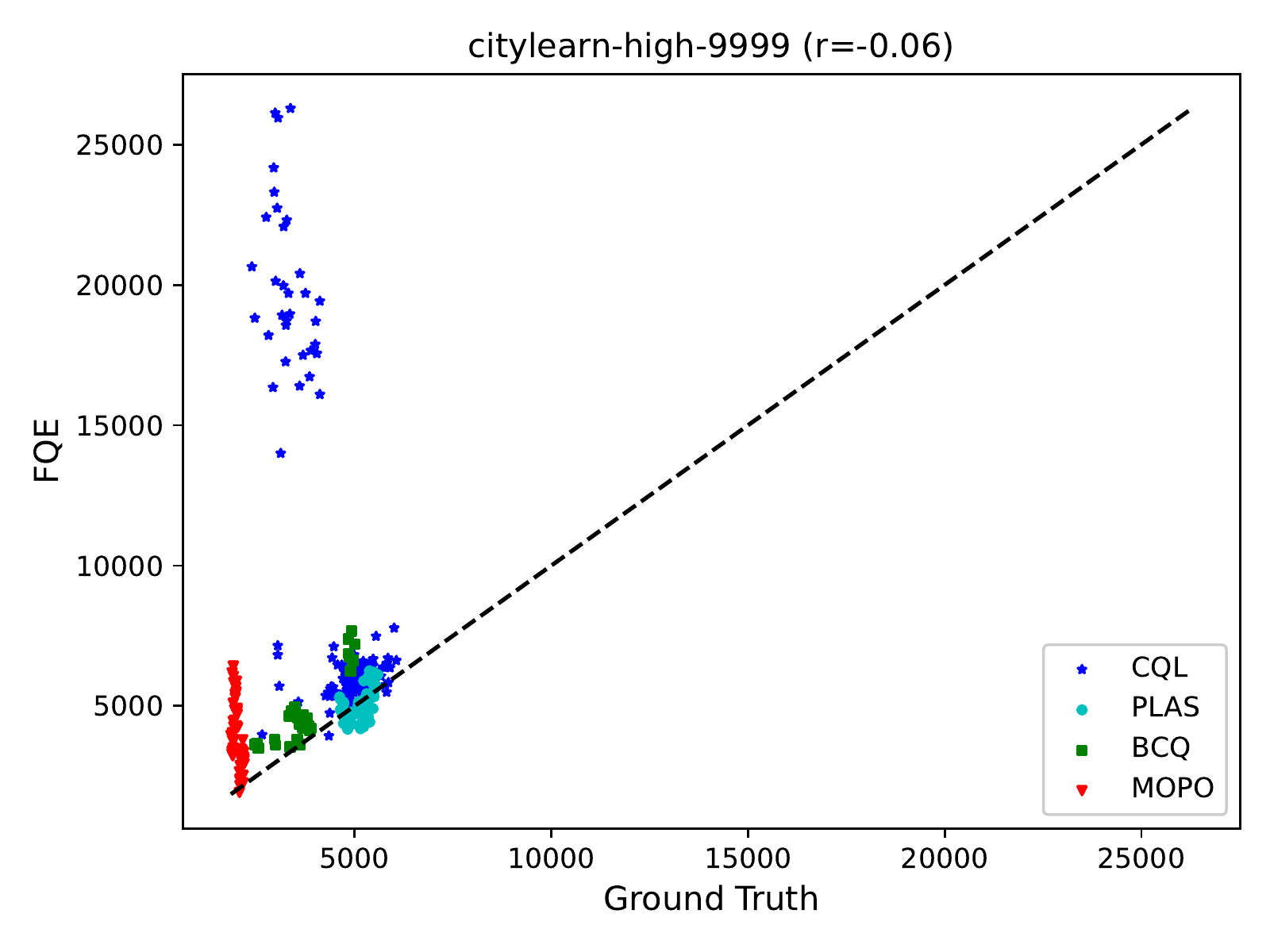}
	\end{minipage}
\end{figure*}

\end{document}